\documentclass[format=acmsmall, authorversion]{acmart}
\usepackage{dutchcal}
\usepackage{multirow}
\usepackage{color}
\usepackage{booktabs}
\usepackage{bm}
\usepackage{chngpage}
\usepackage{makecell}
\usepackage{pifont}
\usepackage{xcolor}
\usepackage{color}
\usepackage[edges]{forest}
\usepackage{titlesec}
\usepackage{eucal}
\usepackage{subcaption}
\usepackage{colortbl}
\usepackage{wrapfig}
\usepackage{array}

\newcolumntype{L}[1]{>{\raggedright\let\newline\\\arraybackslash\hspace{0pt}}m{#1}}
\newcolumntype{C}[1]{>{\centering\let\newline\\\arraybackslash\hspace{0pt}}m{#1}}
\newcolumntype{R}[1]{>{\raggedleft\let\newline\\\arraybackslash\hspace{0pt}}m{#1}}
\ProvidesForestLibrary{edges}
\newcommand{\cmark}{\ding{51}}
\newcommand{\xmark}{\ding{55}}
\newcommand{\term}[1]{\index{\lowercase{#1}}\emph{#1}}
 
\definecolor{green1}{RGB}{34,153,84}
\definecolor{green2}{RGB}{46,125,50}
\definecolor{green3}{RGB}{67,160,71}
\definecolor{green4}{RGB}{102,187,106}
\definecolor{green5}{RGB}{165,214,167}
\definecolor{green_line}{RGB}{0,77,64}
\definecolor{Indianred}{RGB}{205,92,92}
\definecolor{LightCoral}{RGB}{240,128,128}
\definecolor{DarkSalmon}{RGB}{233,150,122}
\definecolor{LightSalmon}{RGB}{255,160,122}
\definecolor{darkred}{RGB}{100,30,22}
\definecolor{lightred}{RGB}{250,219,216}
\definecolor{cottoncandy}{rgb}{1.0, 0.74, 0.85}
\definecolor{gainsboro}{rgb}{0.86, 0.86, 0.86}
\definecolor{mycyan}{cmyk}{.3,0,0,0}
\definecolor{cjh}{rgb}{0,0.8,0.6}

\def\ttr{{\mathtt{r}}}

\definecolor{m1}{HTML}{DDDAF6}
\definecolor{m2}{HTML}{E5DAE2}
\definecolor{m3}{HTML}{C2D3D2}
\definecolor{m4}{HTML}{FDD9C8}
\definecolor{m5}{HTML}{FFF3D1}

\theoremstyle{definition}
\newtheorem{definition}{Definition}[section]

\tikzset{%
    parent/.style =          {align=center,text width=2cm,rounded corners=3pt, line width=0.3mm, fill=gray!10,draw=gray!80},
    child/.style =           {align=center,text width=2.3cm,rounded corners=3pt, fill=blue!10,draw=blue!80,line width=0.3mm},
    grandchild/.style =      {align=center,text width=2cm,rounded corners=3pt},
    greatgrandchild/.style = {align=center,text width=1.5cm,rounded corners=3pt},
    greatgrandchild2/.style = {align=center,text width=1.5cm,rounded corners=3pt},    
    referenceblock/.style =  {align=center,text width=1.5cm,rounded corners=2pt},
    pretrain/.style =           {align=center,text width=1.8cm,rounded corners=3pt, fill=blue!10,draw=blue!80,line width=0.3mm},   
    pretrain_work/.style =           {align=center, text width=5cm,rounded corners=3pt, fill=blue!10,draw=blue!80,line width=0.3mm},  
    template/.style =           {align=center,text width=1.8cm,rounded corners=3pt, fill=red!10,draw=red!80,line width=0.3mm},   
    template_work/.style =           {align=center,text width=5cm,rounded corners=3pt, fill=red!10,draw=red!0,line width=0.3mm},    
    answer/.style =           {align=center,text width=1.8cm,rounded corners=3pt, fill= cyan!10,draw= cyan!80,line width=0.3mm},   
    answer_work/.style =           {align=center,text width=5cm,rounded corners=3pt, fill= cyan!10,draw= cyan!0,line width=0.3mm},      
    multiple/.style =           {align=center,text width=1.8cm,rounded corners=3pt, fill= orange!10,draw= orange!80,line width=0.3mm},   
    multiple_work/.style =           {align=center,text width=5cm,rounded corners=3pt, fill= orange!10,draw= orange!0,line width=0.3mm},        
    tuning/.style =           {align=center,text width=1.8cm,rounded corners=3pt, fill= magenta!10,draw= magenta!80,line width=0.3mm},   
    tuning_work/.style =           {align=center,text width=5cm,rounded corners=3pt, fill= magenta!10,draw= magenta!0,line width=0.3mm},
    part0/.style =           {align=center,text width=1.8cm,rounded corners=3pt, fill=gainsboro,draw=gray!80,line width=0.3mm},   
    part1/.style =           {align=center,text width=1.8cm,rounded corners=3pt, fill=Indianred,draw=darkred,line width=0.3mm}, 
    part1_work/.style = {align=center,text width=5.5cm,rounded corners=3pt, fill=Indianred,draw=darkred,line width=0.3mm}, 
    part2/.style =           {align=center,text width=1.8cm,rounded corners=3pt, fill=LightCoral,draw=darkred,line width=0.3mm},   
    part3/.style =           {align=center,text width=1.8cm,rounded corners=3pt, fill=LightSalmon,draw=darkred,line width=0.3mm}, 
    part4/.style =           {align=center,text width=1.8cm,rounded corners=3pt, fill= cottoncandy,draw= darkred,line width=0.3mm},
    part5/.style =           {align=center,text width=1.8cm,rounded corners=3pt, fill= orange!10,draw= orange!80,line width=0.3mm},
}

\AtBeginDocument{%
  \providecommand\BibTeX{{%
    \normalfont B\kern-0.5em{\scshape i\kern-0.25em b}\kern-0.8em\TeX}}}

\setcopyright{acmcopyright}
\copyrightyear{2023}
\acmYear{2023}
\acmDOI{XX.XXXX/XXXXXXX.XXXXXXX}
\acmJournal{CSUR}

\begin{document}
\title{Knowledge Graph Embedding: A Survey from the Perspective of Representation Spaces}

\author{Jiahang Cao}
\affiliation{%
    \institution{Sun Yat-sen University}
    \streetaddress{No.132, EastWaihuan Road}
    \city{Guangzhou}
    \country{China}
    \postcode{510006}
}
\authornotemark[1]
\email{caojh7@mail2.sysu.edu.cn}

\author{Jinyuan Fang}
\affiliation{%
    \institution{Sun Yat-sen University}
    \streetaddress{No.132, EastWaihuan Road}
    \city{Guangzhou}
    \country{China}
    \postcode{510006}
}
\email{fangjy6@gmail.com}

\authornote{Both authors contributed equally to the paper.}

\author{Zaiqiao Meng}
\affiliation{%
    \institution{University of Glasgow}
    \city{Glasgow}
    \country{UK}
}
\authornotemark[2]
\email{zaiqiao.meng@glasgow.ac.uk}

\author{Shangsong Liang}
\affiliation{%
    \institution{Sun Yat-sen University}
    \streetaddress{No.132, EastWaihuan Road}
    \city{Guangzhou}
    \country{China}
    \postcode{510006}
}
\affiliation{%
    \institution{Mohamed bin Zayed University of Artificial Intelligence}
    \streetaddress{Masdar City}
    \city{Abu Dhabi}
    \country{UAE}
    \postcode{00001}
}
\email{liangshangsong@gmail.com}

\authornote{Corresponding authors.}

\renewcommand{\shortauthors}{Cao et al.}
\begin{abstract}
     Knowledge graph embedding (KGE) is an increasingly popular technique that aims to represent entities and relations of knowledge graphs into low-dimensional semantic spaces for a wide spectrum of applications such as link prediction, knowledge reasoning and knowledge completion. In this paper, we provide a systematic review of existing KGE techniques based on representation spaces. Particularly, we build a 
    fine-grained classification to categorise the models based on three mathematical perspectives of the representation spaces: (1) Algebraic perspective, (2) Geometric perspective, and (3) Analytical perspective. We introduce the rigorous definitions of fundamental mathematical spaces before diving into KGE models and their mathematical properties. We further discuss different KGE methods over the three categories, as well
    as summarise how spatial advantages work over different embedding needs. By collating the experimental results from downstream tasks, we also explore the advantages of mathematical space in different scenarios and the reasons behind them. We further state some promising research directions from a representation space perspective, with which we hope to inspire researchers to design their KGE models as well as their related applications with more consideration of their mathematical space properties.
\end{abstract}




\maketitle

\section{Introduction}

\begin{figure}[t]
    \begin{subfigure}[b]{0.2\textwidth}
        \centering
        \includegraphics[scale=0.25]{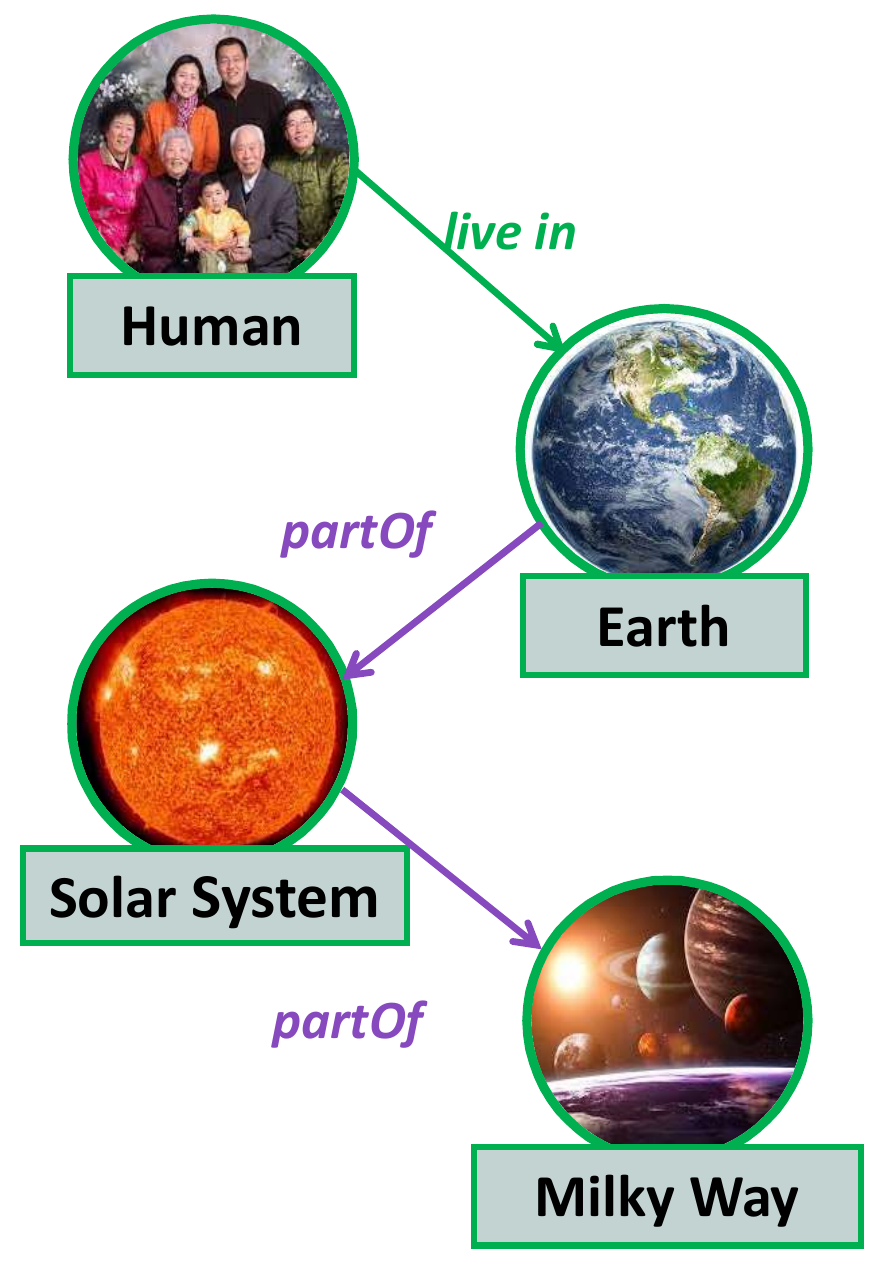}
    \caption{Chain structure.}
    \label{fig:KG_chain}
    \end{subfigure}
    \begin{subfigure}[b]{0.276\textwidth}
        \centering
        \includegraphics[width=1\linewidth]{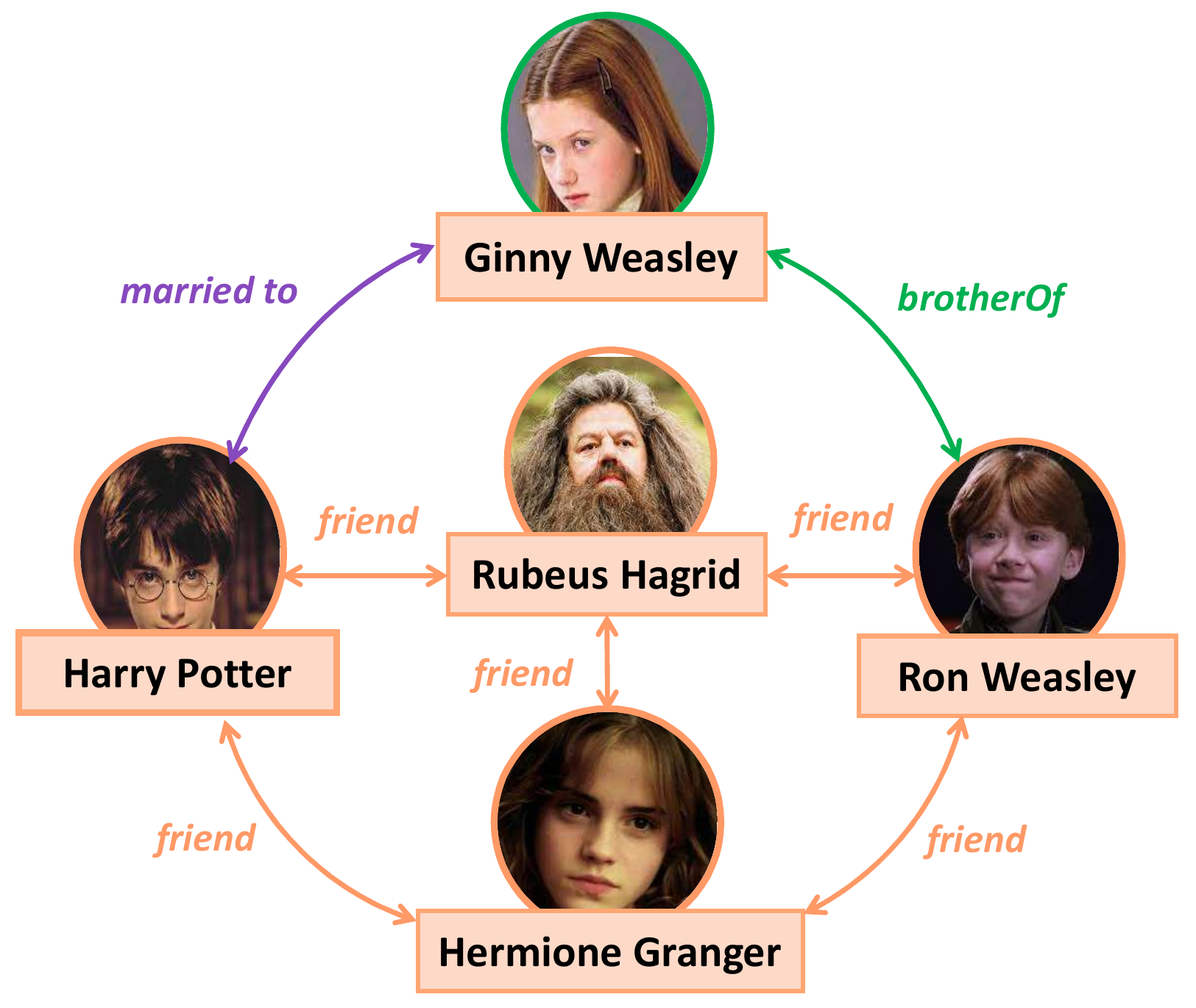}
        \caption{Ring structure.}
    \label{fig:KG_ring}
    \end{subfigure}
    \centering
    \begin{subfigure}[b]{0.425\textwidth}
        \centering
        \includegraphics[width=1\linewidth]{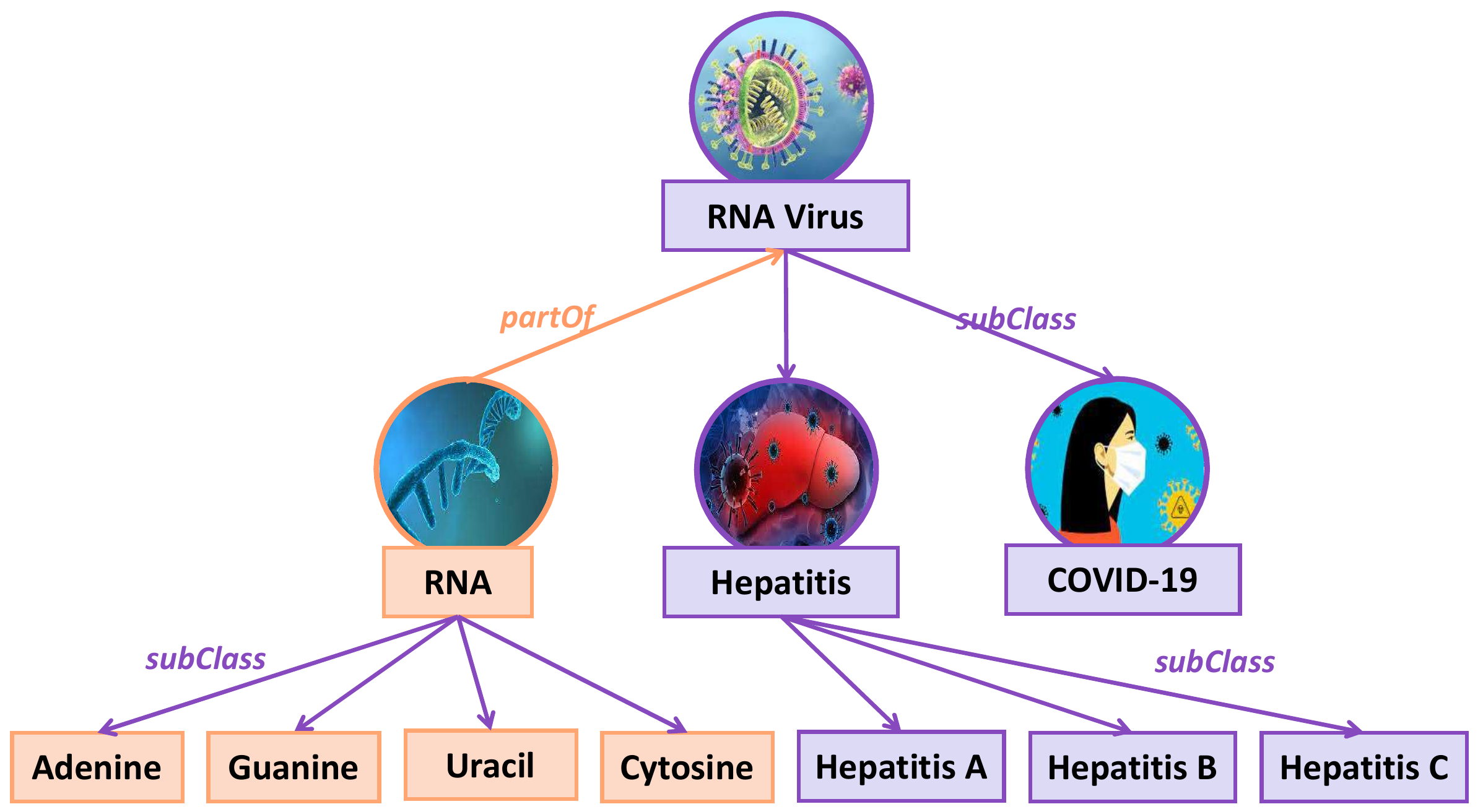}
        \caption{Hierarchy structure.}
    \label{fig:KG_hie}
    \end{subfigure}
    \caption{An illustration of three types of KG structures. (a) shows the most common chain structure in KGs, which can usually be directly modelled in Euclidean space (e.g., TransE, TransH, etc.). (b) is a ring structure of KGs that can be captured in hypersphere space~\cite{cao2022geometry}. (c) stands for hierarchical structure which is usually encoded in hyperbolic or spherical spaces~\cite{chami2020low,sun2020knowledge}. }
    \label{fig:illu_KG_struc}
\end{figure}

Knowledge Graphs are a type of multi-relational graphs that store factual knowledge in real-world. Nodes in KGs represent real-world entities (e.g., names, events and products) and edges represent the relationships between entities. 
Normally, a KG can be efficiently stored as knowledge triples, where each triple consists of two entities and one factual relation between them (i.e., \textit{<head entity, relation, tail entity>}). 
For example, in the triple \textit{<RNA virus, subclass, COVID-19}>, \textit{RNA virus} and \textit{COVID-19} are real-world entities and \textit{subclass} {represents} the relation between \textit{RNA virus} and \textit{COVID-19}.
Over the recent years, rapid growth has been witnessed in building large-scale KGs, such as YAGO~\cite{suchanek2007yago}, Wikidata~\cite{vrandevcic2014wikidata}, Freebase~\cite{bollacker2008freebase} and DBepedia~\cite{auer2007dbpedia}. Due to their effectiveness in storing and representing factual knowledge, they have been successfully applied in question answering~\cite{saxena2020improving,yasunaga2021qa}, recommendation system~\cite{zhou2020improving,sun2020multi}, information retrieval~\cite{wise2020covid,gaur2022iseeq} and other domain-specific applications~\cite{long2020integrated,li2020real}.
Despite the KGs are effective in representing structured factual information, 
they are difficult to manipulate due to the large-scale and {complicated graph structure, i.e., the relationships between entities are intricate and complex, such as the ring structure and hierarchy structure depicted in Figure~\ref{fig:KG_hie}}. 
Therefore, how to effectively and efficiently extract and 
leverage useful information in large-scale KGs for downstream tasks, such as link prediction~\cite{sun2019rotate,zhang2019quaternion,chami2020low} and entity classification~\cite{yu2022jaket,xie2016representation,ji2015knowledge}, is a tough task. To tackle this challenging task, the \textbf{Knowledge Graph Embedding} (KGE) technique was proposed, and has been receiving a lot of attention in the machine learning community~\cite{bordes2013translating,sun2019rotate,chami2020low,10.1145/3447772,liang2021cross,meng2021mixture}. The essential idea of KGE is to learn to embed entities and relations of a KG into a low-dimensional space (i.e. vectorial embeddings), where the embeddings 
are required to preserve the semantic meaning and relational structure of the original KG. The learned embeddings of entities and relations can then be leveraged to solve downstream applications, such as KG completion~\cite{zhang2021drug,abboud2020boxe,zhang2020few,bordes2013translating, wang2014knowledge}, question answering~\cite{yasunaga2021qa,zheng2021knowledge,dong2015question,xu2016question,liang2021profiling}, information extraction~\cite{fei2021enriching,zhang2021fine,liang2016dynamic,hoffmann2011knowledge,wu2010open} and entity classification~\cite{li2021robust,shen2012linden}. 

Many KGE techniques have been proposed to learn the embeddings of entities and relations in KGs~\cite{sun2019rotate,wang2014knowledge,lin2015learning,ji2016knowledge,xiao2015transa}. 
Some KGE methods propose to learn KG embeddings by preserving \emph{relational patterns} between entities in KGs. For example, in order to capture the transformation relationships between entities, TransE~\cite{bordes2013translating} was proposed to embed KGs into \textit{Euclidean Space} and represent relations between entities as translation vectors between entity embeddings in the vector space. Moreover, in order to preserve and infer other relational patterns including symmetry, antisymmetry, inversion and composition in KGs, RotatE~\cite{sun2019rotate} was proposed to map KGs into \textit{Complex Vector Space}, where relations are represented as rotations between entities.

Another line of KGE methods proposes to learn KG embeddings by preserving \emph{structural patterns} of KGs. This line of works was motivated by the fact that large-scale KGs usually contain many complex and compound structures. 
For example, in Figure~\ref{fig:illu_KG_struc}, we provide an illustration of three typical types of structure patterns in KGs, namely chain structure, ring structure and hierarchy structure.
In order to effectively capture the hierarchy structures in KGs, ATTH~\cite{chami2020low} was proposed to embed KGs into  \textit{Hyperbolic Space} with trainable curvatures, where richer transformations can be used to separate nodes than Euclidean space~\cite{nickel2017poincare}, while capturing logical patterns simultaneously. 

In addition, some KGE methods also try to embed KGs in other mathematical spaces to model some desirable properties in KGs. For example, KG2E~\cite{he2015learning} is the first {``density-based” embedding~\cite{vilnis2014word} technique,} 
{which uses Gaussian distributions as embeddings instead of deterministic vectors, to model the uncertainties of entities and relations.} Moreover, TorusE~\cite{ebisu2018toruse} chooses a compact \textit{Lie Group} as its embedding manifold to deal with the regularisation problems, and ModulE~\cite{chai2022module} also {introduces} \textit{Group} theory to model both entities and relations as group element, which can accommodate and outperform most of the existing KGE models.

From the perspective of representation space, we found that {the} above KGE methods mostly learn embeddings in different mathematical spaces, e.g., Euclidean space, Hyperbolic space and Probability space, to capture different relational and structural patterns in KGs. Indeed, different mathematical spaces have their unique strengths, which are beneficial to capture different patterns and properties in KGs. 
{Therefore, we argue that representation space plays a significant role in KGE methods, as it determines the patterns and properties of KGs that can be captured and preserved by KG embeddings.} In addition to the KGE domain, some studies~\cite{murphy2012machine,peng2021hyperbolic,burnaev2021manifold} also demonstrate the importance of mathematical space in traditional machine learning.

Some surveys have been devoted to discussing traditional machine learning models from the perspective of mathematical space~\cite{murphy2012machine,peng2021hyperbolic}.
However, there is not yet a systematic review of KGE methods from the perspective of mathematical space. 
Existing surveys about KGE methods focus either on the encoding model or the applications of KGE methods. For example, 
Wang et al.~\cite{wang2017knowledge} classify KGE methods based on their embedding functions and categorise them into three folds: translation-based models, semantic matching models and additional information-based models. Ji et al.~\cite{ji2021survey} provide a full-scaled view to introduce KGE from four aspects: 
{representation learning}, scoring function, encoding models and auxiliary information. 
Lu et al.~\cite{lu2020utilizing} survey KGE methods with a concentration on utilising textual information.

\begin{figure}[t]
    \centering
    \resizebox{0.95\textwidth}{!}{
        \includegraphics[scale=0.5]{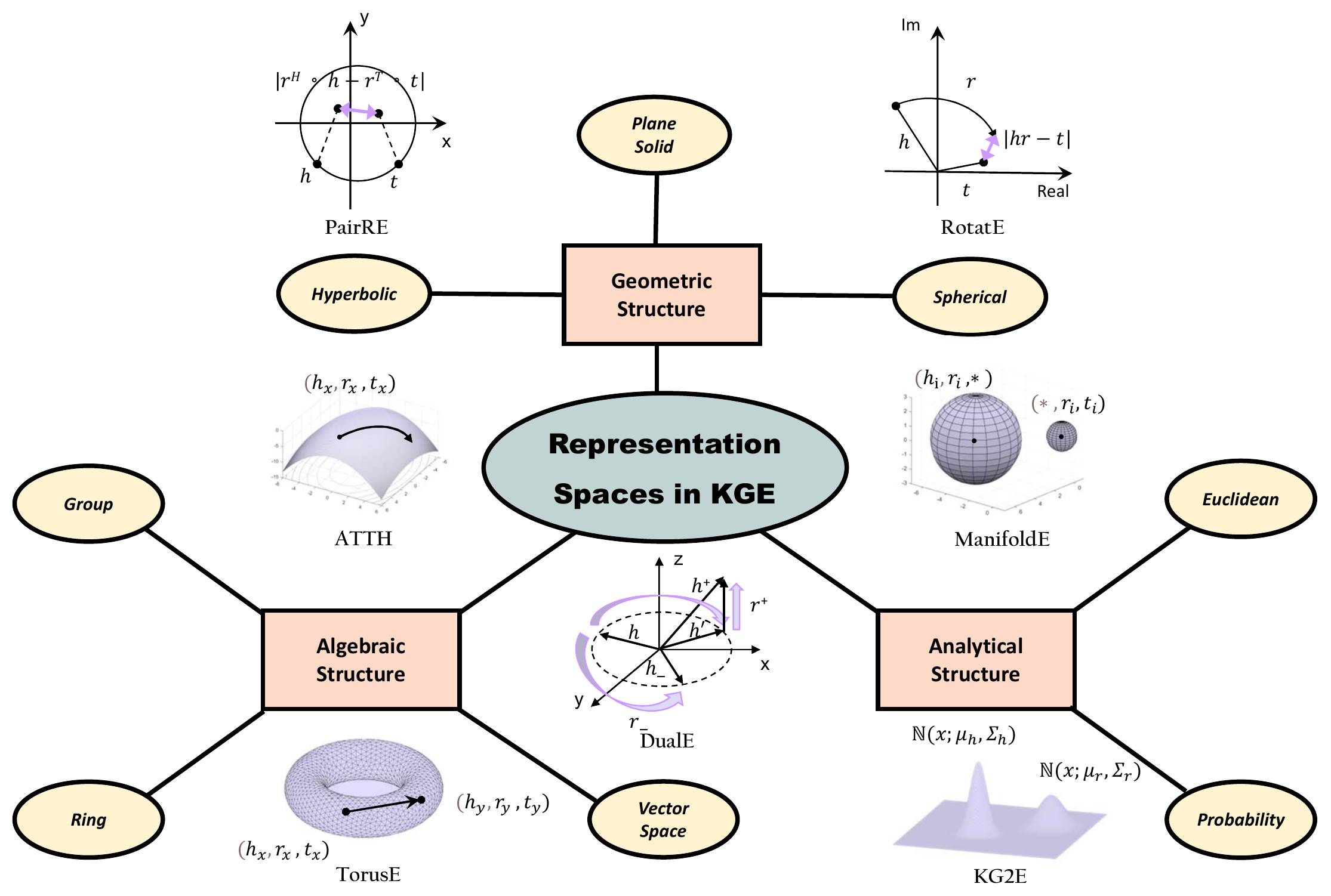}
    }
    \vspace{-1em}
    \caption{Three perspectives and corresponding instances for introducing representation spaces in knowledge graph embedding: (a) Algebraic Structure. (b) Geometric Structure.  (c) Analytical Structure.}
    \label{fig:rep_in_kge}
\end{figure}

\textit{Accordingly, in this paper we aim at providing a comprehensive survey on representation spaces for knowledge graph embedding techniques,
summarising different properties of representation spaces as well as providing guidance for building KGE methods. }
In order to have a better understanding of KGE methods from a novel spatial view, inspired by the fundamental mathematical space system, we build a systematic, comprehensive and multi-angle categorisation to classify existing KGE methods based on their representation spaces. 
Specifically, we propose to classify existing KGE methods into three categories, namely \textbf{Algebraic Structure}, \textbf{Geometric Structure} and \textbf{Analytical Structure}. 
{These three structures have their own mathematical focus, but they are intrinsically linked and jointly make the mathematical system more complete and concrete~\cite{demailly2012analytic,pflaum2001analytic}.}
Figure \ref{fig:rep_in_kge} provides an overview of our classification framework and some representative KGE methods that belong to each category (the detailed version can be found in Section~\ref{sec:spaces_in_KGE}). In this survey, we will introduce the definitions and properties of the above three mathematical structures and introduce some representative KGE methods that belong to these categories in detail. Moreover, we will summarise the experimental results of different KGE methods and provide some suggestions and guidance for building more expressive and powerful KGE methods. Furthermore, we will point out new trends and further directions of KGE methods from the perspective of representation space. 

{Recently, LLMs such as T5~\cite{raffel2020exploring} and GPT-4\footnote{\url{https://openai.com/gpt-4}} have achieved remarkable success in a variety of natural language processing tasks, such as text generation~\cite{li2022pretrained}, machine translation~\cite{lewis2019bart} and question answering~\cite{izacard2020leveraging}. Despite their success in many applications, LLMs still suffer from the following shortcomings: (1) They are known to suffer from the hallucination issue~\cite{ji2023survey,bang2023multitask}, i.e., generating statements that are not factually correct. (2) Since LLMs are pretrained on some general domain corpus, they may not generalise well on some domain-specific tasks, such as biomedical tasks. (3) LLMs are black-box models and it is difficult for them to provide sufficient explainability for their predictions, which is critical in some medical tasks~\cite{pan2023unifying}. In comparison, KGs store factual knowledge about the real world in a structured way and may provide a potential solution to address the shortcomings of LLMs. Indeed, there are some recent works that leverage KGs to enhance the performance and explainability of LLMs. In order to alleviate the hallucination issue, some work \cite{zhang2019ernie,rosset2020knowledge,shen20exploiting,wang21kepler} propose to incorporate KGs into the pretraining of LLMs to encode factual knowledge. For example, the KEPLER~\cite{wang21kepler} proposes to learn LLMs and KGE in a unified manner, which is achieved by using a combination of KGE objective and masked language modelling objective for training. Moreover, there are some works~\cite{lin19kagnet,yasunaga21qa,meng21mixture,zhang22greaselm} that leverage domain-specific KGs to enhance the performance of LLMs on some domain-specific tasks. For example, the MoP model~\cite{meng21mixture} infuses the biomedical knowledge stored in the biomedical knowledge graph UMLS~\cite{bodenreider2004unified} into different BERT models to enhance their performance on several downstream biomedical tasks. Furthermore, there are also some works that leverage KGs to improve the explainability of LLMs. One line of research focuses on using KGs for LLM probing, which aims to understand the relational knowledge stored in the LLMs~\cite{petroni19language,jiang20how,meng22rewire}. Particularly, LAMA~\cite{petroni19language} is the first work to probe the relational knowledge in BERT using KGs. LAMA evaluates the knowledge in BERT by converting facts in KGs into cloze statements and using LLMs to predict the missing entity. Therefore, we believe KGs and KGEs are still useful in the context of LLMs since they can be complementary to LLMs, which is helpful in improving the performance and explainability of LLMs~\cite{pan2023unifying}.}

{To the best of our knowledge, we are the first survey to summarise KGE models by establishing a comprehensive mathematical spatial architecture.} To sum up, the contributions of our work can be summarised as follows:
\begin{itemize}
    \item This is the first paper that comprehensively surveys the relationships between mathematical spaces and KGE techniques. Particularly, we summarise properties of different mathematical spaces used in KGE methods, 
    so as to clearly understand their mathematical properties for different KGE approaches.
    \item We categorise existing KGE models according to their representation spaces, while providing detailed descriptions and comparisons of these works from the perspective of mathematical spaces. 
    \item We provide ideas of space selection for the KGE task based on our analysis on the essential properties of different spaces, which could help researchers and practitioners better understand the space characteristics, and provide guidance for building their KGE models (including loss function, optimisations, etc.). 
    \item 
    We put forward some suggestions and future directions for the KGE tasks by showing some unique properties in different mathematical spaces/structures. These properties can be inspired and generalised to other scenarios such as natural language processing, 
    computer vision, 
    etc., not only for the KGE task.
\end{itemize}

The remainder of this article is organised as follows. 
Section~\ref{sec:preliminaries} introduces notations and the rigorous definitions of fundamental mathematical spaces, as well as the relationships between them. This section will provide some preliminary knowledge about various representation spaces, and build connection between these spaces and the three key components of KGE models (i.e. embedding mapping, score function and representation training). 
Since the fundamental mathematical spaces could not cover diverse spaces used in existing KGE methods, 
we develop a systematic and comprehensive framework to categorise KGE methods from the perspective of representation space. To highlight the excellent effect that different mathematical features could give to KGE, section~\ref{sec:spaces_in_KGE} introduces the proposed classification category, properties of different spaces, as well as
summarise how spatial advantages work in KGE models. Subsequently, Section~\ref{sec:applications} introduces some spatially related KG downstream tasks. Through the results, the advantages of mathematical space in particular scenes and which features are critical to the tasks are well summarised.
Finally, we present our conclusion and future work in Section~\ref{sec:futuredirec}, in which we summarise the respective strengths of three different mathematical structures and the reasons behind them, which will help inspire us to construct state-of-the-art algorithms in more fields, no limited to KGE.

\section{Preliminaries}
\label{sec:preliminaries}
In this section, we introduce the notations used throughout the paper, provide two lines of preliminary knowledge related to our survey: knowledge graph embedding (KGE) and fundamental mathematical spaces.
Specifically, we introduce our notations in subsection~\ref{subsec:notationsAndBackground}, provide an overview of knowledge graph embedding methods in subsection~\ref{subsec:kge} and briefly introduce some basic mathematical spaces and their relationships in subsection~\ref{subsec:relationship}.

\subsection{Notations and Mathematical Background}
\label{subsec:notationsAndBackground}

The definitions of some mathematical terminologies and symbols that appear in the text are shown in Table \ref{tab:symbol_tem}. We denote mathematical spaces 
by blackboard bold characters (e.g., $\mathbb{S}$ denotes topological space). Particularly, we use $\mathbb{R}$ and $\mathbb{C}$ to denote the field of real numbers and the field of complex numbers, respectively. We represent scalars with normal characters (e.g., $x \in \mathbb{R}$ denotes a real scalar), while vectors and matrices are denoted by the bold lowercase characters (e.g., $\mathbf{z} \in \mathbb{R}^n$ denotes a real vector) and the bold uppercase characters (e.g., $\mathbf{X} \in \mathbb{R}^{m \times n}$ denotes a real-valued matrix), respectively.

We represent a knowledge graph as $\mathcal{G} = \{\mathcal{E}, \mathcal{R}, \mathcal{T}\}$, where $\mathcal{E}$ denotes the set of entities (nodes), $\mathcal{R}$ denotes the set of relations (the types of edges) and $\mathcal{T}$ represents the relational facts (edges) in the knowledge graph. Facts observed in $\mathcal{G}$ are stored as a collection of triples: $\mathcal{T} =\{ (h, r, t) \}$, where each triple consists of a head entity $h \in \mathcal{E}$, a tail entity $t \in \mathcal{E}$, and a relation $r \in \mathcal{R}$ between them, e.g., \textit{<Beijing, isCapitalOf, China}> represents the fact that Beijing is the capital of China. We use lower case and bold character to denote the embeddings of entities and relations. Specifically, for a fact triple $(h, r, t)$, we represent the embeddings of head entity, tail entity and the relation between them as $\mathbf{h}$, $\mathbf{t}$ and $\mathbf{r}$, respectively.

\begin{table*}[!t]
    \centering
    \footnotesize
    \renewcommand{\arraystretch}{1.2}
    \caption{Descriptions of symbols and terminologies.}
    \begin{tabular}{L{3cm}L{8.8cm}<{}}
        \toprule
        \textbf{Symbol or Terminology}               & \multicolumn{1}{c}{\textbf{Definition}}                                               \\ \midrule
         $\varnothing $                               & The set containing no elements is called an empty set $\varnothing $.     \\ \midrule
        $\mathbb{R}^n$,$\mathbb{C}^n$...             & Commonly, $\mathbb{R}^n$ represents $n$-dimensional (Real) Vector Space, $\mathbb{C}^n$ denotes $n$-dimensional Complex Vector Space. Other spaces' symbols will be explained in additional detail below.                                                                                                  \\ \midrule
        $\mathbb{S}$,   $\tau$              & A set $\mathbb{S}$  is a finite or infinite collection of objects in which order has no significance, and multiplicity is generally ignored. And the set $\tau$ is an \textit{open set} if every point in $\tau$ has a neighbourhood lying in the set.                           \\ \midrule
        \textit{Intersection}($\cap$), union($\cup$) & The intersection of two sets $\mathcal{A}$ and $\mathcal{B}$ is the set of elements common to $\mathcal{A}$ and $\mathcal{B}$($\mathcal{A}\cap\mathcal{B}$). The union of two sets $\mathcal{A}$ and $\mathcal{B}$ is the set obtained by combining the members of each($\mathcal{A}\cup\mathcal{B}$). \\ \midrule
        \textit{Field}                               & A field is any set of elements that satisfies the field axioms for both addition and multiplication and is a commutative division algebra.                                                                                                                                                             \\ \midrule
        \textit{Complete(ness)}                      & A space is a complete metric space in which every Cauchy sequence is convergent.                                                                                                                                                                                                                       \\ \midrule
        \textit{Complex conjugate}                   & The complex conjugate of a complex number $z=a+bi$ is defined to be $\overline{z}=a-bi$.                                                                                                                                                                                                               \\ \midrule
        \textit{Homeomorphism}                       & A homeomorphism, also called a continuous transformation, is an equivalence relation and one-to-one correspondence between points in two geometric figures or topological spaces that is continuous in both directions.                                                                                \\
        \bottomrule
    \end{tabular}
    \label{tab:symbol_tem}
\end{table*}

\subsection{Knowledge Graph Embedding}
\label{subsec:kge}
\emph{Given a knowledge graph $\mathcal{G}$, the goal of KGE is to learn a mapping function $f$, which projects the entities and relations in $\mathcal{G}$ into a dense and low-dimensional space. The learned embeddings are expected to preserve the structural and attribute information of the original knowledge graph as much as possible, such that they can be leveraged to effectively and efficiently infer the relationships between entities. }

Normally, the paradigm of learning knowledge graph embeddings consists of three components, namely \textit{embedding mapping}, \textit{score function} and \textit{representation training}. The \textit{embedding mapping component} projects the entities and relations to a low-dimensional space and represents them as embedding vectors. A variety of mathematical spaces have be used to define the embeddings of entities and relations. For example, TransE~\cite{bordes2013translating} proposes to learn embeddings in Euclidean space to model the transformation of entities, while RotatE~\cite{sun2019rotate} proposes to embed KGs into complex vector space to model symmetry/antisymmetry of relations. 
Additionally, many other mathematical spaces have also been leveraged in KGEs, such as probability space~\cite{xiao2015transg,he2015learning}, hyperbolic space~\cite{balazevic2019multi,pan2021hyperbolic} and spherical space~\cite{xiao2015one,dong2021hypersphere,cao2022geometry}.
The main focus of this survey is to provide an thorough review of KGE methods from the perspective of embedding space, summarising different properties of different embedding spaces and providing guidance for building KGE methods.

The \textit{score function} is another key component of KGE methods. Score function, denoted as $s(\cdot)$, is a function defined on the mapped embedding space, which is used to measure the plausibility that a triple holds. Specifically, the score function is defined on the embedding vectors of a triple, i.e., $s(\mathbf{h}, \mathbf{r}, \mathbf{t})$, which is supposed to assign higher scores to positive triples (real facts) while assigning lower scores to negative triples (false facts). Based on different representation spaces, different score functions have been proposed to measure the plausibility of triples~\cite{bordes2013translating,sun2019rotate,xiao2015one}. For example, in Euclidean representation space, the translation-based score function $s(\mathbf{h}, \mathbf{r}, \mathbf{t}) =- || \mathbf{h} + \mathbf{r} - \mathbf{t}||_{1/2}$~\cite{bordes2013translating} is widely used to measure the confidence that a triple is positive. Based on the different properties of representation spaces, different score functions can be defined. Please refer to Section~\ref{sec:spaces_in_KGE} for detailed discussion of different score functions.

The \textit{representation training} component of KGE methods aims to learn the entity and relation embeddings by maximising the scores of positive triples while minimising the scores of negative triples. Since it is infeasible to obtain the precise positive and negative triples in a knowledge graph, one convention in the field of KGE is to regard observed triples, i.e., $\mathcal{T}$, as positive examples while sampling unobserved triples, i.e., $\mathcal{T}^{-}$, as negative examples~\cite{bordes2013translating,wang2014knowledge}. The negative triples can be generated by randomly replacing the head entity or the tail entity in an observed triple with a random entity sampled from the entity set, i.e., 
\begin{align}
    \mathcal{T}^{-} = & \{ (h^{-}, r, t) \mid (h,r,t) \in \mathcal{T} \wedge (h^{-}, r, t) \notin \mathcal{T} \wedge h^{-} \in \mathcal{E} \} \cup \notag \\
    & \{(h, r, t^{-}) \mid (h,r,t) \in \mathcal{T} \wedge (h, r, t^{-}) \notin \mathcal{T} \wedge t^{-} \in \mathcal{E} \}. 
\end{align}

Given the positive and negative triples (i.e. $\mathcal{T}$ and $\mathcal{T}^{-}$), various objective functions can be used to train representations of KGE models. For example, the \textit{margin-based ranking loss}~\cite{bordes2013translating,wang2014knowledge,lin2015learning,ji2015knowledge} is a widely adopted objective function for training representations of entities and relations. It is defined as: 
\begin{align}
    \mathcal{L}_{margin} = \sum_{(h,r,t) \in \mathcal{T}} \sum_{(h^{-}, r, t^{-}) \in \mathcal{T}^{-}} \max \left(0, \gamma - s(\mathbf{h}, \mathbf{r}, \mathbf{t}) + s(\mathbf{h^{-}, \mathbf{r}, \mathbf{t}^{-}}) \right),
\end{align}
where $\gamma$ is a margin hyperparameter. The margin-based ranking loss assumes that observed triples are more valid than unobserved triples, and therefore favours higher scores of observed triples $\mathcal{T}$ than unobserved triples $\mathcal{T}^{-}$. Another widely used objective function for learning knowledge graph embeddings is the \textit{cross-entropy loss}~\cite{trouillon2016complex,dettmers2018convolutional,lacroix2018canonical,sun2019rotate}, which is defined as:
\begin{align}
    \mathcal{L}_{CE} = \sum_{(h,r,t) \in \mathcal{T} \cup \mathcal{T}^{-}} \log (1 + \exp(-y_{hrt} \cdot s(\mathbf{h}, \mathbf{r}, \mathbf{t}))),
\end{align}
where $y_{hrt} \in \{ -1, 1 \}$ is the label of a triple $(h,r,t)$, and $y_{hrt}=1$ indicates that the triple is a positive example while $y_{hrt}=-1$ indicates that the triple is a negative example. After defining the objective function, the knowledge graph embeddings are learned by minimising the objective function via stochastic optimisation, where a small batch of positive examples and negative examples are sampled for optimisation at each training iteration. It is worth noting that most KGE models~\cite{bordes2013translating,socher2013reasoning,yang2014embedding,trouillon2016complex,nickel2016holographic} have additional constraints that the norm of knowledge graph embeddings is less than or equal to $1$. Such constraints can prevent the model to trivially minimise the objective function by simply increasing the norm of the embedding vectors~\cite{bordes2013translating}.

\subsection{Fundamental Mathematical Spaces}
\label{subsec:relationship}

In general, various types of fundamental mathematical spaces have been widely used for representation learning in computer science. However, there is no easy way to elaborate all these spaces due to the complex inclusion and overlapping of these spaces. Here we first introduce some basic spaces through algebraic definitions. We start by introducing the topological space, which is the most basic mathematical space. 
\textit{Topological Space} is a kind of mathematical structure defined on a set, from which the concept of \textit{convergence, connectivity, continuity} etc., are introduced~\cite{rudin1973functional}.
A special case of topological space is the \textit{Metric Space}, where the \textit{distance} between any elements in the set is introduced in the topological space. The \textit{Vector Space}, also referred to as \textit{Linear Space}~\cite{rudin1973functional}, is another fundamental space that are widely used in representation learning. The vector space is a set that defines both \textit{addition} and \textit{multiplication} satisfying eight operation conditions (see~\S~\ref{subsec:vector_space} for details).
On the basis of vector space, 
\textit{Normed Vector Space} is defined, which additionally introduces the concept of \textit{norm} to define the \textit{length} between elements in the set. Particularly, \textit{Inner Product Space} is a special normed vector space, 
which additionally introduce \textit{inner product} to define the concept of \textit{angle} between elements in the set.
\textit{Euclidean Space} is a finite dimensional inner product space which is widely used nowadays~\cite{lang2012algebra,rudin1973functional}.

\begin{figure}[t]
    \centering
    \includegraphics[scale=0.5]{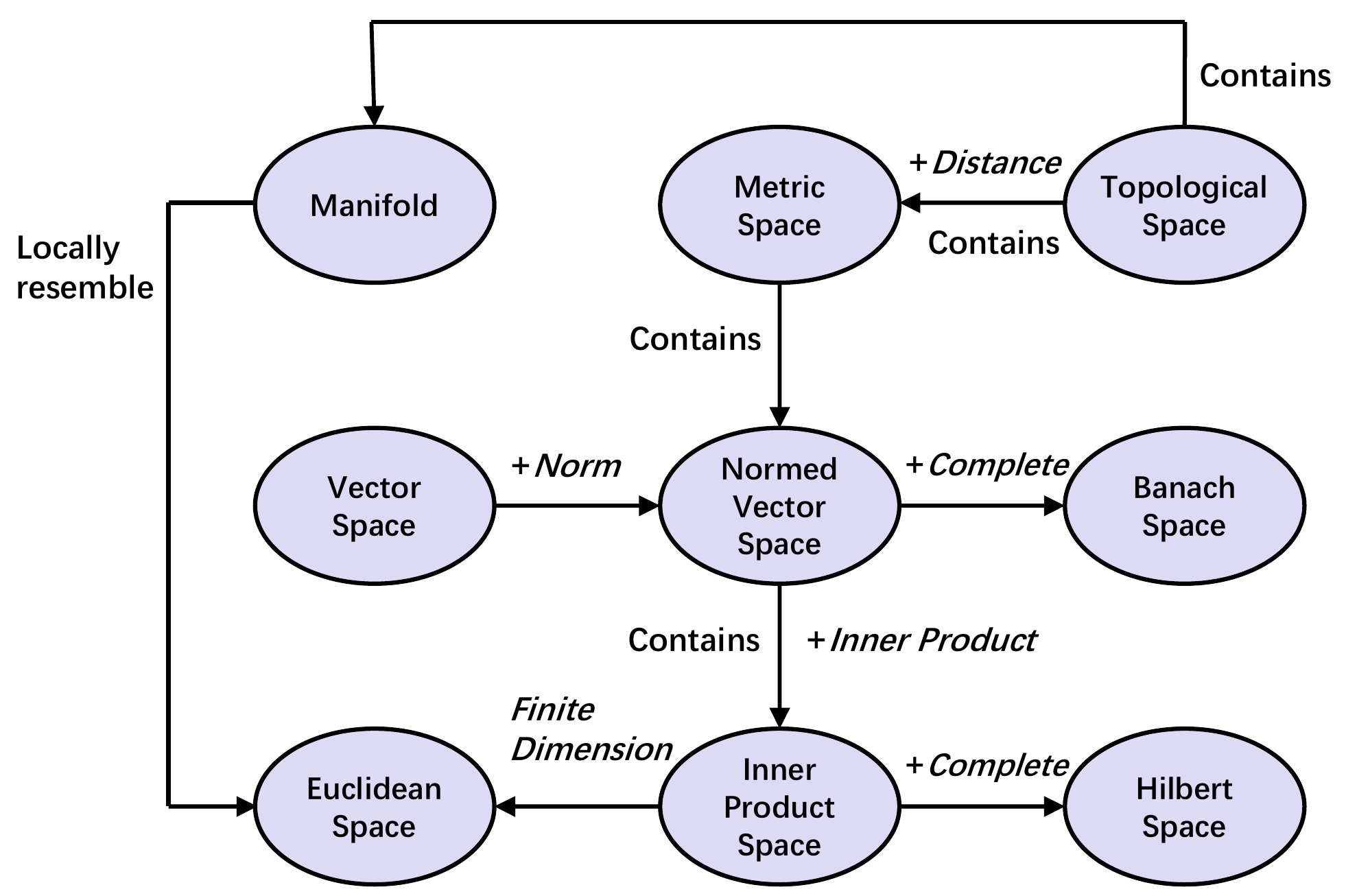}
    \caption{The relationships between different mathematical spaces. The basic spatial relationship is $\left \{ \textit{Inner Product Vector Space} \right \}\subset \left \{ \textit{Normed Vector Space} \right \}\subset\left \{ \textit{Metric Space} \right \} \subset \left \{ \textit{Topological Space} \right \}$. In addition, we tease out more detailed relations, such as the \textit{complete} case of Inner Product Space is Hilbert Space.}
    \label{fig:space_relationships}
\end{figure}

The inclusion relationship of the above spaces can be summarised by: $\left \{ \textit{Inner Product Vector Space} \right \}$ $\subset $ $\left \{ \textit{Normed Vector Space} \right \}\subset\left \{ \textit{Metric Space} \right \} \subset \left \{ \textit{Topological Space} \right \}$. Going from the left to the right, each category is carried by the next one. In inner product vector space, we can use \textit{inner product} to express both the \textit{length} and \textit{angle} of vectors, since the \textit{inner product} induces a \textit{norm}. But in a normed vector space, we could just measure the distance between two points through space metric. Then in metric space and topological space, there is less basic concept that can be measured. In another word, the ``capacity'' of spaces is getting weaker.
To sum up, the \textit{inner product} induces \textit{norm}, \textit{norm} induces distance, and distance induces topology, therefore: \textit{every inner product space is a normed vector space, every normed vector space is a metric space and every metric space is a topological space.} It should be noticed that linear space is an algebraic structure while topological space is a topological structure. So they are not juxtaposed since they are considered in different categories. The relationships between different mathematical spaces are presented in Figure~\ref{fig:space_relationships}. In what follows, we provide rigorous definitions of above mathematical spaces, which also be summarised in Table~\ref{tab:basic_space}.

\subsubsection{Topological Space}
\label{subsec:vectortopo_space}

\begin{definition}
A \textit{topological space} is a set $\mathbb{S}$ in which a collection of subsets $\tau$  (called \textit{open sets}, see Table \ref{tab:symbol_tem}) is specified, with the following properties~\cite{rudin1973functional}: $\mathbb{S}$ is open, $\varnothing $ is open, the intersection (see Table \ref{tab:symbol_tem}) of any two open sets is open, and the union of every collection of open sets is open. Such a collection $\tau$ is called a \textit{topology} on $\mathbb{S}$. When clarity seems to demand it, the topological space corresponding to the topology $\tau$ will be written as $(\mathbb{S}, \tau)$ rather than $\mathbb{S}$. Metric space, uniform space~\cite{treves2016topological,lee2010introduction} are examples of topological space.
\end{definition}

\subsubsection{Vector Space (Linear Space)}
\label{subsec:vector_space}

\begin{definition}
Let $\Phi$ stands for either $\mathbb{R}$ or $\mathbb{C}$, i.e., real field or complex filed (See Table \ref{tab:symbol_tem}). A \textit{scalar} is a member of the \textit{scalar field} $\Phi$. A \textit{vector space} over $\Phi$ is a set $\mathbb{X}$, whose elements are called vectors, and in which two operations, \textit{addition} and \textit{scalar multiplication}, are defined, with the following 
algebraic properties~\cite{rudin1973functional}:
\begin{itemize}
    \item[(a)] 
          For any two vectors $\mathbf{x}$ and $\mathbf{y}$ in $\mathbb{X}$, 
          the addition between them is represented as $\mathbf{x}+\mathbf{y}$, with the following properties:
          \begin{equation}
              \mathbf{x} + \mathbf{y} = \mathbf{y} + \mathbf{x}  \text{ and }  \mathbf{x} + (\mathbf{y} + \mathbf{z}) = (\mathbf{x} + \mathbf{y}) + \mathbf{z} .
          \end{equation}

          The vector space $\mathbb{X}$ contains a unique vector $\mathbf{0}$ (the \textit{zero vector} or \textit{origin} of $\mathbb{X}$) such that $\mathbf{x} + \mathbf{0} = \mathbf{x}$ for every $\mathbf{x} \in \mathbb{X}$, and 
          each $\mathbf{x} \in \mathbb{X}$ corresponds a unique vector $-\mathbf{x}$ such that $\mathbf{x} + (-\mathbf{x}) = \mathbf{0}$.

    \item[(b)] 
          For any scalar $\alpha, \beta \in \Phi$ and vector $\mathbf{x} \in \mathbb{X}$, the scalar multiplication between them is denoted as $\alpha \mathbf{x}$, with the following properties: 
          \begin{equation}
              \alpha (\mathbf{x} + \mathbf{y}) = \alpha \mathbf{x} + \alpha \mathbf{y}, (\alpha + \beta) \mathbf{x} = \alpha \mathbf{x} + \beta \mathbf{x} . 
          \end{equation}
          
          Like the zero vector in vector space, the zero element of the scalar field can be defined in a similar way, which is denoted as $0$. 
\end{itemize}

Particularly, a \textit{real vector space} is the one for which $\Phi = \mathbb{R}$, while a \textit{complex vector space} is the one for which $\Phi = \mathbb{C}$. In the following, any statements about vector spaces in which the scalar field is not explicitly specified means that they can be applied to both real vector space and complex space.
\end{definition}

\subsubsection{Normed Space}

\begin{definition}
A vector space $\mathbb{X}$ is said to be a \textit{normed space} if 
each vector $\mathbf{x} \in \mathbb{X}$ is associated with a nonnegative real number $\left \| \mathbf{x} \right \|$, called the \textit{norm} of $\mathbf{x}$~\cite{rudin1973functional}, in such a way that: 
\begin{itemize}
    \item[(a)] $\left \| \mathbf{x} + \mathbf{y} \right \|\leq \left \| \mathbf{x} \right \| + \left \| \mathbf{y} \right \|$ for all $\mathbf{x}$ and $\mathbf{y}$ in $\mathbb{X}$,
    \item[(b)] $\left \| \alpha \mathbf{x} \right \| = \left | \alpha \right | \left \| \mathbf{x} \right \|$ if $\mathbf{x} \in \mathbb{X}$ and $\alpha$ is a scalar,
    \item[(c)] $\left \|  \mathbf{x} \right \| > 0$ if $\mathbf{x}\neq \mathbf{0}$.
\end{itemize}
\quad The word ``norm'' is also used to denote the \textit{function} that maps $\mathbf{x}$ to $\left \| \mathbf{x}  \right \|$.
Every normed space can be regarded as a particular metric space, in which the distance $d(\mathbf{x},\mathbf{y})$ between $\mathbf{x}$ and $\mathbf{y}$ is defined as $\left \| \mathbf{x} - \mathbf{y} \right \|$. 
The properties of the distance function $d$ in metric space are: 
\begin{itemize}
    \item[(a)] $0\leq d(\mathbf{x},\mathbf{y})< \infty $ for all $\mathbf{x}$ and $\mathbf{y}$,
    \item[(b)] $d(\mathbf{x},\mathbf{y}) = 0$ if and only if $\mathbf{x} = \mathbf{y}$,
    \item[(c)] $d(\mathbf{x},\mathbf{y}) = d(\mathbf{y},\mathbf{x})$ for all $\mathbf{x}$ and $\mathbf{y}$,
    \item[(d)] $d(\mathbf{x},\mathbf{z}) \leq d(\mathbf{x},\mathbf{y}) + d(\mathbf{y},\mathbf{z})$ 
    for all $\mathbf{x},\mathbf{y},\mathbf{z}.$
\end{itemize}
\quad A \textit{Banach Space} is a special normed space which is \textit{complete} (see Table \ref{tab:symbol_tem}) in the metric defined by its norm, which means that every Cauchy sequence is required to converge. {(A Cauchy sequence is a sequence whose elements become arbitrarily close to each other as the sequence progresses~\cite{lang2012algebra}.)}
\end{definition}

\begin{table*}[!t]
    \centering
    \footnotesize
    \renewcommand{\arraystretch}{1.2}
    \caption{Descriptions of the basic spaces in mathematics. It is important to note that they are listed here for a general introduction, not because they are juxtaposed. The specific relationships between these spaces can found in~\S\ref{subsec:relationship}.}
    \label{tab:basic_space}
    \vspace{-0.5em}
    \begin{tabular}{p{3cm}p{5cm}p{3cm}}
        \toprule
        \textbf{Space}             & \textbf{Description}                                                                                        & \textbf{Property}                         \\
        \midrule
        \multirow{2}{*}{\term{Topological Space}}
           & a geometrical space in which \textbf{closeness} is defined.                                                 & \makecell[l]{$\bullet$ \textit{open set}} \\
        \midrule
        \multirow{2}{*}{\term{Vector Space}}
        & \multirow{2}{*}{\parbox{5cm}{a set of vectors in which \textbf{addition} and \textbf{scalar multiplication} are defined.}} & 
        \makecell[l]{$\bullet$ \textit{addition}}  \\ 
        &&\makecell[l]{$\bullet$ \textit{multiplication}}  \\
        \midrule
        \multirow{3}{*}{\term{Normed Space}}        & \multirow{3}{*}{\parbox{5cm}{a vector space over the real or complex numbers, on which a \textbf{norm} is defined. \\ }      }               &
        \makecell[l]{$\bullet$ \textit{addition}}\\
        &&\makecell[l]{$\bullet$ \textit{multiplication}}\\
        &&\makecell[l]{$\bullet$ \textit{norm}: $\left \| x \right \|$}\\
        \midrule
        \multirow{3}{*}{\term{Inner Product Space}} & \multirow{3}{*}{\parbox{5cm}{a vector space with a binary operation called an \textbf{inner product}. } }                                  & \makecell[l]{$\bullet$ \textit{addition}}
        \\ 
        && \makecell[l]{$\bullet$ \textit{multiplication}} \\
        &&\makecell[l]{$\bullet$ \textit{inner product}:$\left \langle x,x \right \rangle$}  \\
        \midrule
        \multirow{3}{*}{\term{Euclidean Space}} & \multirow{3}{*}{\parbox{5cm}{a commonly used \textbf{finite} dimensional inner product space over real numbers.} }                                   & \makecell[l]{$\bullet$ \textit{addition}}  \\ 
        && \makecell[l]{$\bullet$ \textit{multiplication}} \\
        &&\makecell[l]{$\bullet$ \textit{inner product}:$\left \langle x,x \right \rangle$} \\
        \midrule
        \multirow{2}{*}{\term{Manifold}} & a topological space which is \textbf{locally Euclidean}.                                 & \makecell[l]{$\bullet$ each point has \\a certain neighbourhood} \\
        \bottomrule
    \end{tabular}
\end{table*}

\subsubsection{Inner Product Space}

\begin{definition}
A complex vector space $\mathbb{H}$ is called an \textit{inner product space} or \textit{unitary space} if 
each ordered pair of vectors $\mathbf{x}$ and $\mathbf{y}$ in $\mathbb{H}$ is associated with a complex number $\langle\mathbf{x}, \mathbf{y}\rangle$, called the \textit{inner product} or \textit{scalar product} of $\mathbf{x}$ and $\mathbf{y}$, such that the following rules hold~\cite{rudin1973functional}:
\begin{itemize}
    \item[(a)] $\langle\mathbf{y}, \mathbf{x}\rangle = \overline{\langle\mathbf{x}, \mathbf{y}\rangle}$ (The bar denotes complex conjugation (see Table \ref{tab:symbol_tem})),
    
    \item[(b)] $\langle\mathbf{x} + \mathbf{y}, \mathbf{z}\rangle = \langle\mathbf{x}, \mathbf{z}\rangle + \langle\mathbf{y}, \mathbf{z}\rangle$,
    
    \item[(c)] $\langle\alpha \mathbf{x}, \mathbf{y}\rangle $= $\alpha\langle\mathbf{x}, \mathbf{y}\rangle$ if $\mathbf{x} \in \mathbb{H}, \mathbf{y} \in \mathbb{H}, \alpha \in \mathbb{C}$,
    
    \item[(d)] $\langle\mathbf{x}, \mathbf{x}\rangle \geq 0$ for all $\mathbf{x} \in \mathbb{H}$,
    \item[(e)] $\langle\mathbf{x}, \mathbf{x}\rangle = 0$ only if $\mathbf{x} = 0$.
\end{itemize}
Particularly, if the normed space is complete, it is called a \textit{Hilbert Space}.
\end{definition}

\subsubsection{Euclidean Space}

\begin{definition}
\textit{Euclidean space} is the basic space of geometry, intended to represent physical space~\cite{ball1960short}. Generally, Euclidean space refers to Euclidean vector space, which is a finite dimensional inner product space over real numbers. Based on the algebraic definition of Euclidean space, a plane or solid (i.e. Euclidean geometry) can be well represented by lines and points. 
\end{definition}

\subsubsection{Manifold}

\begin{definition}
A topological space $\mathbb{M}$ is said to be a \textit{manifold} or \textit{locally Euclidean of dimension n} if every point of $\mathbb{M}$ has a neighbourhood in $\mathbb{M}$ that is homeomorphic (see Table \ref{tab:symbol_tem}) to an open subset of $n$-dimensional Euclidean space~\cite{lee2010introduction}. Manifolds constitute a generalisation of objects and the concept of a manifold is central to many parts of geometry since it allows complicated structures, such as sphere, curved surface, etc. to be described in terms of well-understood topological properties of simpler spaces.
\end{definition}

\section{Representation Spaces in Knowledge Graph Embedding}
\label{sec:spaces_in_KGE}

Since a KG usually consists of many complicated structures (e.g., 1-to-N, N-to-N, and hierarchical relationships), 
researchers have proposed to embed KGs in different representation spaces in order to better preserve such complicated structural information~\cite{xiao2015one,ebisu2018toruse,zhang2019quaternion,guo2021bique,chami2020low}. Indeed, different representation spaces have their unique structures and properties, as we show in Section~\ref{sec:preliminaries}.
However, in addition to the fundamental mathematical spaces introduced in Section \ref{sec:preliminaries}, there are more spaces that provide better properties for KGE. For example, in hyperbolic space, the region and length increase exponentially with the radius, which provides more available space for embedding task~\cite{balazevic2019multi,liang2019unsupervised,liang2019collaborative,chami2020low,nickel2017poincare,wu2022learning}. Moreover, in Lie group, embedding vectors will never diverge unlimitedly and therefore regularisation of embedding vectors is no longer required for effective learning~\cite{ebisu2018toruse}. 
As a result, KGE methods built on different representation spaces are able to capture and preserve different structural and attribution information in original KGs. However, there is neither a systematic review of KG embedding methods from the perspective of representation spaces, nor any literature showing how to properly choose representation space given particular KGE tasks. In this paper, we aim to fill this gap by summarising KGE methods based on the structures and properties of their mathematical representation spaces. 

Note that in Section~\ref{sec:preliminaries}, we introduced some algebraic definitions of some basic spaces. Based on them, some geometric perspectives, such as Euclidean geometry and hyperbolic geometry, can also be introduced accordingly. At the same time, we notice that there are various kinds of mathematical spaces in KGE, which play a significant role and belong to different mathematical structures. 
The diverse spaces in KGE often have complicated relationships. For example, manifolds and Euclidean geometry have inclusion relations since there are overlapping structures between them. In addition, some spaces are not juxtaposed such as spherical space and probability space because they originate from different mathematical structures, which cannot be discussed in the same category.

\begin{figure*}
    \footnotesize
    \resizebox{1\textwidth}{!}{
    \begin{forest}
        for tree={
        forked edges,
        grow'=0,
        draw,
        rounded corners,
        node options={align=center,},
        text width=4cm,
        s sep=12pt,
        calign=child edge, calign child=(n_children()+1)/2,
        l sep = 10 pt,
        l=40pt
        },
        [Knowledge Graph Embedding, part0,fill=m1,draw=black
        [Geometric Structure , part0,fill=m2,draw=black
        [Euclidean Geometry,part0,fill=m3,draw=black
        [Cartesian Coordinate,part0,fill=m4,draw=black
        [TransE~\cite{bordes2013translating}; RotatE~\cite{sun2019rotate};\\ PairRE~\cite{chao2021pairre}; InterHT~\cite{wang2022interht};\\ TripleRE~\cite{yu2022triplere}; TranS~\cite{zhang2022trans};\\ CompoundE~\cite{ge2022compounde};HopfE~\cite{bastos2021hopfe}; 
        ,fill=m5,draw=white,
        draw=black,align=left,
        inner xsep=1.5mm,text width=4.2cm]
        ]
        [Polar Coordinate,part0,fill=m4,draw=black
        [HAKE~\cite{zhang2020learning}; H$^2$E~\cite{wang2021knowledge};\\ HBE~\cite{pan2021hyperbolic},fill=m5,draw=white,rounded corners=3pt,align=left,
        draw=black,text width=4.2cm,inner xsep=1.5mm]
        ]
        [Spherical Coordinate,part0,fill=m4,draw=black
        [STKE~\cite{wang2022stke},fill=m5,draw=white,rounded corners=3pt,text centered,text width=4.3cm,draw=black]
        ]
        ]
        [Hyperbolic Geometry,part0,fill=m3,draw=black
        [MuRP~\cite{balazevic2019multi}; ATTH~\cite{chami2020low}; \\ HBE~\cite{pan2021hyperbolic}; HyperKA~\cite{sun2020knowledge};\\
        UltraE~\cite{xiong2022ultrahyperbolic}; RotL~\cite{wang2021hyperbolic};\\HypHKGE~\cite{zheng2022hyperbolic}; H$^2$E~\cite{wang2021knowledge};\\
        GIE~\cite{cao2022geometry}; MuRMP~\cite{wang2021mixed},fill=m5,draw=white,rounded corners=3pt,align=left,inner xsep=0pt,draw=black,text width=4cm
        ]
        ]
        [Spherical Geometry,part0,fill=m3,draw=black
        [ManifoldE~\cite{xiao2015one};TransC~\cite{lv2018differentiating};\\ GIE~\cite{cao2022geometry}; MuRS~\cite{wang2021mixed}\\
        HyperspherE~\cite{dong2021hypersphere}; SEA~\cite{gregucci2023link},fill=m5,draw=white,rounded corners=3pt,align=left,inner xsep=0pt,draw=black,text width=4cm]
        ]
        ]
        [Algebraic Structure, part0,fill=m2,draw=black
        [Group , part0,fill=m3,draw=black
            [TorusE~\cite{ebisu2018toruse}; DihEdral~\cite{xu2019relation};\\ NagE~\cite{yang2020nage}; ModulE~\cite{chai2022module};\\ KGLG~\cite{ebisu2019generalized}; DensE~\cite{lu2020dense};\\
            GrpKG~\cite{yang2021knowledge},fill=m5,draw=white,rounded corners=3pt,align=left,inner xsep=0pt,draw=black,text width=4cm]
        ]
        [Ring, part0,fill=m3,draw=black
            [M$\ddot{o}$biusE~\cite{chen2021mobiuse},fill=m5,draw=white,rounded corners=3pt,align=left,inner xsep=0pt,draw=black,text width=4cm]
        ]
        [Vector Space, part0,fill=m3,draw=black
        [Real Vector Space,part0,fill=m4,draw=black
        [TransE $\&$ its extensions~\cite{bordes2013translating,wang2014knowledge};\\ RESCAL~\cite{nickel2011three}; DisMult~\cite{yang2014embedding};\\ MQuatE~\cite{yu2021mquade}; StructurE~\cite{zhang2022structural};\\ TimE~\cite{zhang2021knowledge}; LineaRE~\cite{peng2020lineare};\\ ReflectE~\cite{zhang2022knowledge}; ExpressivE~\cite{pavlovic2022expressive}, fill=m5,draw=white,rounded corners=3pt,align=left,inner xsep=1.5mm,draw=black,text width=4.2cm
        ]
        ]
        [Complex Vector Space,part0,fill=m4,draw=black
    [ComplEx~\cite{trouillon2016complex}; RotatE~\cite{sun2019rotate};\\ QuatE~\cite{zhang2019quaternion}; QuatDE~\cite{gao2021quatde};\\ BiQUE~\cite{guo2021bique}; DualE~\cite{cao2021dual};\\DualQuatE~\cite{gao2021dual}; CORE~\cite{ge2022core};\\ HA-RotatE~\cite{wang2021hierarchical}; Rotate4D~\cite{le2023rotate4D},fill=m5,draw=white,rounded corners=3pt,align=left,inner xsep=1.5mm,draw=black,text width=4.2cm]
        ]
        [Other models in Vector Space,part0,fill=m4,draw=black
    [ConvE~\cite{dettmers2018convolutional}; R-GCN~\cite{schlichtkrull2018modeling};\\ KG-BERT~\cite{yao2019kg}; DKRL~\cite{xie2016DKRL},fill=m5,draw=white,rounded corners=3pt,align=left,inner xsep=1.5mm,draw=black,text width=4.2cm]
        ]
        ]
        ]
        [Analytical Structure,part0,fill=m2,draw=black
        [Probability Space,part0,fill=m3,draw=black
        [TransG~\cite{xiao2015transg}; KG2E~\cite{he2015learning}; DiriE~\cite{wang2022dirie}; GaussianPath~\cite{wan2021gaussianpath},fill=m5,draw=white,draw=black]
        ]
        [Euclidean Space,part0,fill=m3,draw=black
        [FieldE~\cite{nayyeri2021knowledge}; TANGO~\cite{han2021temporal},fill=m5,draw=white,draw=black]
        ]
        ]
        ]
    \end{forest}}
    \caption{The systematic categorisation of KGE models based on three mathematical perspectives.}
    \label{fig:tree_KGE}
\end{figure*}
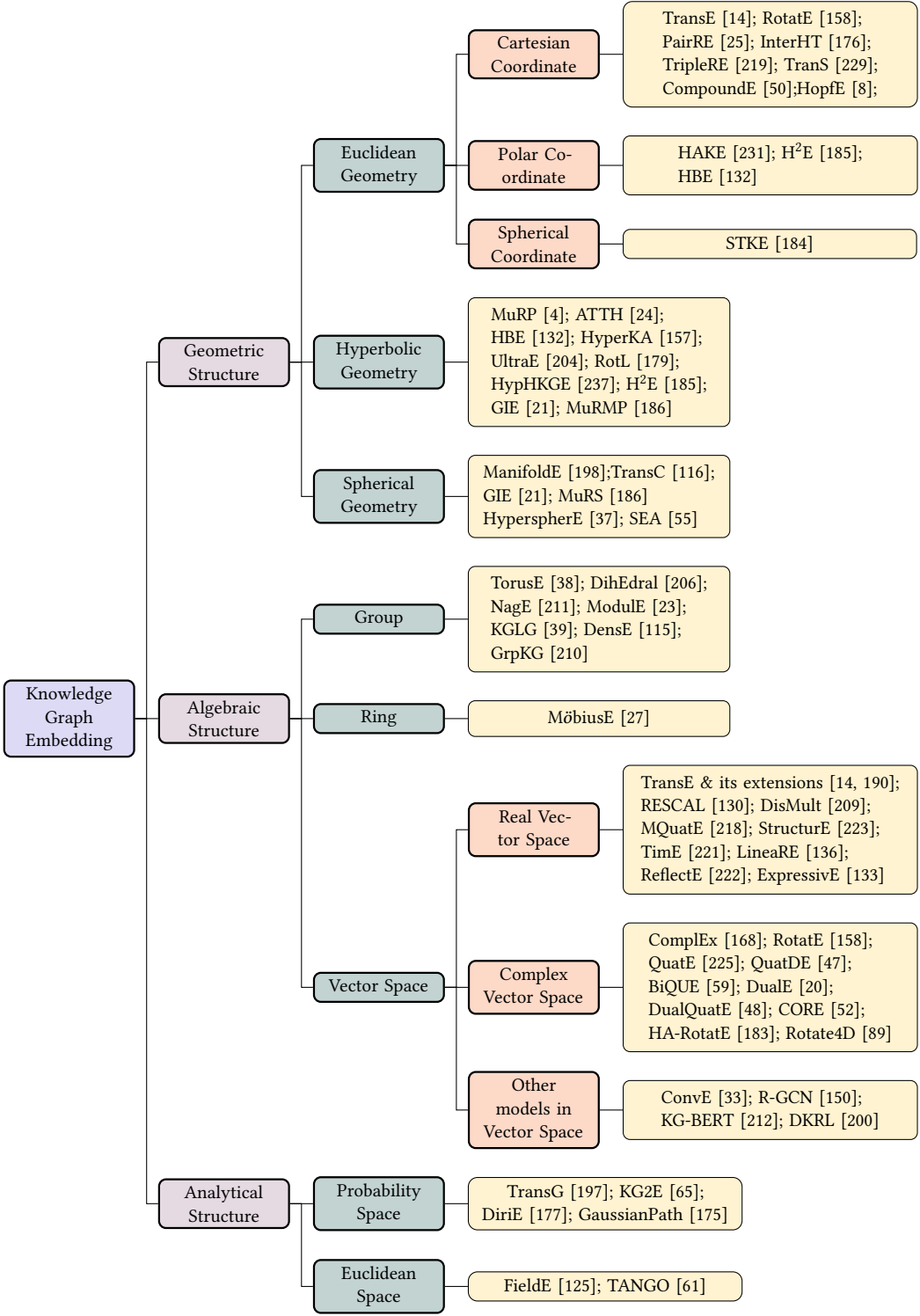

As a result, in order to better understand the influence of different mathematical representation spaces on KGE methods, we build a systematic, comprehensive, multi-angle categorisation to classify the special spaces and categorise KGE models more accurately based on three mathematical structures, namely  
\textbf{Algebraic Structure}, \textbf{Geometric Structure} and \textbf{Analytical Structure}.
Most KGE models fall under these three structures, which proves the rationality of our categorisation, as shown in Figure \ref{fig:tree_KGE}. It is worth noting that we will add additional definitions in Section~\ref{sec:spaces_in_KGE} for KGE spaces that are not mentioned in Section~\ref{sec:preliminaries}.

In this section, 
we will first describe the definitions and properties of the above three mathematical structures, after which we will provide some representative KGE methods that learn embedding in the corresponding mathematical structure, as well as summarise how spatial advantages work in KGE models.

\subsection{Algebraic Structure}
An algebraic structure is a nonempty set on which one or more finite operations satisfying the axiom are defined~\cite{greub2012linear,wilkinson2013linear}. Some 
representative algebraic structures include \textit{vector space, group} and \textit{ring}.
For example, the vector space $\mathbb{X}$ is an algebraic structure that 
involves many common \textit{binary operations}, such as addition, subtraction and multiplication. 
From the algebraic point of view, most models in machine learning and even knowledge graph embedding involve algebraic operations, such as m$\ddot{o}$buis embedding~\cite{balazevic2019multi,ungar2001hyperbolic,chen2021mobiuse}, group embedding~\cite{ebisu2018toruse,lu2020dense}, user and word embedding~\cite{liang2019collaborative,liang2019unsupervised,wu2022learning} etc., which all belong to the methods in the category of algebraic structure.

\subsubsection{Vector Space} 
~\\
The \textbf{Vector Space} is the most widely adopted mathematical space in the field of machine learning, whose definition is provided in Section~\ref{subsec:vector_space}. Since there are two  
convenient algebraic operations defined in vector space: \textit{vector addition} and \textit{scalar multiplication}, it is also widely used by many KGE methods. These methods utilise these operations to project both entities and relations into the same vector space, with the objective of preserving the relational interactions among entities in the vector representation space. As discussed in Section~\ref{subsec:vector_space}, the vector space can be classified as \textit{real vector space} and \textit{complex vector space} based on the scalar field. {In accordance with these classification criteria, we categorise vector space-based KGE methods into three distinct groups: real vector space-based, complex vector space-based, and other models within the vector space domain, such as neural network-based and external information-based models. In what follows, we delve into each of these groups of KGE methods in depth. }

\paragraph{Real Vector Space.}
One representative KGE methods based on real vector space is TransE~\cite{bordes2013translating}. Given a knowledge graph, it projects all entities and relations into a low-dimensional real vector space $\mathbb{R}^k$. 
Specifically, it directly represents the head entity $h$, the relation $r$ and the tail entity $t$ in a fact triple as embedding vectors $\mathbf{h}$, $\mathbf{r}$, $\mathbf{t}$ $\in \mathbb{R}^k$, respectively. 
If the triple $(h,r,t)$ holds, the embedding of the tail entity $\mathbf{t}$ should be as much as close to the head entity embedding $\mathbf{h}$ 
translated by the relation embedding $\mathbf{r}$, which conforms to the principal: $\mathbf{h} + \mathbf{r} \approx \mathbf{t}$. {It is very simple and effective to directly express the relations between entities by \textit{addition} operation.}
Therefore, the score function in TransE is defined as:
\begin{align}
s(\mathbf{h},\mathbf{r},\mathbf{t}) = - || \mathbf{h} + \mathbf{r} - \mathbf{t}||_{1/2}.    
\end{align}

However, despite TransE is simple and effective, it performs poorly when dealing with multi-relation (e.g., 1-to-$N$, $N$-to-1, $N$-to-$N$) data~\cite{toutanova2015observed}. 
For example, suppose that a relation $r_1$ is a $1$-to-$N$ relation, then for a head entity $h_1$, it may have relation $r_1$ with two different tail entities $t_1$ and $t_2$. TransE would enforce these two tail entities to have approximately the same embedding, i.e., $\mathbf{t}_1 \approx \mathbf{t}_2$, which is inaccurate since $t_1$ and $t_2$ are two different entities. The same analysis also applies to $N$-to-$1$ and $N$-to-$N$ relations.
To overcome the problems of TransE in modelling muti-relation data, TransH~\cite{wang2014knowledge} projects the head entity $h$ and tail entity $t$ into the hyperplane where the relationship $r$ resides. Specifically, TransH assumes that each relation embedding $\mathbf{r}$ lies in a different relation-specific hyperplane $\mathbf{w}_r$. In order to measure the plausibility that a triple $(h, r, t)$ holds, the head entity embedding $\mathbf{h}$ and tail entity embedding $\mathbf{t}$ are first projected into the relation-specific hyperplane $\mathbf{w}_r$:
\begin{align}
    \mathbf{h}_{\perp} = \mathbf{h} - \mathbf{w}_{r}^\top \mathbf{h} \mathbf{w}_{r}, 
    \mathbf{t}_{\perp} = \mathbf{t} - \mathbf{w}_{r}^\top \mathbf{t} \mathbf{w}_{r}, 
\end{align}
where $\mathbf{h}_{\perp}$ and $\mathbf{t}_{\perp}$ represent the projected head and tail entity embeddings, respectively. Therefore, the score function of a triple $(h, r,t)$ is defined as:
\begin{align}
    s(\mathbf{h},\mathbf{r},\mathbf{t}) = - ||\mathbf{h}_{\perp} + \mathbf{r} - \mathbf{t}_{\perp} ||_2^2 .
\end{align}

\begin{figure*}[!t]
    \begin{subfigure}[b]{0.37\textwidth}
        \centering
        \includegraphics[viewport= 250 140 580 450,clip,page=1,width=\linewidth]{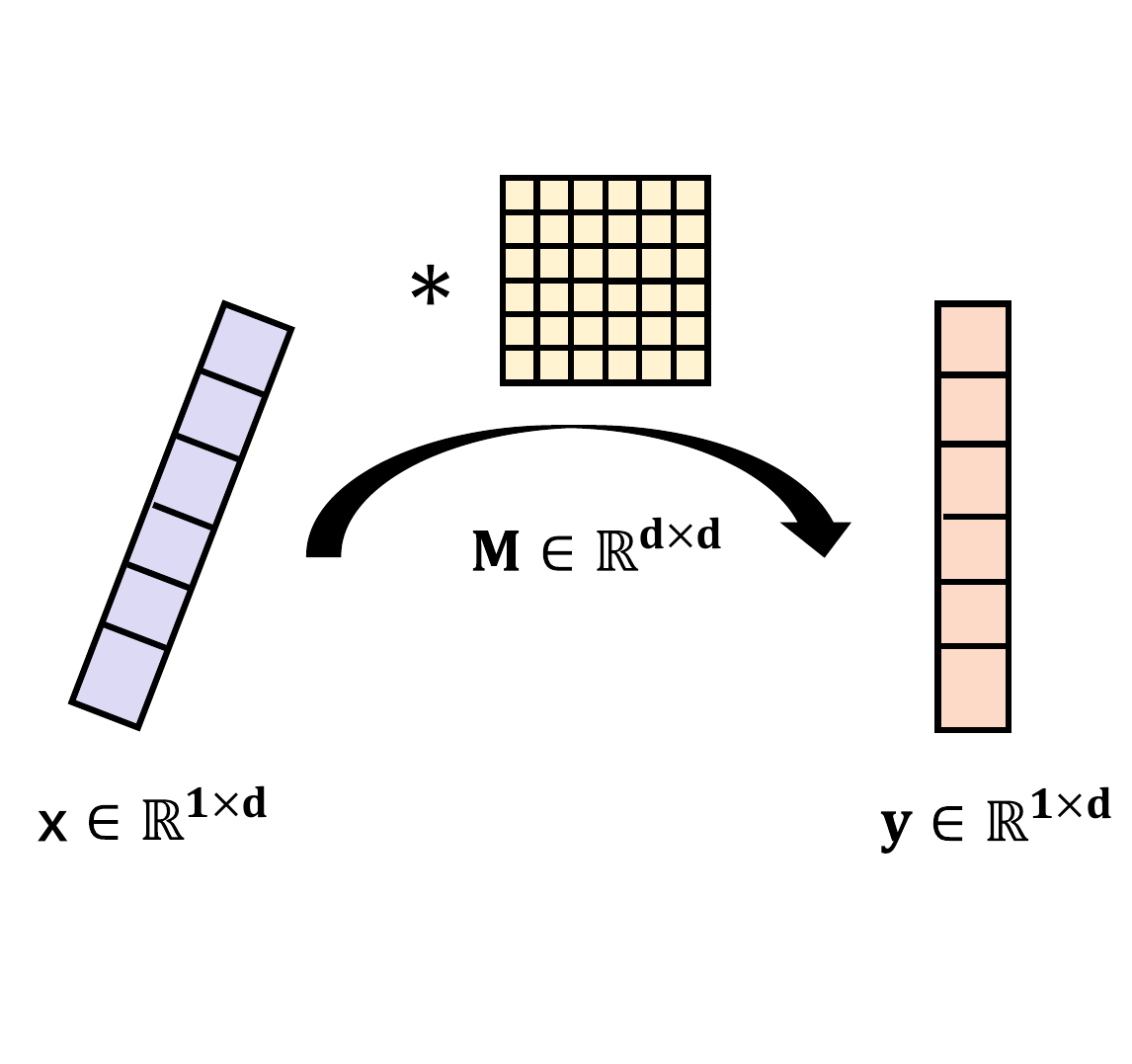}
        \caption{Tensor Projection in Real Vector Space.}
        \label{fig:tensor_project}
    \end{subfigure}
    \begin{subfigure}[b]{0.37\textwidth}
        \centering
        \includegraphics[viewport= 265 140 600 450,clip,page=2,width=\linewidth]{Images/alg1.pdf}
        \caption{Similarity Matching in Real Vector Space.}
        \label{fig:semantic_match}
    \end{subfigure}

    \begin{subfigure}[b]{0.37\textwidth}
        \includegraphics[viewport= 265 160 565 450,clip,page=3,width=1\linewidth]{Images/alg1.pdf}
        \caption{Angle Transformation in Complex Vector Space.}
        \label{fig:angle_trans}
    \end{subfigure}
    \begin{subfigure}[b]{0.37\textwidth}
        \includegraphics[viewport= 265 160 565 450,clip,page=5,width=1\linewidth]{Images/alg1.pdf}
        \caption{3-Dimensional Rotation in Quaternion.}
        \label{fig:quat_rota}
    \end{subfigure}
    \caption{An Illustration of some kind of algebraic operations which can be applied for knowledge graph embedding from algebraic perspective.}
    \label{fig:Illustration_algebric}
\end{figure*}

By projecting the entity embedding to relation-specific hyperplane, TransH allows entities to have different embeddings in different relations, which 
ensures that the embedding of $t$ is different even if the head entity or the relation is the same. {From TransH, we can see that \textit{projection} plays a key role, and projection is a very common operation in vector space that is used to establish various connections (See Figure \ref{fig:tensor_project})}.

Despite TransH can effectively handle multi-relations in KGs, it still assumes that the relation embeddings and entity embeddings belong to the same embedding space,
which will limit diversity. However, an entity may have multiple aspects and different relations may focus on different aspects of entities. To address this problem, TransR~\cite{lin2015learning} proposes to embed entities in an entity space $\mathbb{R}^k$ ($\mathbf{h}, \mathbf{t} \in \mathbb{R}^k$), and embed relations in a different relation space $\mathbb{R}^d$ ($\mathbf{r} \in \mathbb{R}^d$, $k$ and $d$ are not necessarily identical).  
The entity embedding and relation embedding are correlated by the relation-specific mapping matrix ($\mathbf{M}_r\in \mathbb{M}^{k \times d}$), which projects entity embedding from entity space to relation space. Therefore, the score function in TransR is defined as:
\begin{align}
    s(\mathbf{h},\mathbf{r},\mathbf{t}) = - ||\mathbf{M}_r \mathbf{h} + \mathbf{r} - \mathbf{M}_r \mathbf{t} ||_2^2 .
\end{align}

RESCAL~\cite{nickel2011three} models the semantic interactions between entities by using bilinear operations with score function:  
\begin{align}
    s(\mathbf{h},\mathbf{r},\mathbf{t}) =  \mathbf{h}^\top \mathbf{M}_r\mathbf{t}  =  \bigg(\sum_i\sum_j[\mathbf{M}_r]_{ij}\cdot\mathbf{h}_i\cdot\mathbf{t}_j \bigg), 
\end{align}
where each relation is defined as a matrix $\mathbf{M}_r\in \mathbb{M}^{d \times d}$, and ${[\mathbf{M}_r]}_{ij}$ denotes the i-th row and j-th column of the matrix $\mathbf{M}_r$. {Similar linear models such as DisMult~\cite{yang2014embedding}, HolE~\cite{nickel2016holographic}, ANALOGY~\cite{liu2017analogical}, SimplE~\cite{kazemi2018simple}, TuckER~\cite{balavzevic2019tucker} and LowFER~\cite{amin2020lowfer} have proven their great performance on downstream tasks.} {As a result, with simple and efficient linear operations, the plausibility of facts can be measured by matching latent semantics of entities and relations (See Figure \ref{fig:semantic_match}).}

Recently, there are some KGE models that learn embeddings in the real vector space as well.    
For example, ReflectE~\cite{zhang2022knowledge} uses reflection transformation to map properties and entities by Householder matrix. LineaRE~\cite{peng2020lineare} interprets a relation as a linear function of entities to capture connectivity patterns. TimE~\cite{zhang2021knowledge} projects entities into the nonlinear time domain to obtain better diversity distribution. It could be found that many mathematical operations of real vector space are worth exploring and applying to KGE.

\paragraph{Complex Vector Space.}
Compared with real vector embedding, complex vector embedding (or complex embedding for brevity) can handle a large variety of binary relations~\cite{trouillon2016complex}, such as symmetric and antisymmetric. 
Complex embedding methods for KGs, which embed entities and relations in the \textbf{Complex Vector Space} (i.e., $\mathbf{h}, \mathbf{r}, \mathbf{t} \in \mathbb{C}^k$), have also been widely studied.
ComplEx~\cite{trouillon2016complex} is the first model to use complex embedding in KGE. 
Specifically, it defines a score function with the Hermitian dot product in complex vector space:
\begin{align}
    s(\mathbf{h},\mathbf{r},\mathbf{t}) = Re \big( \mathbf{h}^\top diag(\mathbf{r})\bar{\mathbf{t}} \big) = Re \bigg(\sum_i\mathbf{r}_i\cdot\mathbf{h}_i\cdot\mathbf{t}_j \bigg), 
\end{align}
where $\bar{\mathbf{t}}$ is the conjugate (see Table \ref{tab:symbol_tem}) of $\mathbf{t}$, $Re(\cdot)$ 
denotes the operation to obtain the real part of a complex number. 
Since the score function is not symmetric anymore, the facts
with antisymmetric relations can receive different scores depending on the ordering of entities. Thus ComplEx can effectively capture antisymmetric relations while retaining the efficiency benefits of the dot product. 

Motivated by the Euler's rule $e^{i\theta} = cos\theta + sin\theta$, RotatE~\cite{sun2019rotate} maps entities and relations into the complex vector space and defines each relation as a rotation from the source entity to the target entity with the principal $\mathbf{t} = \mathbf{h} \circ \mathbf{r}$ (where $\circ$ denotes the Hadamard product, i.e., element-wise product). Different entities can be directly modelled through the angular transformation (See Figure ~\ref{fig:angle_trans}) so as to capture some patterns including symmetry, 
antisymmetry, inversion, and composition. 
Compared with ComplEx, QuatE~\cite{zhang2019quaternion} takes advantage of quaternion representations to enable richer and more expressive semantic matching between head and tail entities with the use of Hamilton product ($\otimes$, i.e., quaternion multiplication), where a quaternion $Q$ is defined as $Q = a+b\textbf{i}+c\textbf{j}+d\textbf{k}$ and (\textbf{i, j, k}) are imaginary units satisfying Hamilton's rule: $\textbf{i}^2 = \textbf{j}^2 = \textbf{k}^2 = \textbf{ijk} = -1$.
BiQUE~\cite{guo2021bique} extends the quaternion system to a more powerful algebraic system called biquaternion $q = (w_r+w_i\textbf{I})+(x_r+x_i\textbf{I})+(y_r+y_i\textbf{I})+(z_r+z_i\textbf{I})$, where $w_r,x_r,y_r,z_r,w_i,x_i,y_i,z_i\in \mathbb{R}$. By utilising the Hamilton product of biquaternions, BiQUE imbues itself with a strong geometric interpretation (i.e., the Euclidean/hyperbolic rotation). {In addition to the fact that quaternions have more degree of freedom in the four dimensional space. It is worth mentioning that the interpolation between two quaternions is extremely easy, which helps to establish rotations (See Figure ~\ref{fig:quat_rota}).}
Recently, DualE~\cite{cao2021dual} employs a novel framework which can embrace both translation and rotation operations in the dual quaternion space($Q_{dual} = a + \epsilon b$, where $a$ and $b$ are quaternions while $\epsilon$ is a dual unit). DualQuatE~\cite{gao2021dual} combines the idea of DualE and QuatE. HA-RotatE~\cite{wang2021hierarchical} and CORE~\cite{ge2022core} inherit the structure of RotatE and expand the ability of KGs embedding. 

\paragraph{{Neural Network Models in Vector Space.}}
{Neural networks~\cite{dettmers2018convolutional} are employed in KGE models for learning embedded features. However, the black-box nature of neural networks precludes our understanding of the specific mathematical features captured by these neural network-based KGE models. Nevertheless, given that these models yield vector representations, we categorise them in conventional vector spaces.}

{ConvE~\cite{dettmers2018convolutional} initially reshapes the head entity and relation into a 2D matrix, subsequently utilising 2D convolution in conjunction with multiple layers to represent interactions between entities and relations. The scoring function is defined as: 
\begin{align}
    s(\mathbf{h},\mathbf{r},\mathbf{t}) = \sigma \big( \text{vec} \big( \sigma\big ( [\mathbf{h}_{2D},\mathbf{r}_{2D} ]\ast \omega \big) \big) \mathbf{W} \big)\mathbf{t} ,
\end{align}
where $\sigma$ denotes activation function, and $\text{vec}(\cdot)$ means the vectorisation operation. $\mathbf{h}_{2D}$ and $\mathbf{r}_{2D}$ are 2D reshaping of $\mathbf{h}$ and $\mathbf{r}$ (i.e., $\mathbf{h}_{2D}, \mathbf{r}_{2D} \in \mathbb{R}^{k_w \times k_h}$, if $\mathbf{h}, \mathbf{r}\in \mathbb{R}^k \text{ with } k=k_wk_h$), respectively. $\omega$ is the convolutional filter. 
Owing to the potent feature extraction capacity of the nonlinear neural network layer, ConvE exhibits high expressiveness and delivers exceptional performance.}

{R-GCN~\cite{schlichtkrull2018modeling} represents relationships between entities and relations by employing Graph Convolutional Networks (GCNs~\cite{kipf2016semi,yao2023improving}), which focus on local graph neighbourhoods to manage large-scale relational data. To compute the forward-pass update for an entity, the hidden state $\mathbf{y}$ of layer $l+1$ undergoes a propagation process~\cite{yao2023improving}: 
\begin{align}
    \mathbf{y}_i^{l+1} = \sigma \bigg( \sum_{r\in \mathcal{R}}\sum_{j\in \mathcal{N}_i^r} \frac{1}{c_{i,r}} \mathbf{W}_r^l y_j^l + \mathbf{W}_0^l y_i^l \bigg),
\end{align}
where $\mathcal{N}_i^r$ denotes the set of neighbor indices of node $i$ under relation $\mathbf{r}\in \mathcal{R}$ and $c_{i,r}$ is a normalisation constant. CompGCN~\cite{vashishth2019composition} leverages a variety of composition operators derived from KGE methods to concurrently embed nodes and relations in a graph. This approach has demonstrated the capability to generalise various existing multi-relational GCN techniques, including R-GCN~\cite{schlichtkrull2018modeling}, Directed-GCN~\cite{marcheggiani2017encoding} and Weighted-GCN~\cite{shang2019end}.}

{KG-BERT~\cite{yao2019kg}, a transformer-based~\cite{vaswani2017attention} model, interprets triples as text sequences and conducts knowledge embedding by leveraging a pre-trained language model BERT~\cite{devlin2018bert}. KG-BERT is capable of utilising abundant linguistic information present in the extensive text and emphasising the most pertinent words associated with a triple. 
Recently, Knowformer~\cite{li2023position} employs position-aware relational compositions to encode the semantics of entities appearing in varying positions within a relational triple and achieves state-of-the-art. It has been proved that these compositions assist the self-attention mechanism~\cite{vaswani2017attention} in differentiating entity roles based on their positions.}

\paragraph{{Incorporate Auxiliary Information in KGE.}}
{
Learning knowledge graph embeddings using auxiliary information constitutes a significant subfield within KGE approaches.
While this topic 
may not be as closely connected to the mathematical representation space, we will provide a concise overview of some classic external-information-based KGE models to maintain a comprehensive review.
Commonly used auxiliary information in existing work includes text descriptions, entity types, relational structures or paths, and other information~\cite{wang2015knowledge,tabacof2019probability,safavi2020evaluating,feng2016gake}. DKRL~\cite{xie2016DKRL} investigates more profound knowledge representation by utilising CNNs to extract the semantics of entity descriptions in a representation learning manner. TKRL~\cite{xie2016TKRL} considers hierarchical entity types as projection matrices and employs the type information as relation-specific type constraints. It has been demonstrated that TKRL effectively captures hierarchical type information, which is crucial for constructing representations of knowledge graphs. HRS~\cite{zhang2018knowledge} treats relations in knowledge graphs as a three-layer hierarchical relation structure, which can be effortlessly integrated into other KGE models to acquire abundant structural information. RSN~\cite{guo2019learning} and Interstellar~\cite{zhang2020interstellar} tackle the issue of previous models' insufficient ability to capture long-term relational dependencies of entities effectively by employing relational paths. Recently, TransO~\cite{li2023transo} has been proposed as a method to seamlessly integrate all available ontology information (i.e., type information, relation constraint information, and hierarchical structure information.) within the knowledge embedding process, thereby enhancing decision-making capabilities in complex situations. In addition, combined with other external information such as image data~\cite{xie2017image}, conceptual information~\cite{hao2019universal,guan2019knowledge,huang2023concept2box} also holds significant importance in the realm of knowledge graph embedding.}

\subsubsection{Group}
~\\
A \textbf{\textit{group}}~\cite{bourbaki1989commutative,lang2012algebra,rotman2000first} is an algebraic structure composed of a set and an operation, which is ubiquitous in all fields inside and outside mathematics. For example, symmetry groups describe the symmetries of geometry~\cite{robinson1938representations}, and lie groups are used in particle physics~\cite{georgi1974unity}. Because of its unique abstract algebra properties, groups are also widely used in machine learning. In this section, we will start with the definition of groups and then describe the KGE models 
that leverage groups as embedding space. 

\begin{definition}
A (binary) \textbf{\textit{operation}} on a set $\mathbb{G}$ is a function:
\begin{equation*}
    \ast : \mathbb{G} \times \mathbb{G} \rightarrow \mathbb{G}.
\end{equation*}
\end{definition}

\begin{definition}
A \textbf{\textit{group}} is a set $\mathbb{G}$ equipped with an operation $\ast$ and a special element $e\in \mathbb{G}$, called the \textbf{\textit{identity}}, such that: 
\begin{itemize}
    \item[(a)] (Associativity): For every $a,b,c\in \mathbb{G}$, $(a\ast b) \ast c= a\ast (b\ast c)$;
    \item[(b)] (Existence of identity): $e\ast a = a$ for all $a\in \mathbb{G}$;
    \item[(c)] (Existence of inverses): for every $a\in \mathbb{G}$, there is $a'\in \mathbb{G}$ with $a'\ast a = e$.
\end{itemize}
\end{definition}

To tackle TransE’s regularisation problem, TorusE~\cite{ebisu2018toruse} 
proposes to embed KGEs into a special algebraic structure---Torus. A torus is an Abelian Lie group, which is derived from the vector space through the nature projection $\pi : \mathbb{R}^n \rightarrow T^n, x\mapsto\left [ x \right ]$ ($T^n$ denotes quotient space~\cite{ebisu2018toruse}). With the help of torus, the model never diverges unlimitedly. 

DihEdral~\cite{xu2019relation} is the first attempt to employ finite non-Abelian group in KG embedding to account for relation compositions. 
Since the elements in a dihedral group are well constructed by rotation and reflection operations, and the multiplication between elements can be Abelian or non-Abelian, DihEdral is capable to capture all desired properties: (skew-)symmetry, inversion and (non-) Abelian composition. DensE~\cite{lu2020dense} decomposes a relation operator into a SO(3) group-based (SO(3): Special Orthogonal Group in 3 dimensions) rotation as well as a scaling transformation. NagE~\cite{yang2020nage} proves for the first time that the group algebraic structure is significant for designing relational embedding models. \textit{Specifically, the definition of group can naturally satisfy the basic properties (e.g., inversion, composition) of knowledge graphs}, which means the  group-based models should have 
great potential to deal with KGE tasks. Other recent models based on group structure such as ModulE ~\cite{chai2022module} consider both entity and relation as group elements so as to achieve state-of-the-art performance. 

\subsubsection{Ring}
~\\
In mathematics, \textbf{\textit{rings}}~\cite{bourbaki1989commutative,lang2012algebra,rotman2000first} are algebraic structures that generalised fields. Commutative rings are one of the main branches of ring theory. Examples include the set of integers with addition and multiplication, and the set of polynomials with the same operations. 
Ring theory was later proved useful in geometry and analysis ~\cite{faith2012algebra}. 
In this section, we will first introduce the definition of ring, and then discuss the KGE models that use ring as embedding space.

\begin{definition}
A \textbf{\textit{ring}} is a set $\mathbb{S}$ equipped with two binary operations: $+$ (\textit{addition}) and $\cdot$ (\textit{multiplication}) which satisfy the following axioms,
\begin{itemize}
    \item[(a)]$\mathbb{S}$ is an Abelian group under addition, meaning that: 
          \begin{itemize}
              \item[$\bullet$] (Associative): $(a+b)+c = a+(b+c)$ for all $a,b,c\in R$;
              \item[$\bullet$] (Commutative): $a+b = b+a$ for all $a,b\in \mathbb{S}$;
              \item[$\bullet$] (Additive identity): There is an element $0$ in $\mathbb{S}$ such that $a+0=a$ for all $a\in \mathbb{S}$;
              \item[$\bullet$] (Additive inverse): For each $a$ in $\mathbb{S}$ there exists $-a$ such that $a+(-a) = 0$.
          \end{itemize}
    \item[(b)] $\mathbb{S}$ is a monoid under multiplication, meaning that:
          \begin{itemize}
              \item[$\bullet$] (Associative): $(a\cdot b)\cdot c = a\cdot(b\cdot c)$ for all $a,b,c\in \mathbb{S}$;
              \item[$\bullet$] (Multiplicative identity): There is an element 1 in $\mathbb{S}$ such that $a\cdot 1 = a $ and $1\cdot a = a$ for all $a\in \mathbb{S}$.
          \end{itemize}
    \item[(c)] Multiplication is distributive with respect to addition, meaning that:
          \begin{itemize}
              \item[$\bullet$] (Left distributivity): $a\cdot (b+c) = (a\cdot b)+(a\cdot c)$ for all $a,b,c\in \mathbb{S}$;
              \item[$\bullet$] (Right distributivity): $(b+c)\cdot a = (b\cdot a)+(c\cdot a)$ for all $a,b,c\in \mathbb{S}$.
          \end{itemize}
\end{itemize}
\end{definition}

M$\ddot{o}$biusE~\cite{chen2021mobiuse} extends KGE to manifold-based embedding, in which the entities and relations are embedded to the surface of M$\ddot{o}$bius ring. With the scoring function ($\textit{dist} (\textbf{h}\oplus\textbf{r},\textbf{t})$, where $\oplus$ and \textit{dist} represent addition and distance function specially defined on M$\ddot{o}$bis ring, respectively), M$\ddot{o}$biusE has much more expressiveness and flexibility than TorusE~\cite{ebisu2018toruse} due to the extra properties on ring. As M$\ddot{o}$biusE subsumes TorusE, it naturally inherits all the desired properties of TorusE including symmetric/antisymmetry, inversion, and composition. \textit{It is worth mentioning that the M$\ddot{o}$bius band is a non-oriented surface and the concept of clockwise and counterclockwise cannot be clearly defined, which may be helpful for certain tasks that are closely related to orientation.}

\subsection{Geometric Structure}
A geometric structure~\cite{goldman1988varieties} on a manifold is a complete Riemannian metric which is locally homogeneous (i.e., any two points have isometric neighbourhoods). Although there are few detailed definitions of geometric structures, here we 
focus more on the knowledge graph embedding models built on different geometric models/spaces and analyse them in detail from three geometric perspectives: Euclidean Geometry, Hyperbolic Geometry and Spherical Geometry. We summarise the operations of these three geometries in Table \ref{table:summaryofgeometry}.

\begin{table*}[!t]
    \centering
    \footnotesize
    \renewcommand{\arraystretch}{1}
    \setlength\tabcolsep{6pt}
    \caption{Summary of operations in Euclidean, Hyperboloid and Spherical models~\cite{nayyeri2021knowledge,lou2020differentiating,weber2018curvature}. {It {is worth noting} that the category of geometric spaces associated with a model is {closely related} to the distance function utilised in its scoring function. Models based on Euclidean geometry typically employ the well-known Euclidean distances (i.e., $||\cdot||$), where varying transformation relations for $\mathbf{h}$, $\mathbf{r}$, and $\mathbf{t}$ give rise to distinct models. Conversely, hyperbolic and spherical models tend to use hyperbolic distance ($d_{\mathbb{H}}$) and spherical distance ($d_{\mathbb{S}}$) with unique properties for their respective scoring functions. Notably, some models employing spherical embedding techniques may not exclusively rely on spherical distances. Instead, they may model entities as spheres or through other spherical geometric embeddings. Visual representations of these three geometric embeddings are included in the table.}}
    \resizebox*{!}{0.4\columnwidth}{
    \begin{tabular}{llll}
        \toprule
                                     & \textbf{Euclidean}                            & \textbf{Hyperboloid}                                                                                                                                              & \textbf{Spherical}                                                                               \\
        \midrule
        Manifold $\mathcal{M} $      & $\mathbb{R}^n$                                & $\mathbb{H}^n_K = \left \{ x\in \mathbb{R}^{n+1}:\left \langle x,x \right \rangle = \frac{1}{K} \right \}$                                                        & $\mathbb{S}^n_K = \left \{ x\in \mathbb{R}^{n+1}:\left \langle x,x \right \rangle = 1 \right \}$ \\ \midrule
        Distance $d(x,y)$            & $\left \langle \sqrt{x-y,x-y} \right \rangle$ & $\frac{1}{\left | K \right |}cosh^{-1}(K\left \langle x,y \right \rangle)$                                                                                        & $cos^{-1}(\left \langle x,y \right \rangle)$                                                     \\ \midrule
        Exponential map $exp^K_x(v)$ & x+v                                           & $cosh(\sqrt{\left | K \right |}\left \| v \right \|)x+v\frac{sinh(\sqrt{\left | K \right |}\left \| v \right \|)}{\sqrt{\left | K \right |}\left \| v \right \|}$ & $cos(\left \| v \right \|)x+sin(\left \| v \right \|)\frac{v}{\left \| v \right \|}$             \\ \midrule
        Curvature $\mathfrak{C} $    & 0                                             & $<0    $                                                                                                                                                          & $>0 $                                                                                            \\ \midrule
        Sum of angles $\mathfrak{A}$ & $\pi$                                         & $<\pi$                                                                                                                                                            & $>\pi$                                                                                      \\ \midrule
        \multirow{3}{*}{\shortstack{Scoring function\\$s(\mathbf{h},\mathbf{r},\mathbf{t})$}} & 
TransE: $-|| \mathbf{h} + \mathbf{r} - \mathbf{t}||$ 
& MuRP: $-d_{\mathbb{H}}(\mathbf{x}_{\mathbf{h}}^{(\mathbf{r})},\mathbf{x}_{\mathbf{t}}^{(\mathbf{r})})^2+b_{\mathbf{h}}+b_{\mathbf{t}}$ 
& MainfoldE: $-|| \varphi(\mathbf{h})+\varphi(\mathbf{r}) - \varphi(\mathbf{t})||^2$
         \\ 
         & RotatE: $-|| \mathbf{h} \circ \mathbf{r} - \mathbf{t}||$ &
          ATTH: $-d_{\mathbb{H}}(\mathbf{Q}(\mathbf{h},\mathbf{r}),\mathbf{t}^{\mathbf{H}})^2+b_{\mathbf{h}}+b_{\mathbf{t}}$
        & MuRS: $-d_{\mathbb{S}}(\mathbf{x}_{\mathbf{h}}^{(\mathbf{r})},\mathbf{x}_{\mathbf{t}}^{(\mathbf{r})})^2+b_{\mathbf{h}}+b_{\mathbf{t}}$
          \\
        &PairRE: $-|| \mathbf{h} \circ \mathbf{r}^{H} -\mathbf{t}\circ \mathbf{r}^{T}||$
         &HyperKA: $-d_{\mathbb{H}}(\mathbf{x}_{\mathbf{h}}^{(0)}\oplus \mathbf{x}_{\mathbf{r}}^{(0)}, \mathbf{x}_{\mathbf{t}}^{(0)})$
         &SEA: $-|| \mathbf{e}-\mathbf{q}_i||+b_{\mathbf{h}}+b_{\mathbf{e}}$
         \\ \midrule
        ~Illustration                & \begin{minipage}[c]{0.2\columnwidth}
            \centering
            \raisebox{-.5\height}{\includegraphics[viewport= 200 150 700 450,clip,page=1,width=\linewidth]{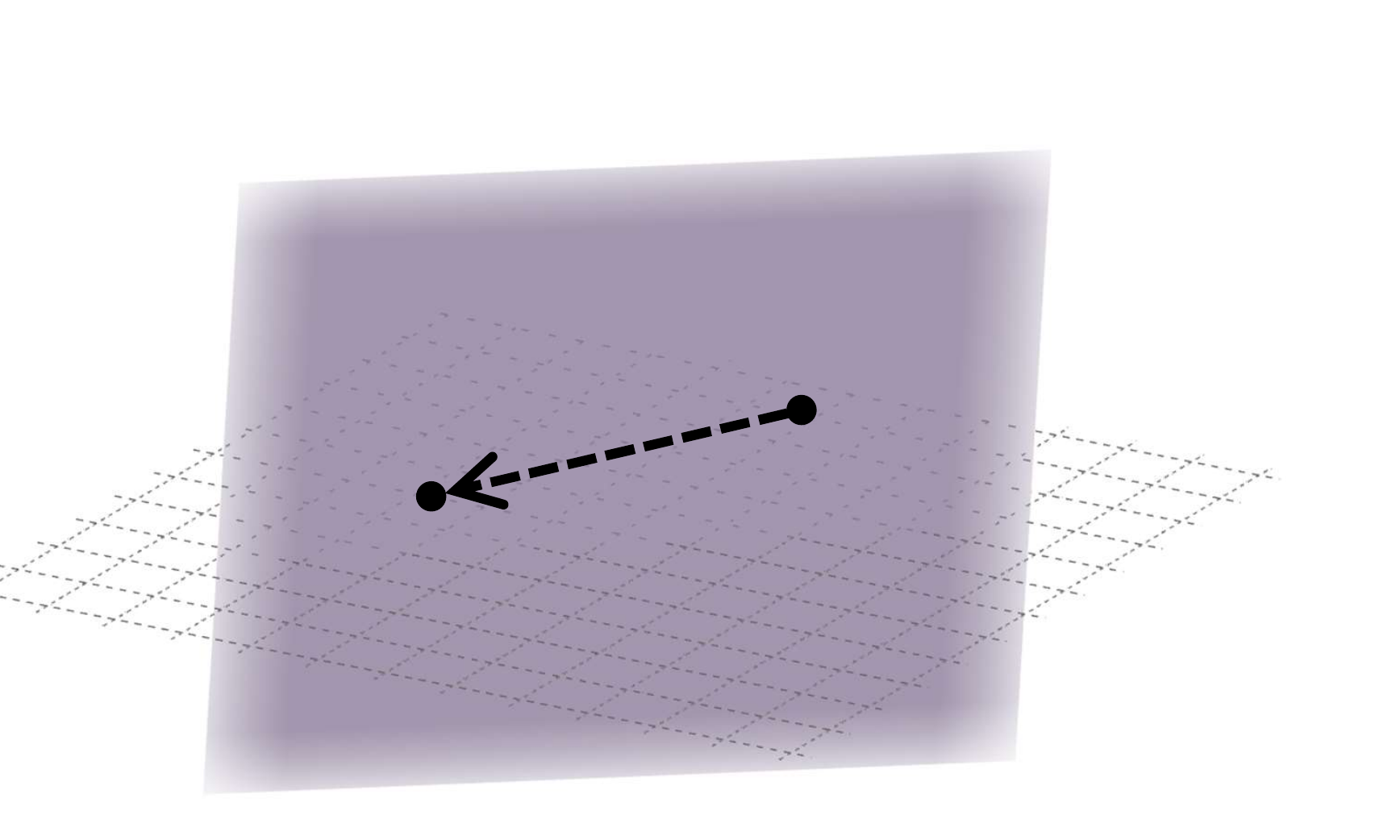}}
        \end{minipage}                    & \begin{minipage}[c]{0.2\columnwidth}
            \centering
            \raisebox{-.5\height}{\includegraphics[viewport= 150 150 650 450,clip,page=2,width=\linewidth]{Images/eu_hy_sp_purple.pdf}}
        \end{minipage}                                                                                                                                        & \begin{minipage}[c]{0.2\columnwidth}
            \centering
            \raisebox{-.5\height}{\includegraphics[viewport= 200 150 700 450,clip,page=3,width=\linewidth]{Images/eu_hy_sp_purple.pdf}}
        \end{minipage}                                                                       \\
        \bottomrule
    \end{tabular}}
    \label{table:summaryofgeometry}

\end{table*}

\subsubsection{Euclidean Geometry}
~\\
Euclidean geometry is the study of geometrical shapes (plane and solid) and figures based on different axioms and theorems. It is basically introduced for flat surfaces or plane surfaces. This part of geometry was employed by the Greek mathematician \textit{Euclid}, who has also described it in his book, \textit{Elements}~\cite{fitzpatrick2007euclid}. Geometry is derived from the Greek words `geo' which means earth and `metrein' which means `to measure'. 

Euclidean geometry deals with things like points, lines, angles, squares, triangles, and other shapes. Based on different coordinate systems, in this part, we will divide KGE models built in Euclidean geometry into four part: \textit{Cartesian Coordinate, Polar Coordinate} and \textit{Spherical Coordinate}. It is worth noting that: Euclidean geometry under geometric structure and vector space under algebraic structure are very similar and have some 
common concepts. 
This is because geometric features and algebraic features are usually closely related. But in this section we will emphasise the advantages of KGE models from the geometric perspective.

\paragraph{Cartesian Coordinate.}
Most of the translation-based models are based on the common Cartesian coordinates. For example, TransE follows the principle: $\mathbf{h} + \mathbf{r} \approx \mathbf{t}$, i.e., the vector of head entity add the relation vector is equal to the tail vector through a translation. These three vectors are connected from head to tail, forming a closed path in Cartesian coordinates.
RotatE shows that it's ingenious to define each relation as a rotation from the source entity to the target entity. For example, two relations $\mathbf{r}_1$ and $\mathbf{r}_2$ are inverse if and only if their embeddings are conjugate: $\mathbf{r}_1 = \bar{\mathbf{r}}_2$ ($\mathbf{r}_1:(cos(\theta_1),sin(\theta_1)),\mathbf{r}_2:(cos(\theta_2),sin(\theta_2))$), which means they are symmetric about the real axis (i.e., $sin(\theta_1) = -sin(\theta_2)$). All relation patterns can be clearly illustrated by geometric transformation in Cartesian coordinates. 
Moreover, QuatE~\cite{zhang2019quaternion} extends complex space to quaternion where there are two planes of rotations.
{Inspired by RotatE, Tang et al.~\cite{tang2020orthogonal} extend RotatE from 2D complex domain to high dimensional space with orthogonal transformations, which preserves the ability of modelling different patterns while achieving better performance. Furthermore, numerous RotatE-based models~\cite{wang2021hierarchical,huang2021knowledge,gao2020rotate3d,le2023knowledge,le2023rotate4D} exist, which expand upon the foundational principles of the original RotatE framework. For example, Rotate3D~\cite{gao2020rotate3d} and Rotate4D~\cite{le2023rotate4D} define rotational relations in higher dimensional space.}
However, it remains challenging for KGE models to handle complex relations. To mitigate this problem, PairRE~\cite{chao2021pairre} uses paired vectors for each relation $\mathbf{r} = \left [ \mathbf{r}^H,\mathbf{r}^T \right ]$, where the value of $\mathbf{r}^H$ and $\mathbf{r}^T$ can be changed 
to fit the complex relations. 
Through the scoring function, relation vectors can project entities to arbitrary positions inside a unit circle lying on the Cartesian coordinates. 
Further analysis also proves that PairRE can capture subrelation in addition to those that RotatE can model~\cite{sun2019rotate}. 

\begin{figure*}[!t]
\begin{subfigure}[b]{0.33\textwidth}
    \centering
    \includegraphics[viewport= 250 150 600 450,clip,page=1,width=\linewidth]{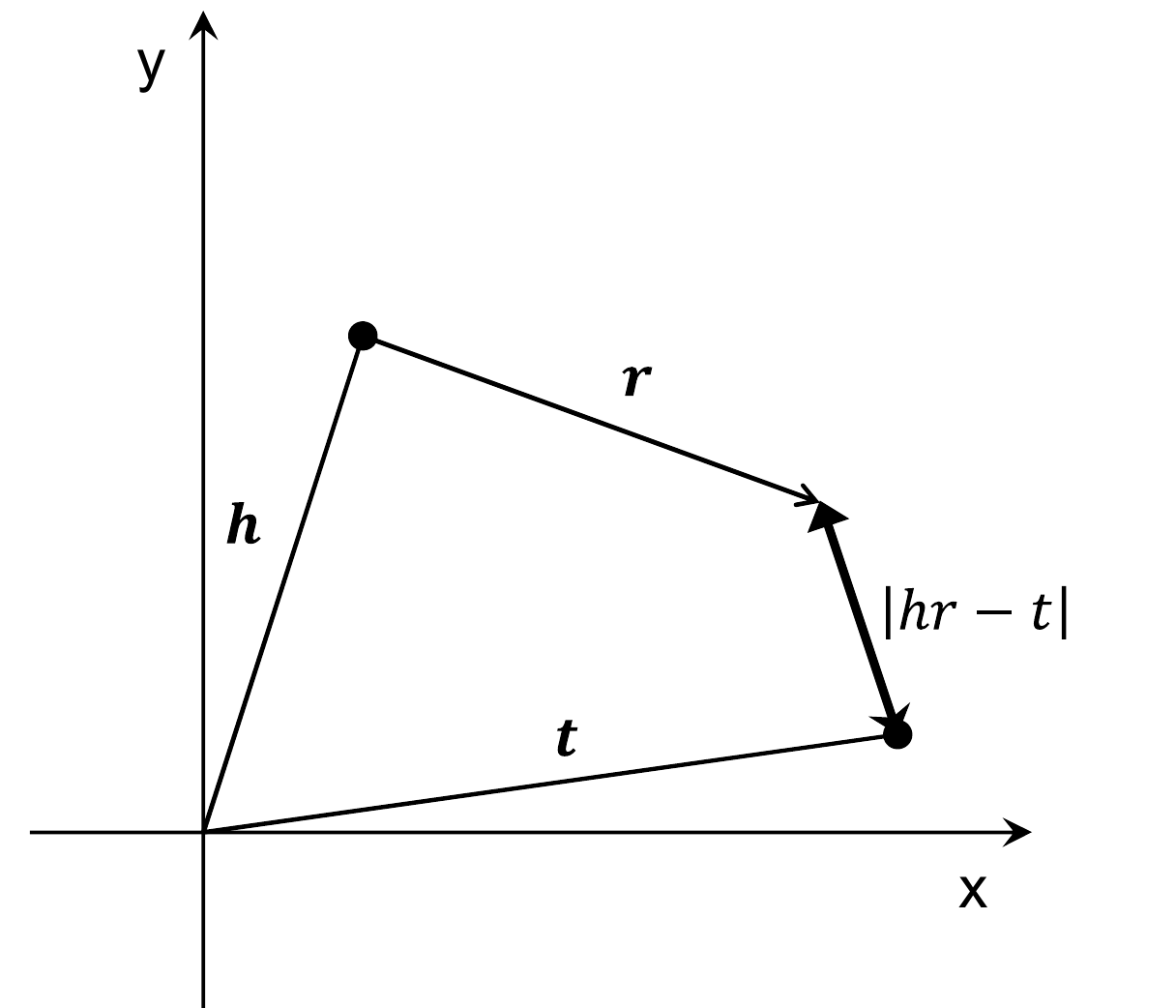}
    \caption{TransE}
    \label{fig:car_transe}
\end{subfigure}
\begin{subfigure}[b]{0.33\textwidth}
    \centering
    \includegraphics[viewport= 280 200 650 450,clip,page=2,width=1.2\linewidth,scale=1.5]{Images/cartesian6_final.pdf}
    \caption{RotatE}
    \label{fig:car_rotate}
\end{subfigure}    
\begin{subfigure}[b]{0.3\textwidth}
    \centering
    \includegraphics[viewport= 250 150 600 450,clip,page=3,width=\linewidth]{Images/cartesian6_final.pdf}
    \caption{QuatE}
    \label{fig:car_quate}
\end{subfigure}    

\begin{subfigure}[b]{0.32\textwidth}
    \centering
    \includegraphics[viewport= 250 165 600 450,clip,page=4,width=\linewidth]{Images/cartesian6_final.pdf}
    \caption{PairRE}
    \label{fig:car_pairre}
\end{subfigure}
\begin{subfigure}[b]{0.32\textwidth}
    \centering
    \includegraphics[viewport= 250 165 600 450,clip,page=5,width=\linewidth]{Images/cartesian6_final.pdf}
    \caption{CompoundE}
    \label{fig:car_compounde}
\end{subfigure}    
\begin{subfigure}[b]{0.32\textwidth}
    \centering
    \includegraphics[viewport= 250 165 600 450,clip,page=6,width=\linewidth]{Images/cartesian6_final.pdf}
    \caption{TranS}
    \label{fig:car_trans}
\end{subfigure}    
\caption{KGE models in Cartesian Coordinate. ``$\leftrightarrow$'' illustrates the distance (scoring) function.}
\label{fig:Illustration_in_car}
\end{figure*}

Other Cartesian coordinate-based KGE models have also emerged recently. For example, TripleRE \cite{yu2022triplere} inherits PairRE’s projection part and builds the translation part by its own way. InterHT~\cite{wang2022interht} merges the information from tail entity to head entity representation. TranS~\cite{zhang2022trans} is proposed to efficiently capture single relations {and HousE~\cite{li2022house} utilise Householder parameterisation~\cite{householder1958unitary} to capture crucial relation patterns.} CompoundE~\cite{ge2022compounde} successfully embeds KGs by leveraging three fundamental Euclidean geometric operations. {CompoundE3D~\cite{ge2023knowledge} updates the compound transformation to further match the rich underlying characteristics of a KG. ConE~\cite{ma2023conte} combines an explicit relation and a latent relation as collaborative relation for solving circular relation problems.} The above geometric models based on Cartesian coordinates are illustrated in Figure~\ref{fig:Illustration_in_car}. 
{\textit{In summary, the prevalent geometric transformations employed in KGE include translation, rotation, reflection, and scaling. These transformations have been proven effective in capturing essential relational patterns and mapping characteristics.}}

\paragraph{Polar Coordinate.}
In order to naturally reflect the hierarchy of KGs, HAKE~\cite{zhang2020learning} models entities and relations in the polar coordinate system. The radial coordinate aims to model entities ($\mathbf{h}_m,\mathbf{t}_m,\mathbf{r}_m \in \mathbb{R}^k$) at different levels, and the angular coordinate aims to distinguish entities ($\mathbf{h}_p,\mathbf{t}_p,\mathbf{r}_p \in \left [ 0,2\pi \right )$ ) at the same level of hierarchy. Combining the modules and phase information, HAKE significantly outperforms the SOTA hierarchy-based models while also completely capturing relational patterns.
Although H$^2$E~\cite{wang2021knowledge}'s representation space belongs to hyperbolic geometric, it also uses modulus and phase information to embed entities--- which is named Hyperbolic Polar Embedding. Modulus embedding models the inter-level hierarchy and Phase Embedding models the intra-level hierarchy.
Recently, HBE~\cite{pan2021hyperbolic} is also a hyperbolic geometry-based KGE model but with polar coordinate system by Mobius multiplication and Mobius addition in an extended Poincare ball.

\paragraph{Spherical Coordinate.}
{
STKE~\cite{wang2022stke} represents entities and relations in a spherical coordinate system, but for temporal knowledge graphs. Each entity contains the radial part $r$, the azimuth part $\theta$ and the polar part $\varphi$. With the help of spherical coordinates, STKE treats temporal changes as scaling and rotation of entity embeddings, which can dynamically distinguish different time-constrained entities.}

\subsubsection{Hyperbolic Geometry}
~\\
MuRP~\cite{balazevic2019multi} is proposed to embed hierarchical multi-relational data in the Pointcar$\acute{e}$ ball of hyperbolic space, where the Pointcar$\acute{e}$ ball is defined as a $d$-dimensional manifold with the form: $\mathbb{B}^d_c  = \left \{ x\in \mathbb{R}^d:c\left \| x \right \| ^2<1, c>0\right \}$. {\textit{Compared with Euclidean space, hyperbolic surface can be seen to have more spaces to represent entities and capture hierarchy information with increasing radius~\cite{sala2018representation}}}, 
so that MuRP outperforms Euclidean KGE models and achieves better performance, especially in hierarchical datasets.
Instead of learning in a hyperbolic space with fixed curvature as in MuRP, ATTH~\cite{chami2020low} leverages expressive hyperbolic isometries to simultaneously capture logical patterns and hierarchies. For each relation (e.g., rotation, reflection), it learns a specific absolute curvature $c_r$ to avoid precision errors. 
Based on ATTH, models in different conditions are also proposed (i.e., ATTE, ROTE/H, REFE/H).
As also mentioned in section about polar coordinate, HBE~\cite{pan2021hyperbolic} employs an extended Pointcar$\acute{e}$ ball to capture hierarchical structures using polar coordinate system (Pointcar$\acute{e}$ disk) to solve the boundary constraints which might happen in conventional Pointcar$\acute{e}$ ball.
Sun et al.~\cite{sun2020knowledge} also represent KG embedding in a hyperbolic space by HyperKA, but firstly incorporate hyperbolic translation embedding with graph neural network (GNN).
However, with the constant emergence of hyperbolic models, Kai Wang et al.~\cite{wang2021hyperbolic} could not help asking: \textit{Is Hyperbolic Geometry necessary?} {Considering that hyperbolic-based model consistently necessitates greater computational complexity, which subsequently leads to the requirement for increased training resources, it raises the question of whether the benefits outweigh the associated costs?} 
To tackle this issue, RotL and Rot2L are proposed to simplify the hyperbolic operation in RotH. 
By defining Flexible Addition, RotL can reduce the computation complexity of RotH~\cite{chami2020low} and save over $50\%$ training time. However, it should not be ignored that knowledge graphs have multiple mixed relations, and excessive focus on hierarchical information often neglects capturing other information. Hyperbolic hierarchical transformations are introduced in HypHKGE~\cite{zheng2022hyperbolic} to extract hierarchies. MuRMP~\cite{wang2021mixed} utilises the mix-curvature model combined with GNN to better capture intrinsic heterogeneous structure in the KGs. UltraE~\cite{xiong2022ultrahyperbolic} considers an ultrahyperbolic manifold to overcome the non-hierarchical embedding problems. {Several contemporary KGE models (e.g., SEPA~\cite{gregucci2023link} and FFTAttH~\cite{xiao2022complex}) grounded in hyperbolic spaces have also demonstrated commendable performance.}

\subsubsection{Spherical Geometry}
~\\
Before introducing the spherical geometry models, in order to avoid the confusion of the concepts of spherical coordinate system and spherical geometry, we first make a specific distinction between them. The spherical coordinate system mentioned above can be consider as a tool to describe points' positions naturally. And the basic concepts behind it are points and straight lines based on the Euclidean geometry. However, in spherical geometry~\cite{harris1998spherical}, the basic concepts are point and great circles where any two lines meet in two points, and there are also no parallel lines. {\textit{Spheres are compatible with ring structures as the circular pattern of the vector field generated by spherical embedding has a natural circularity~\cite{suzuki2018riemannian}.
Consequently, spherical-based models also can yield competitive outcomes on complex relational datasets.}}

TransC~\cite{lv2018differentiating}, in order to differentiate instances and concepts, encodes each concept in knowledge graph as a sphere (e.g. concept $c_i$ is encoded as a sphere $s_i(\textbf{p}_i,m_i)$, where $\textbf{p}_i$ denotes the centre, $m_i$ denotes the radii) and each instance as a vector in the same semantic space. For example, for distinguishing the relations between concepts and sub-concepts (i.e. $\textit{subClassOf}$ ), TransC construct several possible positions (e.g. inclusion, intersection, separation)  between two concept spheres for different conditions.
Comparing with TransC, HypersphereE~\cite{dong2021hypersphere} extends the sphere into hypersphere so as to not neglect the uncertainty of instances.
Another special model named ManifoldE~\cite{xiao2015one}, which expands point-wised embedding to manifold-based embedding. Instead of adopting previous translational-based principle $\textbf{h}+\textbf{r} = \textbf{t}$, ManifoldE employs manifold-based principle $\mathbf{M}(\textbf{h,r,t}) = D_r^2$ ($\mathbf{M}$ is the manifold function) in sphere and hyperplane respectively. In Sphere condition, entity which is always considered as a point in classic models is now extended to a whole high dimensional sphere. Through this mean, ManifoldE avoids much noise so as to best distinguish true facts. {The smoothness of the sphere surface makes the embedding very flexible, and there can be countless mappings from the centre of the circle to the surface. }
As we know that KGs usually contain rich types of structure such as hierarchical and cyclic typed structures. Embedding KGs in single curvature space, such as Euclidean or hyperbolic space, overlooks the intrinsic heterogeneous structures of KGs, and therefore cannot accurately capture their structures. Hence, M$^2$GNN, as a mixed-curvature model, is designed to address this issue. The scoring function is defined as $\phi_{\mathbb{P}^{d_o,d_h,d_s}_{K_o,K_h,K_s}(e_h,r,e_t)}$, where $\mathbb{P}$ denotes the mixed-curvature space, $d_o,d_h,d_s$ are the dimensions of the component spaces of Euclidean, hyperbolic, and spherical, $K_{o/h/s}$ are the corresponding curvature of spaces. {Recently, SEA~\cite{gregucci2023link} utilises spherical geometry to consolidate various extant representations of KGE queries, thereby capturing diverse logical and structural patterns. Intriguingly, SEPA~\cite{gregucci2023link}, an alternative version of the SEA, can also be projected onto the Pointcar$\acute{e}$ ball to encompass more intricate structural representation.}

\subsection{Analytical Structure}
An analytical structure is usually thought of as a structure of having a measure. For instance, metric (or distance) is well defined in Euclidean space so that we can integrate, differentiate and other analytical operations. Similarly, in the probability space, the probabilistic measure is defined, so it can consider as an analytical structure. In this section, we will deeply dig into the KGE models by dividing them into two main spaces: Probability Space and Euclidean Space.

\subsubsection{Probability Space}
~\\
To the best of our knowledge, KG2E~\cite{he2015learning} is the first density-based KGE model which represents each entity/relation by a multi-dimensional Gaussian distribution $\mathcal{N}(\boldsymbol{\mu},\boldsymbol{\Sigma} )$ in probability space, where the mean vector $\boldsymbol{\mu}$ indicates its position and the covariance metric $\boldsymbol{\Sigma}$ indicates the corresponding (un)certainty which impacts on others. In addition, two similar methods based on KL-divergence and expected likelihood, are proposed to inspect the difference of asymmetric and symmetric respectively.
Previous models have formally considered the issue of multiple relation semantics in KGs. However, the traditional translational-based models always assign only one vector for one vector, ignoring the fact that a relation may have multiple meanings. Thus, TransG~\cite{xiao2015transg} is proposed by leveraging a Bayesian non-parametric infinite mixture model to handle multiple relation semantics by generating multiple translation components for a relation. In TransG, entities are generated by a certain Gaussian distribution, where the $m-$th component translation vector of relation $r$ is represented as: 
$\boldsymbol{\mu}_{r,m} = \mathbf{t} - \mathbf{h} \sim \mathcal{N}(\boldsymbol{\mu}_t - \boldsymbol{\mu}_h,(\sigma_h^2+\sigma_t^2)E)$. 
Through this process, TransG would automatically select the best match between h, t and r. Other probabilistic-based models such as DBKGE~\cite{liao2021learning} and GaussianPath~\cite{wan2021gaussianpath} also harness the uncertainty of KGs by Gaussian representation. {DiriE~\cite{wang2022dirie} is proposed by embedding entities as Dirichlet distributions and relations as multinomial distributions. This method uses Bayesian inference to assess the relationships between entities and subsequently learns binary embeddings of knowledge graphs for modelling intricate relation patterns and uncertainty. Recently, It$\hat{o}$E~\cite{nayyeri2023knowledge} formulates the relations in a KG as stochastic It$\hat{o}$ processes, enabling transitions between two nodes to occur with an associated likelihood. This approach permits It$\hat{o}$E to represent multiple stochastic trajectories including loops connected to paths and is mathematically substantiated as a generalisation of several state-of-the-art models. The aforementioned methodologies demonstrate that \textit{probabilistic embedding is not only capable of acquiring unstructured patterns but also adept at capturing additional uncertain information~\cite{xu2021understanding}.}}


\subsubsection{Euclidean Space}
~\\
In order to improve the structure preservation capabilities of KGE models, Nayyeri et al.~\cite{nayyeri2021knowledge} propose a novel KGE model named FieldE which employs ordinary differential equations (ODEs) for embedding KGs into a Euclidean space. Each entity is represented by a vector in $\mathbb{R}^n$ denoted by $\textbf{e}(t)$ and each relation is represented as a vector field $f_{\theta_{r}}$ on a Riemannian Manifold, where $\textbf{e}(t)$ lies on a trajectory (continuous) on the manifold $\mathcal{M}$ solving the ODE: $\frac{d\textbf{e}(t)}{dt} = f_{\theta_{r}}(\textbf{e}(t))$. Hence, FieldE could capture the continuity of changes in the embedding space and describe the underlying geometry by nature. ODE method can also be used in temporal KGs, in which TANGO~\cite{han2021temporal} is proposed to learn continuous-time representations of entities and relations dynamically by Neural ODE method~\cite{chen2018neural}. {\textit{These analytical methods facilitate the acquisition of dynamic and continuous representations of entities and relations, which in turn enhance memory efficiency, adaptive computation, and parameter effectiveness in various subsequent tasks~\cite{kobyzev2020normalizing}.}} {In addition to continuous analysis, other analytical perspectives such as derivability, differentiability, and integrability are also worth exploring in KGEs.}

\section{Downstream Tasks of Knowledge Graph Embedding}
\label{sec:applications}

After introducing a systematic review of existing KGE models from the perspective of mathematical structure, this section focuses on KGE-based downstream tasks.
We highlight some important and popular applications which are usually employed to evaluate the performance of embedding models. After summarising and comparing the performance of some KGE models, 
{we examine their strengths and weaknesses from diverse spatial perspectives and offer some advice for building KGE models. The summary of the advantages and disadvantages of different KGE models from the spatial perspective is provided in Table~\ref{tab:space_summary}.}

In what follows, we first describe the process of \textbf{Link prediction}, a fundamental task in KGE domain, 
and focus on one popular task: \textbf{Hierarchy Acquisition} in the link prediction scenario. In addition, we analyse and discuss the task of \textbf{Pattern Inference}. 
{Additionally, we present the model's results regarding the time/space complexity and discuss the applications of the KGE models in other downstream tasks. We also provide some suggestions for building KGE models based on the empirical results in Section~\ref{subsec:suggestions}.}

\begin{table*}[!tb]
    \caption{The link prediction results on the WN18RR and FB15k-237 datasets, which are classified by its geometric space. The best scores of 32-dimensional models are in \textbf{bold}, {the second best scores are \underline{underlined} and} the best average scores are \textbf{coloured}. TransE, ComplEx, QuatE, RotatE, MuRE, MuRMP, {and 5$\star$E} results are taken from ~\cite{xiong2022ultrahyperbolic}. TuckER, RefE, RotE, AttE, MuRP, RefH, RotH, AttH, and Rot2L results are {derived} from ~\cite{wang2021hyperbolic}. MuRMP is a mix-curvature model. UltraE is an ultrahyperbolic-based model, in which its best results (when $q=6$) are chosen. {Other results are taken from their original paper.} $^*$ denotes: DFieldE$_\mathbb{S}$ and MuRS results are not obtained in the case of low dimensions (d = 32), but in high dimensions. Hence the average score of the spherical-based model is not comparable. $^\triangledown$ denotes: GIE is a geometric interactive model and its result is also in high dimensional conditions, which is not added to the average calculation.}
    \small
    \centering
    \label{tab:2}
    \begin{tabular}{llccccccc}
        \toprule
        \multirow{2}*{\textbf{Type}} & \multirow{2}*{\textbf{Method}} & \multirow{2}*{\textbf{Year}}& \multicolumn{3}{c}{\textbf{FB15K-237}} & \multicolumn{3}{c}{\textbf{WN18RR}}  
        \\
        \cline{4-9}
        ~                            & ~         &                       & \textbf{MRR}                           & \textbf{Hits@10}                    & \textbf{Hits@1}         & \textbf{MRR}            & \textbf{Hits@10}        & \textbf{Hits@1}         \\
        \hline
        \multirow{14}*{\makecell[c]{Euclidean-based \\ Models}} 
        & TransE~\cite{bordes2013translating} & 2013 & 0.295 & 0.466 & 0.210 & 0.366 & 0.515 & 0.274\\
        & ComplEx~\cite{trouillon2016complex} & 2016 & 0.287 & 0.456 & 0.203 & 0.421 & 0.476 & 0.391\\
        & QuatE~\cite{zhang2019quaternion} & 2019 & 0.293 & 0.460 & 0.212 & 0.421 & 0.467 & 0.396\\
        & RotatE~\cite{sun2019rotate} & 2019  & 0.290 & 0.458 & 0.208 & 0.387 & 0.491 & 0.330\\
        & TuckER~\cite{balavzevic2019tucker} & 2019 & 0.306 & 0.475 & 0.223 & 0.428 & 0.474 & 0.401 \\
        & MuRE~\cite{balazevic2019multi} & 2019 & 0.313 & 0.489 & 0.226 & 0.458 & 0.525 & 0.421  \\
        & HAKE~\cite{zhang2020learning} & 2020 & 0.296 & 0.463 & 0.212 & 0.416 & 0.467 & 0.389 \\
        & RefE~\cite{chami2020low}  & 2020 & 0.302 & 0.474 & 0.216 & 0.455 & 0.521 & 0.419 \\
        & RotE~\cite{chami2020low} & 2020 & 0.307  & 0.482 & 0.220 & 0.463 & 0.529 & 0.426 \\
        & AttE~\cite{chami2020low} & 2020 & 0.311 & 0.488 & 0.223 & 0.456 & 0.526 & 0.419 \\
        & Rot2L~\cite{wang2021hyperbolic} & 2021 & 0.326 & 0.503 & 0.237 & 0.475 & 0.554 & 0.434\\
        & EucHKGE~\cite{zheng2022hyperbolic} & {2021} & {0.319} & {0.499} &
        {0.228} & {0.462} & {0.534} & {0.425}\\
        & It$\hat{o}$E$_\mathbb{R}$~\cite{nayyeri2023knowledge} & {2023} & {0.330} & {0.508} & {0.242} & {0.455} & {0.548} & {0.404} \\
        \cline{2-9}
        & \textit{Avg\_score}& - & {0.306} & {0.479} & {0.220} & {0.436} & {0.510} & {0.395} \\
        \hline
        \multirow{10}*{\makecell[c]{Hyperbolic-based\\ Models}} 
        & MuRP~\cite{balazevic2019multi} & 2019 & 0.323 & 0.501 & 0.235 & 0.465 & 0.544 & 0.420 \\
        & RefH~\cite{chami2020low} & 2020 & 0.312 & 0.489 & 0.224 & 0.447& 0.518 & 0.408   \\
        & RotH~\cite{chami2020low} & 2020 & 0.314 & 0.497 & 0.223 & 0.472 & 0.553& 0.428  \\
        & AttH~\cite{chami2020low} & 2020 & 0.324 & 0.501 & 0.236 & 0.466& 0.551 & 0.419  \\
        & HypHKGE~\cite{zheng2022hyperbolic} & 2021 & 0.330 & 0.510 & 0.240 & 0.475 & 0.556 & 0.432 \\
        & DFieldE$_\mathbb{P}$~\cite{nayyeri2021knowledge} & 2021 & 0.330 & 0.510 & \textbf{0.250} & 0.480 & \underline{0.570} & \underline{0.440}\\
        & FFTAttH~\cite{xiao2022complex} & {2022} & {0.331} & {\textbf{0.517}} & {0.239} & {0.476} & {0.558} & {0.432}\\
        & It$\hat{o}$E$_\mathbb{P}$~\cite{nayyeri2023knowledge} & {2023} & - & - & - & {0.474} & {\textbf{0.574}} & {0.426} \\
        & SEPA~\cite{gregucci2023link} & {2023} & {0.332} & {0.509} & {0.243} & {\underline{0.481}} & {0.562} & {\textbf{0.441}} \\
        \cline{2-9}
        & \cellcolor{m4}\textit{Avg\_score}   & \cellcolor{m4} -        & \cellcolor{m4}{0.325}                & \cellcolor{m4}{0.504}             & \cellcolor{m4}{0.236} & \cellcolor{m4}{0.471} & \cellcolor{m4}{0.554} & \cellcolor{m4}{0.427} \\
        \hline
        \multirow{4}*{\makecell[c]{Spherical-based\\Models}} 
        & DFieldE$_\mathbb{S}$~\cite{nayyeri2021knowledge} & 2021                         & 0.360                          & 0.550                               & 0.270         & -        & -           & -          \\
        & MuRS~\cite{wang2021mixed} & 2021 & 0.338 & 0.525 & 0.249 & 0.454 & 0.550 & 0.432  \\
        & It$\hat{o}$E$_\mathbb{S}$~\cite{nayyeri2023knowledge} & {2023} & {\underline{0.334}} & {0.511} & {0.245} & - & - & - \\
        \cline{2-9}
        & \textit{Avg\_score$^*$} & - & {0.344} & {0.529} & {0.255} & 0.454 & 0.550 & 0.432\\
        \hline
        \multirow{4}*{\makecell[c]{Mixed            Models}} 
        & MuRMP~\cite{wang2021mixed}  &2021                       & 0.319          & 0.502                             & 0.232           & 0.470       & 0.547            & 0.426           \\
        & UltraE~\cite{xiong2022ultrahyperbolic} & 2022 & \textbf{0.338}  & \underline{0.514} & \underline{0.247} & \textbf{0.483} & 0.555 & 0.425\\
        & GIE$^\triangledown$\cite{cao2022geometry} & 2022 & 0.362 & 0.552 & 0.271 & 0.491 & 0.575 & 0.452  \\
        \cline{2-9}
        & \cellcolor{m1}\textit{Avg\_score} & \cellcolor{m1}-&\cellcolor{m1}0.329  & \cellcolor{m1}0.508 & \cellcolor{m1}0.240                  & \cellcolor{m1}0.477   & \cellcolor{m1}0.551   &\cellcolor{m1}0.426                   \\
    \bottomrule        
    \end{tabular}
\end{table*}

\subsection{Link Prediction}
Link prediction aims to predict the existence of edges (triples) in knowledge graphs, which is a fundamental task since many existing KGs have missing facts or incorrect edges~\cite{nickel2015review}. 
In particular, the task of link prediction is often formulated as predicting missing entity in an incomplete fact triple, i.e., predicting head entity $h$ in $(?, r, t)$, or tail entity $t$ in $(h, r, ?)$, where $(?, r, t)$ and $(h, r, ?)$ denote fact triples with missing entities. For example, given a triple \textit{<?, subclass, COVID-19}>, the goal is to predict the superclass of COVID-19. 
Therefore, link prediction is also referred to as knowledge graph completion~\cite{nickel2015review}, entity prediction~\cite{lin2015modeling} or entity ranking~\cite{bordes2014semantic}.

Given learned entity and relation embeddings with KGE methods, link prediction is carried out through a ranking procedure. Specifically, in order to predict the tail entity of an incomplete fact triple $(h, r, ?)$, we take each entity $t'$ in the KG as {a} candidate answer and calculate the plausibility of the triple $(h,r,t')$, which is achieved by calculating the score function of the employed KGE method. For example, for KGE methods that learn KG embeddings in Euclidean space, the translation-based score function $s(\mathbf{h}, \mathbf{r}, \mathbf{t}) = -||\mathbf{h} + \mathbf{r} - \mathbf{t}||_{1/2}$ is often used to assign scores for triples. In contrast, for KGE methods that learn embeddings in complex space, the rotation-based score function $s(\mathbf{h}, \mathbf{r}, \mathbf{t}) = -||\mathbf{h} \circ \mathbf{r} - \mathbf{t}||^2$ is often utilised. 
After obtaining the scores of candidate answers, we can rank those candidate entities in the descending order of their scores, and select the highest ranked entity as the prediction result. A similar procedure can also be used to predict the missing head entity $h$ in $(?, r, t)$. For evaluation, a common practice is to record the ranks of correct answers in {the} previous ranked list and leverage those ranks to calculate the evaluation metrics.

\subsection{Hierarchy Acquisition}
Nowadays, more and more KGE models not only aim to obtain SOTA in link prediction task, but also pay special attention to whether they can capture hierarchy properties. 
{One main reason why these works focus on capturing hierarchy structures} is that: \textbf{Multi-relational knowledge graphs often exhibit multiple simultaneous hierarchies}~\cite{dettmers2018convolutional,toutanova2015observed}. However, conventional models (e.g., TransE, RotatE ) merely emphasis on capturing these hierarchies. Therefore, we focus on KGE models from the view of mathematical space, 
{aiming to find} the most appropriate space to capture multi-layered information. {In this} part, we first introduce what hierarchy exactly is, and then draw our conclusion by comparing the performance of KGE models in different spaces on hierarchy-contained datasets. 

\paragraph{What is Hierarchy?} Semantic hierarchy is a ubiquitous property in knowledge graphs. Some relations can induce various hierarchical structure. For instance, \textit{chair} is at a higher lever than \textit{armchair, fighting\_chair} under the relation \textit{hypernym}, and \textit{armchair, fighting\_chair} are parent nodes to their part: \textit{backrest, leg} with the relation \textit{has\_part}. Such hierarchies can be treated as "tree-like" structures intuitively.

\paragraph{Results.} We summarise the performance of KGE models in link prediction task on two hierarchical datasets: WN18RR \cite{dettmers2018convolutional} and FB15K-237 \cite{toutanova2015observed}, where 
the curvature metric $\xi_G$ (The lower the metric $\xi_\mathcal{G}$ is, the more hierarchical the knowledge graph is~\cite{gu2018learning}.) of two datasets are -2.54 and -0.65, respectively. 
{We summarise two key findings: \textbf{(1). Non-Euclidean (e.g., hyperbolic-based, spherical-based) models typically demonstrate a better capacity to capture various KG structures in low dimensions, as opposed to their Euclidean counterparts. (2). Within high-dimensional conditions, both non-Euclidean and Euclidean models exhibit comparable performance in representing KGs.}}

{To ensure a fair comparison, we substantiate our conclusions through multiple aspects. Firstly, we summarise the link prediction results of some state-of-the-art KGE models on the WN18RR and FB15k-237 datasets in Table~\ref{tab:2}. 
By analysing the empirical results of these models in Table~\ref{tab:2}, we can preliminarily infer that, by measured by the average scores metric (i.e., the \textit{Avg\_score}), Euclidean spatial models generally exhibit inferior performance compared to non-Euclidean spatial models in low-dimensional embeddings. In particular, hyperbolic-based models significantly outperform Euclidean baselines on WN18RR and FB15K-237 (highlighted in orange). Subsequently, Table~\ref{tab:compare} enumerates models based on distinct spaces from the same article to corroborate the conclusion. For instance, the performance of RefH/RotH/AttH is evidently superior to their Euclidean counterparts, RefE/RotE/AttE, across multiple evaluation metrics. Concurrently, Tables~\ref{tab:compare} and~\ref{tab:same_period} present KGE models over the same time period, effectively eliminating the influence of temporal factors on the analysis. Therefore, 
\textbf{(1)}. In low dimensions, hyperbolic embeddings present superior performances compared to Euclidean-based embeddings. Concurrently, spherical embedding also yields favourable results (See Table ~\ref{tab:2},~\ref{tab:compare} and~\ref{tab:same_period}). This phenomenon can be mathematically elucidated that: spheres are congruent with ring structures (as depicted in Figure~\ref{fig:KG_ring}) due to the circular nature of the vector field produced by spherical embedding exhibits circularity, facilitating the capture of cyclic structures~\cite{suzuki2018riemannian}. In contrast, vector fields in hyperbolic spaces consistently operate from the narrower regions of the manifold and progress towards its broader sides~\cite{nayyeri2021knowledge}. The curvature of hyperbolic spaces in low-dimensional settings exhibits a direct correlation with the calculated graph curvature~\cite{chami2020low}, making it suitable for tree-like or hierarchical configurations~\cite{nickel2018learning,ganea2018hyperbolic,tay2018hyperbolic,nickel2017poincare}.
\textbf{(2)}. However, when the embedding dimension is large, Euclidean, hyperbolic and spherical embedding methods perform similarly across all datasets (See Table~\ref{tab:compare} and~\ref{tab:same_period}). We explain this behaviour by noting that, given sufficiently large dimensions, both Euclidean and non-Euclidean spaces possess ample capacity to represent intricate hierarchies present in KGs. Consequently, the disparity among manifold selections in high-dimensional settings appears to be minimal. }

{Therefore, hyperbolic space provides effective methods for studying low-dimensional embeddings while maintaining underlying hierarchical structures, allowing hyperbolic-based models to embed tree-like structures with minimal distortion in merely two dimensions~\cite{chami2020low}. And the spherical space demonstrates the capacity to encapsulate ring structures, owing to its inherent property of extracting circularity.} 
Nevertheless, this does not mean that non-Euclidean models are necessarily superior to other spaces since some Euclidean-based models can also achieve high performance through ingenious tricks, such as ReflectE~\cite{zhang2022knowledge}, CompoundE~\cite{ge2022compounde}.
In addition, the results of the mixed models (coloured in purple) which learn embeddings in multiple spaces are more outstanding, which means that better performance can be achieved by using multiple geometries simultaneously with a suitable hybrid method.

\begin{table*}[t]
    \caption{{The link prediction results in different mathematical spaces from the \textbf{same paper} for low-dimensional ($d=32$) and high-dimensional (best for $d\in \{200,400,500\}$) embeddings.
    Models from the same article in different spaces are identified by identical symbols, whereas models originating from separate articles are distinguished by different symbols. [$\clubsuit$]: MuRE/P are from~\cite{balazevic2019multi}; [$\spadesuit$]: Results are from~\cite{balazevic2019multi}, where $\spadesuit_i$ denotes models using different geometric transformation; [$\diamondsuit$]: Euc/HypHKGE are from~\cite{zheng2022hyperbolic}; [$\heartsuit$]: It$\hat{o}$E$_{\mathbb{R}/\mathbb{P}/\mathbb{S}}$ are from~\cite{zheng2022hyperbolic}; [$\dagger$]: MuRS/MP are from~\cite{wang2021mixed}; 
    The best scores are in \textbf{bold}, the second best scores are \underline{underlined}.}
    }
    \small
    \centering
    \label{tab:compare}
    \resizebox*{!}{0.35\linewidth}{
    \begin{tabular}{clcccccccccccc}
        \toprule
        \multirow{2}*{\textbf{Space}} 
        & \multirow{2}*{\textbf{Method}} & \multicolumn{6}{c}{\textbf{WN18RR}} & \multicolumn{6}{c}{\textbf{FB15K-237}} 
        \\
        \cline{3-14}
        ~                            & ~                               & \multicolumn{3}{c}{\textbf{\textit{low-dimension}}}                          & \multicolumn{3}{c}{\textbf{\textit{high-dimension}}}                    & \multicolumn{3}{c}{\textbf{\textit{low-dimension}}}         & \multicolumn{3}{c}{\textbf{\textit{high-dimension}}}               \\ 
        ~ & ~  & \textbf{{MRR}} & \textbf{{Hits@1}} & \textbf{{Hits@10}} & \textbf{{MRR}} & \textbf{{Hits@1}} & \textbf{{Hits@10}} & \textbf{{MRR}} & \textbf{{Hits@1}} & \textbf{{Hits@10}} & \textbf{{MRR}} & \textbf{{Hits@1}} & \textbf{{Hits@10}}\\                            
        \cline{1-14}
        \multirow{6}{*}{$\mathbb{E}$}
        & MuRE~$\clubsuit$ (2019) & 0.458 & 0.421 & 0.525 & 0.475 &  0.436  & 0.554 
        & 0.313& 0.226 &0.489 & 0.336&0.245&0.521   \\
        & RefE~$\spadesuit_1$ (2020)  & 0.455 & 0.419 &  0.521 & 0.473  & 0.430  & 0.561 
        & 0.302 &0.216 &0.474 & 0.351  &0.256  & 0.541   \\
        & RotE~$\spadesuit_2$ (2020)& 0.463 & 0.426 & 0.529 & \underline{0.494}  &  0.446  & \underline{0.585} 
        & 0.307 & 0.220 & 0.482 & 0.346  & 0.251  & 0.538 \\
        & AttE~$\spadesuit_3$ (2020)& 0.456 & 0.419 & 0.526 & 0.490  & 0.443  & 0.581 
        & 0.311  & 0.223  & 0.488 & 0.351 & 0.255 & 0.543 \\
        & EucHKGE~$\diamondsuit$ (2021) & {0.462} & {0.425} & {0.474} & {0.493} & {0.447} & {0.583}
        &0.319&0.228&0.499&\underline{0.354}&\underline{0.261}&\underline{0.545}\\
        & It$\hat{o}$E$_\mathbb{R}$~$\heartsuit$ (2023)& {0.455} & {0.404} & {0.548} & - & - & -
        &\underline{0.330}&0.242&0.508&-&-&-\\
        \cline{1-14}
        \multirow{6}{*}{$\mathbb{H}$} 
        & MuRP~$\clubsuit$ (2019) & 0.465 & 0.420 & 0.544 & 0.481 & 0.440 & 0.566 
        &0.323&0.235&0.501&0.335&0.243&0.518 \\
        & RefH~$\spadesuit_1$ (2020)& 0.447 & 0.408 & 0.518 & 0.461& 0.404 & 0.568 
        &0.312&0.224&0.489 &0.346&0.252&0.536   \\
        & RotH~$\spadesuit_2$ (2020)& 0.472 & 0.428 & 0.553 & \textbf{0.496} & \textbf{0.449}& \textbf{0.586}
        &0.314&0.223&0.497&0.344&0.246&0.535\\
        & AttH~$\spadesuit_3$ (2020)& 0.466 & 0.419 & 0.551 & 0.486& 0.443 & 0.573
        &0.324&0.236&0.501&0.348&0.252&0.540\\
        & HypHKGE~$\diamondsuit$ (2021)& \textbf{0.475} & \textbf{0.432} &{0.556} &0.494&\underline{0.448}& 0.584
        & {0.330} & \underline{0.240} & \underline{0.510} &0.351&0.258& 0.541\\ 
        & It$\hat{o}$E$_\mathbb{P}$~$\heartsuit$  (2023) & \underline{0.474} & \underline{0.426} & \textbf{0.574} & -& - &-
        &-&-&-&-&-&-\\
        \cline{1-14} 
        \multirow{2}{*}{$\mathbb{S}$} 
        & MuRS~$\dagger$ (2021) & - & - & - & 0.454 & 0.432 & 0.550 
        &- & - & - &0.338&0.249&0.525 \\
        & It$\hat{o}$E$_\mathbb{S}$~$\heartsuit$ (2023) & - & - & - & - & - & - 
        &\textbf{0.334}&\textbf{0.245}&\textbf{0.511}&-&-&-\\
        \cline{1-14}
        \multirow{1}{*}{$\mathbb{M}$} 
        & MuRMP~$\dagger$ (2021) & 0.470 & 0.426 & 0.547 & 0.481 & 0.441& 0.569
        &0.319&0.232&0.502&\textbf{0.358}&\textbf{0.273}&\textbf{0.561}\\
    \bottomrule        
    \end{tabular}
    }
\end{table*}

\begin{table*}[t]
    \caption{{The scores or average scores of KGE models in Euclidean space ($\mathbb{E}$) and Hyperbolic space ($\mathbb{H}$) during the \textbf{same period}. [2019] are the scores of MuRE ($\mathbb{E}$) and MuRP ($\mathbb{P}$); [2020] are the average scores of Ref/Rot/AttE ($\mathbb{E}$) and Ref/Rot/AttH ($\mathbb{H}$); [2021] are the scores of EucHKGE ($\mathbb{E}$) and HypHKGE ($\mathbb{H}$}). 
    }
    \small
    \centering
    \label{tab:same_period}
    \resizebox*{!}{0.2\linewidth}{
    \begin{tabular}{cccccccccccccc}
        \toprule
        \multirow{3}*{\textbf{Year}} & \multirow{3}*{\textbf{Space}} & \multicolumn{6}{c}{\textbf{WN18RR}} & \multicolumn{6}{c}{\textbf{FB15K-237}}  
        \\
        \cline{3-14}
        &&\multicolumn{3}{c}{\textbf{\textit{low-dimension}}} & \multicolumn{3}{c}{\textbf{\textit{high-dimension}}} & \multicolumn{3}{c}{\textbf{\textit{low-dimension}}} & \multicolumn{3}{c}{\textbf{\textit{high-dimension}}}\\
        ~                            & ~                               & \textbf{MRR}                           & \textbf{Hits@1}                    & \textbf{Hits@10}         & \textbf{MRR}            & \textbf{Hits@1}        & \textbf{Hits@10} & \textbf{MRR}            & \textbf{Hits@1}        & \textbf{Hits@10} & \textbf{MRR}            & \textbf{Hits@1}        & \textbf{Hits@10}        \\
        \cline{1-14}
        \multirow{2}{*}{\shortstack{2019}} 
        & $\mathbb{E}$ & 0.458 & \textbf{0.421} & 0.525 & 0.475 &  0.436  & 0.554 
        & 0.313& 0.226 &0.489 & \textbf{0.336}&\textbf{0.245}&\textbf{0.521}   \\
        & $\mathbb{H}$  & \textbf{0.465} & 0.420 & \textbf{0.544} & \textbf{0.481} & \textbf{0.440} & \textbf{0.566} 
        &\textbf{0.323}&\textbf{0.235}&\textbf{0.501}&0.335&0.243&0.518\\
        \cline{1-14}
        \multirow{2}{*}{\shortstack{2020}} 
        & $\mathbb{E}$ & 0.458 & \textbf{0.421} & 0.527 & 0.485 & 0.444 & \textbf{0.575}
        &0.304&0.219&0.481&\textbf{0.348}&\textbf{0.254}&\textbf{0.540}\\ 
        & $\mathbb{H}$  & \textbf{0.461} & 0.418 & \textbf{0.540} & \textbf{0.481} & \textbf{0.432} & \textbf{0.575} 
        &\textbf{0.322}&\textbf{0.227}&\textbf{0.495}&0.346&0.250&0.537\\
        \cline{1-14}
        \multirow{2}{*}{\shortstack{2021}} 
         & $\mathbb{E}$ & {0.462} & {0.425} & {0.474} & {0.493} & {0.447} & {0.583}
        &0.319&0.228&0.499&\textbf{0.354}&\textbf{0.261}&\textbf{0.545}\\
        & $\mathbb{H}$ &\textbf{0.475} &\textbf{0.432} &\textbf{0.556} &\textbf{0.494}&\textbf{0.448}& \textbf{0.584}
        &\textbf{0.330} &\textbf{0.240} &\textbf{0.510} &0.351&0.258& 0.541\\
    \bottomrule        
    \end{tabular}
    }
\end{table*}

\subsection{Patterns inference}
\label{subsec:pattern}
Another popular task is about exploring the pattern inference capability since large-scale KGs always exhibit various types of relationships. As described in ~\cite{bordes2013irreflexive}, a excellent model should be able to learn all combinations of these properties: (a) \textbf{symmetry} (e.g., marriage, is\_similar\_to). (b) \textbf{antisymmetry} (e.g., father\_of). (c) \textbf{inversion} (e.g., hypernym and hyponym). (d) \textbf{composition} (e.g., my mother's father is my grandpa.).
However, looking back at the development of previous KGE models, it was a difficult process to build a model that could capture all of the above attributes simultaneously. In this section, we will mainly analyse the advantages of models focusing on inferring relational patterns, and summarise the key factors of capturing those properties.\\
~\\
First, we give these four important patterns' formal definitions as below~\cite{sun2019rotate}:
\begin{definition}
    A relation $\ttr$ is \textbf{symmetric} if $\forall x, y$
    \begin{equation*}
        \mathbf{r}(x, y) \Rightarrow \mathbf{r}(y, x).
    \end{equation*}
    A clause with such form is a \textbf{symmetry} pattern.
\end{definition}

\begin{definition}
    A relation $\mathbf{r}$ is \textbf{antisymmetric} if $\forall x, y$
    \begin{equation*}
        \mathbf{r}(x, y) \Rightarrow \neg \mathbf{r}(y,x).
    \end{equation*}
    A clause with such form is a \textbf{antisymmetry} pattern.
\end{definition}

\begin{definition}
    Relation $\mathbf{r}_1$ is \textbf{inverse} to relation $\mathbf{r}_2$ if $\forall x, y$
    \begin{equation*}
        \mathbf{r}_2 (x, y) \Rightarrow \mathbf{r}_1 (y, x).
    \end{equation*}
    A clause with such form is a \textbf{inversion} pattern.
\end{definition}

\begin{definition}
    Relation $\mathbf{r}_1$ is \textbf{composed} of relation $\mathbf{r}_2$ and relation $\mathbf{r}_3$ if $\forall x, y, z$
    \begin{equation*}
        \mathbf{r}_2 (x, y) \wedge \mathbf{r}_3 (y, z) \Rightarrow \mathbf{r}_1 (x, z).
    \end{equation*}
    A clause with such form is a \textbf{composition} pattern.
\end{definition}

Next, we analyse specific operations in different spaces to explain how existing models can infer and model those patterns. Table.\ref{tab:pattern_inf} shows the patterns inference capabilities of some main KGE works, and their corresponding spaces and operators.

\paragraph{Addition.} The KGE models with addition operation as their core are often found among distance-based models such as SE~\cite{bordes2011learning}, TransE~\cite{bordes2013translating} and TransE's variants. The characteristic of addition is that it is easy to establish the connection between vectors, and the complexity is very low.
In SE~\cite{bordes2011learning}, the scoring function is too sketchy to capture any of these patterns; By defining $h + r \approx t$, TransE~\cite{bordes2013translating} makes good use of vector addition to establish the relationship between relations and entities, and can capture three patterns except symmetry.
Since then, a large number of translation-based models(e.g., TransH~\cite{wang2014knowledge}, TransR~\cite{lin2015learning}, etc.) still retain addition operation but carry out special operation $f_r(\cdot)$ (mostly matrix multiplication based on r). What remains unchanged is that the core of operation is still Addition, that is, to a large extent, addition corresponds to the concept of ``translation''.
However, with the added projection parameters, these models are unable to encode inverse and composition. Although they have made progress in dealing with complex relations, the Trans's variants are a step back in terms of modelling patterns --- a drawback of the simplicity of addition.

\paragraph{Product.} The term ``product'' refers to the results of one or more multiplications. Most existing SOTA models use product operation, such as ComplEx~\cite{trouillon2016complex} and RotatE~\cite{sun2019rotate}. Products in different spaces have different properties to help the KGE models infer more patterns. Here we mainly divide the product operations into the following groups and analyse them in detail.

\begin{itemize}
    \item \textit{Dot Product.} The dot product may be defined algebraically or geometrically. With algebraic definition, inner product is a way to multiply vectors together, with the result of this multiplication being a scalar. With geometric definition, the notions of length and angles can be defined by means of the dot product. In KGE, the fundamental purpose of product is to expect to establish complex relations between relations and entities through multiplication rather than simple addition. For example, the scoring function in DistMult~\cite{yang2014embedding} is defined as $f_r(h,t)=\textbf{h}^{\top}diag(\textbf{r})\textbf{t}=\sum_i\left [ \textbf{r} \right ]_i\cdot\left [ \textbf{h} \right ]_i\cdot\left [ \textbf{t} \right ]_i$, we can see that each score in the summation is a direct multiplication of $\textbf{h}_i\textbf{r}_i\textbf{t}_i$; similar in ComplEx~\cite{trouillon2016complex}, the scoring function is extended to complex space. Note that they are both RESCAL~\cite{nickel2011three}'s extensions, and RESCAL was originally built for implicit semantic matching by factorisation, which is one of the latent advantages of dot products.
    \item \textit{Hadamard Product.} Hadamard Product (also known as element-wise product) is a binary operation by which the elements corresponding to {the} same row and columns of given vectors/matrices are multiplied together to form a new vector/matrix. The difference between {the} dot product and {the} Hadamard product operationally is the aggregation by summation. The dot product of two vectors gives only a scalar number while the Hadamard product of two vectors gives a complete vector, which preserves a large amount of transformed information of KGs. As in RotatE~\cite{sun2019rotate}, through the principle $\textbf{t} = \textbf{h} \circ \textbf{r}$, \textbf{r} becomes a element-wise rotation from the head entity to the tail entity; In HAKE~\cite{zhang2020learning}, each $\textbf{r}_i$ is regarded as a scaling transformation between two moduli; Cross interaction operations are also applied by utilising Hadamard product in CrossE~\cite{zhang2019interaction}. Similarly uses of Hadamard product also appear in ComplEx~\cite{trouillon2016complex}, PairRE~\cite{chao2021pairre}, etc. In general, the Hadamard product is closely related to the concept of ``rotation''. Besides, the Hadamard product appears in lossy compression algorithms such as JPEG, can also be used for describing NN as LSTM~\cite{hochreiter1997long}, GRU~\cite{cho2014learning} or enhancing, suppressing or masking image regions.
    \item \textit{Other Products.} Other subsequent product operations are mostly extended based on the existing advantages of Dot/Hadamard product. Zhang et al.~\cite{zhang2019quaternion} believed that latent inter-dependencies between all components are aptly captured with Hamilton product, encouraging a more compact interaction between entities and relations; What's more advanced than RotatE is that DualE~\cite{cao2021dual} adopts quaternion inner product operation to model both translation and rotation, which improves the capability of inferring three important patterns.
\end{itemize}

\paragraph{Other Operations.} 
Other operations are mostly those with exclusive rules under some special conditions, but show strengths in some specific tasks (e.g., multi-relation tasks). For example, simple matching in SME~\cite{bordes2014semantic}/RESCAL~\cite{nickel2011three}, although it only uses simple linear/bilinear matching, it opens a new field for semantic matching KGE model; Circular correlation in HolE~\cite{nickel2016holographic}, which makes compressions of pairwise tensor product to enhance its efficiency; Orthogonal Transform~\cite{nickel2016holographic} that aims to Unleash the original potential of RotatE into higher dimensions; Mobius addition and attention in ATTH~\cite{chami2020low} are utilised to represent relations as parameterised geometric operations that directly map to logical properties.

\begin{table*}[!t]
    \caption{The pattern inference abilities of several models by using different operators in certain spaces. SE, TransE, TransR, DisMult, ComplEx, and RotatE results are taken from ~\cite{sun2019rotate}. Other results come from origin papers. $\mathbb{R}$ represents the Euclidean space. $\mathbb{R}_p$ represents the Euclidean space with polar coordinate system~\cite{zhang2020learning}. $\mathbb{C}$ represents the complex vector space. $\mathbb{G}$ represents the group. $\mathbb{H}$ represents the hyperbolic space. $\mathbb{UH}$ represents the ultra-hyperbolic space. $\mathbb{Q}$ represents the quaternion vector space. $\mathbb{BQ}$ and $\mathbb{DQ}$ represents the biquaternion vector space and dual quaternion vector space, respectively, both of them are special cases of $\mathbb{Q}$.}
    \label{tab:pattern_inf}
    \begin{center}
    \resizebox*{!}{0.65\linewidth}{
        \begin{tabular}{lcccccc}
            \toprule
            \multirow{2}{*}{\textbf{Method}} & \multirow{2}{*}{\textbf{Manifold}}
                                             & \multirow{2}{*}{\textbf{Operator}}
                                             & \multicolumn{4}{c}{\textbf{Relation Patterns}}                                                                                                                  \\
                                             &                                                 &                                                                  & $Sym$  & $Asym$           & $Inv$  & $Comp$ \\ \hline
            SE~\cite{bordes2011learning}                               & $\mathbb{R}$                                       & Addition($+$)                                                    & \xmark & \xmark           & \xmark & \xmark \\ \hline
            TransE~\cite{bordes2013translating}                           & $\mathbb{R}$                                       & Addition($+$)                                                    & \xmark & \cmark           & \cmark & \cmark \\ \hline
            TransR~\cite{lin2015learning}                           & $\mathbb{R}$                                       & Addition($+$)                                                    & \cmark & \cmark           & \xmark & \xmark \\ \hline
            DistMult~\cite{yang2014embedding}                         & $\mathbb{R}$                                       & Inner Product($\left \langle \cdot  \right \rangle$)             & \cmark & \xmark           & \xmark & \xmark \\ \hline
            HolE~\cite{nickel2016holographic}                          & $\mathbb{R}$                                       & Circular Coorelation($\star$)                                    & \cmark & \cmark           & \cmark & \xmark \\ \hline
            HAKE~\cite{zhang2020learning}                             & $\mathbb{R}_{p}$                                & Hadamard Product($\circ$)                                         & \cmark & \cmark           & \cmark & \cmark \\ \hline
            PairRE~\cite{chao2021pairre}                           & $\mathbb{R}$                                       & Hadamard Product($\circ$)                                         & \cmark & \cmark           & \cmark & \cmark \\ \hline
            ComplEx~\cite{trouillon2016complex}                          & $\mathbb{C}$                              & Hadamard Product($\circ$)                                         & \cmark & \cmark           & \cmark & \xmark \\ \hline
            RotatE~\cite{sun2019rotate}                           & $\mathbb{C}$                              & Hadamard Product($\circ$)                                         & \cmark & \cmark           & \cmark & \cmark \\ \hline
            ExpressivE~\cite{pavlovic2022expressive}                           & $\mathbb{R}$                                       & Hadamard Product($\circ$)                                         & \cmark & \cmark           & \cmark & \cmark \\ \hline
            OTE~\cite{tang2020orthogonal}                              & $\mathbb{R}$                                       & Orthogonal Transform($\phi$)                                     & \cmark & \cmark           & \cmark & \cmark \\ \hline
            QuatE~\cite{zhang2019quaternion}                            & $\mathbb{Q}$                           & {Hamilton Product}($\otimes$)                                     & \cmark & \cmark           & \cmark & \xmark \\ \hline
            BiQUE~\cite{guo2021bique}                            & $\mathbb{BQ}$                               & {Hamilton Product}($\otimes$)                                     & \cmark & \cmark           & \cmark & \cmark \\ \hline
            DualE~\cite{cao2021dual}                            & $\mathbb{DQ}$                            & Dual Quaternion Product($\left \langle \otimes  \right \rangle$) & \cmark & \cmark           & \cmark & \cmark \\ \hline
            Rotate3D~\cite{gao2020rotate3d} & $\mathbb{R}$ & Rotational Product($\odot$)& \cmark & \cmark           & \cmark & \cmark \\ \hline
            HousE~\cite{li2022house}        & $\mathbb{R}$ & Householder Rotation($\mathcal{S}$)& \cmark & \cmark           & \cmark & \cmark \\ \hline
            ATTH~\cite{chami2020low}                             & $\mathbb{H}$                                      & M$\ddot{o}$bius addition($\oplus^c$)                  & \cmark & \cmark           & \cmark & \cmark \\ \hline
            DihEdral~\cite{xu2019relation}                         & $\mathbb{G}$                                           & Matrix Product($\odot^*$)                                          & \cmark & \cmark & \cmark & \cmark \\ \hline
            UltraE~\cite{xiong2022ultrahyperbolic} & $\mathbb{UH}$ & Ultrahyperbolic Transform($f_{\mathbf{r}}$) & \cmark & \cmark           & \cmark & \cmark \\ \bottomrule
        \end{tabular}}
    \end{center}

\end{table*}

\subsection{Knowledge Infusion To Enhance Other Domain Applications}
\label{subsec:other_domain}
Apart from the aforementioned applications that are appropriate for KG embeddings, there are other wider fields where KG representation learning could be infused to enhance.
Knowledge graph based questions answering (KGQA) is a fundamental, but still challenging task. It recognises the user's question input in order to obtain the accurate answers composed of KG entities. Existing methods include semantic parsing models~\cite{yih2015semantic,liang2016neural,lan2020query}, reinforcement learning~\cite{xiong2017deeppath,qiu2020stepwise}, and so on. 
Knowledge reasoning is a process of using known knowledge to infer new knowledge. Researchers always use machine learning methods~\cite{wang2021kepler,zhang2019ernie,wang2019kgat} to infer potential relations between entity pairs and identify erroneous knowledge based on existing data automatically. There are lots of other external applications based on knowledge graph embedding are still worth exploring, such as Recommendation System~\cite{bellini2017auto,wang2019multi,zhao2017meta}, Information Retrieval~\cite{raviv2016document,ensan2017document,liu2015latent} and other specific domain (e.g. Cyber Security~\cite{li2023k,host2023constructing}, Biomedicine~\cite{vilela2023biomedical,carvalho2023knowledge}, etc.). {The recent surge of LLMs~\cite{brown2020language,touvron2023llama,scao2022bloom,cui2023efficient,cao2023relmkg,choi2023knowledge}) has demonstrated exceptional proficiency in handling diverse Natural Language Processing (NLP) tasks, including question answering, machine translation, and text generation. Several studies~\cite{zhang2019ernie,rosset2020knowledge,zhang2020pretrain,sung2021can,yasunaga2021qa,cheng2023editing} have been conducted to substantiate the benefits provided by models integrating knowledge graphs and large language models. }

\subsection{Model Complexity}
\label{subsec:model_complexity}

{In this section, we incorporate Table~\ref{tab:complexity} to examine the time and space complexity of some KGE models. 
Based on the analyses in Table~\ref{tab:complexity}, we can draw the following conclusions. First, models which represent entities and relations as vectors (e.g., TransE, TransH, ComplEx, and UltraE) are more efficient. They usually have space and time complexity that scales linearly with entity dimension $d$. HolE needs more time complexity as it computes circular correlation via the discrete Fourier transform. Second, models which represent relations as matrices (e.g., TransR, SE, and RESCAL) usually have higher complexity in both space and time. Third, advanced methods (e.g., RotE/H, Rot2L, and It$\hat{o}$E) may lead to a linear increase in space complexity but not to order. RotE/H requires more relation parameters since relation transformation vectors and the learnable curvature for different relations are needed. 
It is important to mention that non-Euclidean models exhibit only marginal deviations from Euclidean models in terms of time or space complexity. Regarding the training cost, it has been demonstrated that hyperbolic embedding typically necessitates more training time than Euclidean embedding, owing to the fact that the M$\ddot{o}$bius operations in hyperbolic are far more complex than the Euclidean operations~\cite{wang2021hyperbolic}. Simultaneously, the training cost of various models within the same space tends to fluctuate based on factors such as model size, mapping method~\cite{li2022house} (linear or nonlinear), and the usage of acceleration algorithms~\cite{li2021efficient,zhang2013optimizing,wang2018incorporating}. Consequently, we conclude that determining the complexity of a KGE method is a multidimensional and comprehensive challenge. It is worth emphasising that our survey's primary objective is to analyse KGE methods from the standpoint of representation space, and as such, we do not further explore the intricacies of complexity.}

\begin{table*}[!t]
    \caption{{Comparison of state-of-the-art knowledge graph embedding models, along with their publisher, math space, time complexity, and space complexity. The complexity results are taken from \cite{wang2017knowledge,mohamed2021biological} or referenced from their corresponding papers. RotE/H results are obtained from Rot2L~\cite{wang2021hyperbolic}. $d$ and $k$ are the embedding dimensionality of entities and relations, respectively (usually $d=k$). $N_e$ and $N_r$ are the numbers of entities and relations. $\theta$ denotes the average sparseness degree of projection matrices in TranSparse~\cite{ji2016knowledge}.}}
    \label{tab:complexity}
    \begin{center}
    \resizebox*{!}{0.6\linewidth}{
        \begin{tabular}{lllll}
            \toprule
            \multicolumn{1}{l}{\textbf{Method}} & \multicolumn{1}{l}{\textbf{Publisher}} & \textbf{Math Space} & \textbf{Time Complexity} & \textbf{Space Complexity}       \\ \cmidrule{1-5}
            SE~\cite{bordes2011learning} & AAAI 2011 & Euclidean & $\mathcal{O}(d^2)$ & $\mathcal{O}({N}_ed + {N}_rd^2)$ \\
            RESCAL~\cite{nickel2011three} & ICML 2011 & Euclidean & $\mathcal{O}(d^2)$ & $\mathcal{O}({N}_ed + {N}_rd^2)$ \\
            TransE~\cite{bordes2013translating} & NeurIPS 2013  & Euclidean & $\mathcal{O}(d)$ & $\mathcal{O}({N}_ed + {N}_rd)$ \\
            TransH~\cite{wang2014knowledge} & AAAI 2014 & Euclidean & $\mathcal{O}(d)$ & $\mathcal{O}({N}_ed + {N}_rd)$ \\
            TransR~\cite{lin2015learning} & AAAI 2015 & Euclidean & $\mathcal{O}(dk)$ & $\mathcal{O}({N}_ed + {N}_rdk)$ \\
            TransD~\cite{ji2015knowledge} & ACL-IJCNLP 2015 & Euclidean & $\mathcal{O}(\max(d,k))$ & $\mathcal{O}({N}_ed + {N}_rk)$ \\
            DistMult~\cite{yang2014embedding} & ICLR 2015 & Euclidean & $\mathcal{O}(d)$ & $\mathcal{O}({N}_ed + {N}_rd)$ \\
            ComplEx~\cite{trouillon2016complex} & ICML 2016 & Euclidean & $\mathcal{O}(d)$ & $\mathcal{O}({N}_ed + {N}_rd)$ \\
            TranSparse~\cite{ji2016knowledge} & AAAI 2016 & Euclidean & $\mathcal{O}(dk)$ & $\mathcal{O}({N}_ed + (1-\theta){N}_rdk)$ \\ 
            HolE~\cite{nickel2016holographic} & AAAI 2016 & Euclidean & $\mathcal{O}(d\log d)$ & $\mathcal{O}({N}_ed + {N}_rd)$ \\ 
            ANALOGY~\cite{liu2017analogical} & ICML 2017 & Euclidean & $\mathcal{O}(d)$ & $\mathcal{O}({N}_ed + {N}_rd)$ \\ 
            ConvE~\cite{dettmers2018convolutional} & AAAI 2018 & Euclidean & $\mathcal{O}(d)$ & $\mathcal{O}({N}_ed + {N}_rd)$ \\
            RotE~\cite{chami2020low} & ACL 2020 & Euclidean & $\mathcal{O}(d)$ & $\mathcal{O}({N}_ed + 2N_rd)$ \\
            Rot2L~\cite{wang2021hyperbolic} & ACL Findings 2021 & Euclidean & $\mathcal{O}(d)$ & $\mathcal{O}({N}_ed + 2(N_r+5)d)$ \\
            It$\hat{o}$E$_\mathbb{R}$~\cite{nayyeri2023knowledge} & ACL Findings 2023 & Euclidean & $\mathcal{O}(d)$ & $\mathcal{O}({N}_ed + {N}_rk)$ \\
            RotH~\cite{chami2020low} & ACL 2020 & Hyperbolic & $\mathcal{O}(d)$ & $\mathcal{O}({N}_ed + 3(N_r+1)d)$ \\ 
            It$\hat{o}$E$_\mathbb{P}$~\cite{nayyeri2023knowledge} & ACL Findings 2023 & Hyperbolic & $\mathcal{O}(d)$ & $\mathcal{O}({N}_ed + {N}_rk)$ \\
            ManifoldE$_\mathbb{S}$~\cite{xiao2015one} & IJCAI 2016 & Spherical & $\mathcal{O}(d)$ & $\mathcal{O}({N}_ed + {N}_rd)$ \\
            It$\hat{o}$E$_\mathbb{S}$~\cite{nayyeri2023knowledge} & ACL Findings 2023 & Spherical & $\mathcal{O}(d)$ & $\mathcal{O}({N}_ed + {N}_rk)$ \\
            UltraE~\cite{xiong2022ultrahyperbolic} & KDD 2022 & Ultrahyperbolic & $\mathcal{O}(d)$ & $\mathcal{O}({N}_ed + {N}_rd)$ 
            \\\bottomrule
        \end{tabular}
        }
    \end{center}

\end{table*}

\begin{table*}[!t]
    \centering
    \footnotesize
    \renewcommand{\arraystretch}{1.2}
    \caption{{The advantages and disadvantages of different KGE models from the spatial perspective.}}
    \label{tab:space_summary}
    \vspace{-0.5em}
    \resizebox*{!}{0.85\linewidth}{
    \begin{tabular}{llllll}
        \toprule
        \textbf{Perspective} &\textbf{Subclass/Subspace} &\textbf{Examples} & \textbf{Properies} & \textbf{Advantages} & \textbf{Disadvantages} \\
        \midrule
        \multirow{13}{1.6cm}{\shortstack{Geometric \\Structure}} & 
        \multirow{5}{2.5cm}{Euclidean Geometry} & \multirow{5}{3.3cm}{TransE~\cite{bordes2013translating}; RotatE~\cite{sun2019rotate}; PairRE~\cite{chao2021pairre}; TripleRE~\cite{yu2022triplere}; TranS~\cite{zhang2022trans}; HopfE~\cite{bastos2021hopfe}; CompoundE~\cite{ge2022compounde}; HAKE~\cite{zhang2020learning}; H$^2$E~\cite{wang2021knowledge}}
        & \multirow{5}{3.2cm}{\parbox{3.2cm}{The prevalent Euclidean geometric transformations can effectively capture simple relational patterns and characteristics.}} 
        & \multirow{13}{3.3cm}{\parbox{3.3cm}{$\bullet$ It is easy and intuitive to bridge relationships between entities through a variety of geometric transformations, such as \textit{translation, rotation, reflection} and \textit{scaling}.\\}  \parbox{3.3cm}{~\\ $\bullet$ The distinctive and heterogeneous architecture of knowledge graphs is closely associated with geometric embedding (e.g., Ring KGs \textit{vs.} Spheres~\cite{suzuki2018riemannian}). \\} \parbox{3.3cm}{~\\ $\bullet$ Non-Euclidean models are suitable for complex KGs tasks such as link prediction.}}
        & 
      \multirow{12}{3.2cm}{\parbox{3.2cm}{$\bullet$ Advanced geometric embedding methods (e.g., hyperbolic embedding) tend to be more complex than Euclidean embedding and also require more training time and cost~\cite{wang2021hyperbolic}.\\} \parbox{3.2cm}{$\bullet$ Certain geometries might exhibit diminished benefits for graphs containing long paths~\cite{sala2018representation}.}} \\
      &&&&&\\ &&&&&\\ &&&&&\\ &&&&&\\ \cmidrule{2-4}
      & \multirow{5}{*}{Hyperbolic Geometry} & \multirow{5}{3.3cm}{MuRP~\cite{balazevic2019multi}; ATTH~\cite{chami2020low}; HBE~\cite{pan2021hyperbolic}; HyperKA~\cite{sun2020knowledge}; UltraE~\cite{xiong2022ultrahyperbolic}; 
      It$\hat{o}$E~\cite{nayyeri2023knowledge}; HypHKGE~\cite{zheng2022hyperbolic}; H$^2$E~\cite{wang2021knowledge}; H$^2$E~\cite{wang2021knowledge}; HBE~\cite{pan2021hyperbolic}} 
      & \multirow{5}{3.2cm}{\parbox{3.2cm}{Hyperbolic surface has more (tree-like) spaces to represent and capture hierarchy information.}} 
      & & \\
      &&&&&\\ &&&&&\\ &&&&&\\ &&&&&\\ \cmidrule{2-4}
      & \multirow{4}{*}{Spherical Geometry} & \multirow{4}{3.3cm}{ManifoldE~\cite{xiao2015one}; GIE~\cite{cao2022geometry}; TransC~\cite{lv2018differentiating}; MuRS~\cite{wang2021mixed}; HyperspherE~\cite{dong2021hypersphere}; SEA~\cite{gregucci2023link}} 
      & \multirow{4}{3.2cm}{\parbox{3.2cm}{Sphere excels at capturing ring structures by its circularity nature of spherical embedding.}} 
      & & \\
      &&&&&\\ &&&&&\\ &&&&&\\ \midrule
      \multirow{14}{1.6cm}{\shortstack{Algebraic \\Structure}} & 
        \multirow{5}{2.5cm}{Vector Space} & \multirow{5}{3.3cm}
       {TransE~\cite{bordes2013translating}; RESCAL~\cite{nickel2011three}; TransH~\cite{wang2014knowledge}; DisMult~\cite{yang2014embedding}; 
        ComplEx~\cite{trouillon2016complex}; RotatE~\cite{sun2019rotate}; QuatE~\cite{zhang2019quaternion}; 
        DualE~\cite{cao2021dual}; ConvE~\cite{dettmers2018convolutional};
        R-GCN~\cite{schlichtkrull2018modeling}; KG-BERT~\cite{yao2019kg};  DKRL~\cite{xie2016DKRL}}
        & \multirow{5}{3.2cm}{\parbox{3.2cm}{With simple and efficient operations in vector space, the plausibility of facts can be easily measured by matching the latent semantics of entities and relations.}}&
        \multirow{10}{3.2cm}{\parbox{3.2cm}{~\\$\bullet$ Using proper algebraic operations can help KGE models capture more important patterns of knowledge graphs (\emph{e.g.,} \textit{symmetry, antisymmetry, inversion,} and \textit{composition}).\\} \parbox{3.2cm}{~\\ $\bullet$ The plausibility of facts can be measured by matching latent semantics of entities and relations through efficient linear operations.~\cite{nickel2016holographic,kazemi2018simple} }} 
        & \multirow{10}{3.2cm}{\parbox{3.2cm}{~\\$\bullet$ It should be noted that the demonstrations of various algebraic structures are solely dependent on the specific KG task itself, and the extent to which they can be generalised to other domains remains uncertain.\\} \parbox{3.2cm}{$\bullet$ Some operations, such as M$\ddot{o}$bius addition~\cite{chen2021mobiuse}, may be less intuitive and consequently challenging to extend to numerous scenarios.}}\\
        &&&&&\\ &&&&&\\ &&&&&\\ &&&&&\\ \cmidrule{2-4}
        & \multirow{4}{2.5cm}{Group} & \multirow{4}{3.3cm}
        {TorusE~\cite{ebisu2018toruse}; DihEdral~\cite{xu2019relation}; NagE~\cite{yang2020nage}; ModulE~\cite{chai2022module}; KGLG~\cite{ebisu2019generalized}; DensE~\cite{lu2020dense}; GrpKG~\cite{yang2021knowledge}}
        & \multirow{4}{3.2cm}{\parbox{3.2cm}{The definition of group can naturally satisfy the basic properties (e.g., \textit{inversion, composition}) of KGs.}}& &\\
        &&&&&\\ &&&&&\\ &&&&&\\ \cmidrule{2-4}
        & \multirow{3}{*}{Ring} & 
        \multirow{3}{3.5cm}{M$\ddot{o}$biusE~\cite{chen2021mobiuse}} & \multirow{3}{3.3cm}{Ring structure is helpful for orientation-related tasks due to its non-oriented surface.} & &\\ 
        &&&&&\\ 
        \\ \midrule
        \multirow{11}{1.6cm}{\shortstack{Analytical \\Structure}} &
        \multirow{4}{2.5cm}{Probability Space} &\multirow{4}{3.3cm}{TransG~\cite{xiao2015transg}; KG2E~\cite{he2015learning}; DiriE~\cite{wang2022dirie}; It$\hat{o}$E~\cite{nayyeri2023knowledge}; GaussianPath~\cite{wan2021gaussianpath} }
        & \multirow{4}{3.2cm}{\parbox{3.2cm}{Probabilistic embedding is capable of acquiring unstructured patterns and capturing uncertain information.}}
        &\multirow{8}{3.2cm}{\parbox{3.2cm}{$\bullet$ Concentrating on analytical properties (\emph{e.g.,} \textit{uncertainty, continuity,} and \textit{differentiability}.) in the modelling process enables the KGE systems to capture stable and robust representations~\cite{nayyeri2023knowledge,nayyeri2021knowledge}.}} & \multirow{7}{3.2cm}{\parbox{3.2cm}{$\bullet$ The analytical nature primarily serves as an auxiliary component in the KGE modelling process and is always not the predominant factor in achieving optimal performance in downstream tasks.}} \\ 
        &&&&&\\ &&&&&\\ &&&&&\\\cmidrule{2-4}
        & \multirow{4}{2.5cm}{Euclidean Space} &
        \multirow{4}{3.3cm}{FieldE~\cite{nayyeri2021knowledge}; TANGO~\cite{han2021temporal}}&\multirow{4}{3.2cm}{\parbox{3.2cm}{The analytical methods (e.g., ODE) facilitate the acquisition of dynamic and continuous representations of entities and relations.}} & \\
        &&&&&\\ &&&&&\\ &&&&&\vspace{3pt}\\  
        \bottomrule
    \end{tabular}
    }
\end{table*}

\subsection{Suggestions}
\label{subsec:suggestions}

{In Section~\ref{sec:applications}, we present the performance of {different} KGE models on different downstream tasks and applications from the mathematics spatial perspective. Here we {summarise our analyses} and provide suggestions and guidance for constructing KGE models: 
(1). Firstly, we analyse the performances of KGE models in different geometric spaces {based on the empirical results in} Table~\ref{tab:2}. The results show that the \textit{hyperbolic-based} models have relatively better performance on the FB15K237 and WN18RR datasets. This is due to the fact that most existing KG datasets are known to have tree-like or
hierarchical structures and thus favour hyperbolic embeddings~\cite{tay2018hyperbolic,nickel2017poincare,ganea2018hyperbolic}. Therefore, it is suggested to employ hyperbolic-based models for handling datasets (not limited to knowledge graphs) that exhibit hierarchical structure. 
Meanwhile, models that blend various geometric properties (e.g., UltraE~\cite{xiong2022ultrahyperbolic}, DGS~\cite{iyer2022dual}, GIE~\cite{cao2022geometry} and Concept2Box~\cite{huang2023concept2box}) have also achieved promising performance. Nonetheless, it is {worth noting} that employing sophisticated geometric embeddings (e.g., hyperbolic embedding) 
{will lead to} increased computational complexity, resulting in the need for additional training resources. Consequently, caution must be exercised to manage the model expense or implement a justifiable acceleration of the algorithm when utilising advanced geometric embeddings.
(2). The {results} of pattern inference (in Section~\ref{subsec:pattern}) are discussed from an algebraic perspective. The {important} patterns (e.g., \textit{symmetry, antisymmetry, inversion} and \textit{composition}) of the knowledge graphs are {closely related to} the algebraic operations employed by the KGE models. Table~\ref{tab:pattern_inf} reveals that most models that can simultaneously capture all four patterns gain from \textit{Product} operations, such as Hadamard product and Hamilton product. This suggests us to develop more product-based models to capture the relational patterns of KG concurrently. Moreover, exploring innovative algebraic operations under certain conditions is also worthwhile. 
(3). Section~\ref{subsec:other_domain} enumerates a variety of applications in which KGs can be employed across different domains, encompassing intrinsic applications such as KGQA and knowledge reasoning, as well as extrinsic applications such as leveraging KGs in Biomedicine. Concurrently, the recent rise of LLMs~\cite{brown2020language,touvron2023llama,scao2022bloom,cui2023efficient}, which are closely associated with KGs, presents an opportunity for mutual reinforcement. Nevertheless, LLMs trained on general corpora may not effectively generalise to specific domains or novel knowledge owing to the absence of domain-specific knowledge or new training data. In order to solve the above problems, a potential solution is to integrate Knowledge Graphs into LLMs because of their ability to provide accurate and explicit knowledge. It is also well known that KGs have great symbolic reasoning capabilities and can actively evolve with the continuous input of new knowledge. 
Therefore, infusing KGs into other domains, enhancing LLM reasoning with knowledge graphs and interpretability, enriching knowledge graphs with LLMs, and integrating LLMs with KGs are all promising avenues for future research.
}

\section{Future Directions}
\label{sec:futuredirec}
In this section, we summarise some core advantages of mathematical structures and propose several noteworthy future directions, which we hope to inspire the readers to construct more flexible and adaptable KGE algorithms in the future.

\subsection{Algebraic Operation}
The ability of knowledge representation is often directly related to the properties of model's algebraic operations. The algebraic advantages can sometimes compensate for the model's shortcomings, which allow the model to handle complex and non-intuitive relationships, such as 1-N, N-1 and N-M relations. One future direction is to construct a powerful operation that is able to deal with multiple tasks simultaneously. Recent works such as  CompoundE~\cite{ge2022compounde} have started using combination operations to better handle complex relation types in different conditions.
Another promising method is to build a unified algebraic KGE framework~\cite{hayashi2017equivalence,liu2017analogical} for better understanding the nature of algebraic operations. Nevertheless, algebraically inspired KGE models still have many perspectives to ponder.

\subsection{Geometric Embedding}
Several current KGE methods are inspired by some useful geometric properties. For instance, ManifoldE~\cite{xiao2015one} utilise spherical embedding to extend the entity representation from one point to a sphere. ATTH~\cite{chami2020low} provides a hyperbolic embedding for capturing hierarchies based on hyperbolic isometries. However, most knowledge graphs embrace complicated structures which results in these existing methods perform poorly when modelling the knowledge graph with hybrid structures~\cite{cao2022geometry}. Hence, it is crucial to establish more comprehensive geometric embedding methods. Some works provide mixture-curvature embedding~\cite{wang2021mixed} or geometry interactive~\cite{cao2022geometry} KGE to enhance the ability of geometric embedding. Another meaningful research direction is how to build geometric knowledge learning models with low energy, since complex geometric embedding always needs more training time, especially on large-scale KGs. Wang et al.~\cite{wang2021hyperbolic} provides a fewer parameters ``RotH-like~\cite{chami2020low}'' model. However, it is still a worthwhile direction to build a comprehensive geometric embedding model while keeping low energy consumption.

\subsection{Analytical Optimisation}
Analytical properties of knowledge representation are of great importance but poorly studied. Any of these properties, such as convergence, stability, complexity and derivability, etc., will greatly affect the model performance on downstream tasks. For example, the embedding space has to be differentiable, or the model could not be trained by gradient descent. Several works focus on solving certain analytical properties in KGE. TorusE~\cite{ebisu2018toruse} started by analysing convergence, by choosing a compact Lie group as an embedding space, the model never diverges unlimited and regularisation is no longer required. In addition, it is worth paying special attention to build some highly efficient KGE models through optimisation analytically. Peng et al.~\cite{peng2021highly} adopt segmented embeddings to divide the entity representation space into multiple independent sub-spaces explicitly.
Nevertheless, there still has a long way to go to deal with analytical optimisations, so as to construct more reliable KGE models in the future.

\subsection{Others}
It is worth noting that the above three mathematical perspectives are not independent but complementary to each other. Therefore, one possible research direction is to build  multi-mathematical perspective models. FieldE~\cite{nayyeri2021knowledge} employs the neural ODE that represents relations as trajectories connecting neighbouring nodes in the graph, which can continuously (analytical view) represent the underlying geometries including Euclidean, Poincare Ball, Hyperboloid and Spherical (geometric view). Moreover, it is of great significance to note the above summarised implications are \textbf{not limited to the KGE domain, but indeed can be extended to any domains}. Recent works strongly bear this out. Hyperbolic space can be utilised in generative models~\cite{ding2021deep} geometrically, neural order differential equations (NODE) are considered as another breakthrough in deep learning with strong abilities in supervised learning,
sequential diffusion~\cite{jiang2021learning,kang2021stable,thorpe2021grand++}, etc.
{In summary, it is imperative to harness the attributes of mathematical space in order to investigate further potentialities, encompassing not only Knowledge Graph Embeddings but also extending to various other domains.}

\section{Conclusion}
Over the recent years, the rapid growth of Knowledge Graph Embedding has been witnessed. As we have seen in our analysis of the current state of the art, most of the existing works tend to utilise the properties of different spaces to build knowledge representations. Hence, we present a comprehensive review of the state-of-the-art KGE techniques from the perspective of representation spaces, and discuss the existing KGE methods according to three mathematical perspectives: (1) Algebraic perspective, (2) Geometric perspective, and (3) Analytical perspective. In particular, we first introduce the basic concept of mathematical spaces and the relationship between them. We then introduce and compare the strengths of KGE techniques from different mathematical perspectives in terms of embedding space, scoring function, optimisation, etc. At the same time, some unique characteristics of specific spaces are summarised through the experimental results. {For instance, hyperbolic-based models manifest their competence in assimilating hierarchical information.} We also discuss some future directions such as efficient geometric embedding, and rational optimisation methods. {Furthermore, in the context of the burgeoning big model epoch, we propose several prospective avenues for integrating LLMs and KGs.} We believe that it is vital to analyse the model from the mathematical point of view, to determine more precisely what causes the possible failures in each method and find the appropriate mathematics tools to solve them. This article takes the field of KGE as an example to show the powerful role of mathematical spaces and their properties, thereby inspiring more strong mathematical modelling in more fields.

\bibliographystyle{acm}
\bibliography{cite}

\begin{thebibliography}{100}

\bibitem{abboud2020boxe}
{\sc Abboud, R., Ceylan, I., Lukasiewicz, T., and Salvatori, T.}
\newblock Boxe: A box embedding model for knowledge base completion.
\newblock {\em Advances in Neural Information Processing Systems 33\/} (2020), 9649--9661.

\bibitem{amin2020lowfer}
{\sc Amin, S., Varanasi, S., Dunfield, K.~A., and Neumann, G.}
\newblock Lowfer: Low-rank bilinear pooling for link prediction.
\newblock In {\em International Conference on Machine Learning\/} (2020), PMLR, pp.~257--268.

\bibitem{auer2007dbpedia}
{\sc Auer, S., Bizer, C., Kobilarov, G., Lehmann, J., Cyganiak, R., and Ives, Z.}
\newblock Dbpedia: A nucleus for a web of open data.
\newblock In {\em The semantic web}, vol.~4825. Springer, 2007, pp.~722--735.

\bibitem{balazevic2019multi}
{\sc Balazevic, I., Allen, C., and Hospedales, T.}
\newblock Multi-relational poincar{\'e} graph embeddings.
\newblock {\em Advances in Neural Information Processing Systems 32\/} (2019), 4463--4473.

\bibitem{balavzevic2019tucker}
{\sc Bala{\v{z}}evi{\'c}, I., Allen, C., and Hospedales, T.}
\newblock Tucker: Tensor factorization for knowledge graph completion.
\newblock In {\em Conference on Empirical Methods in Natural Language Processing and the International Joint Conference on Natural Language Processing\/} (2019), pp.~5185--5194.

\bibitem{ball1960short}
{\sc Ball, W. W.~R.}
\newblock {\em A short account of the history of mathematics}.
\newblock Courier Corporation, 1960.

\bibitem{bang2023multitask}
{\sc Bang, Y., Cahyawijaya, S., Lee, N., Dai, W., Su, D., Wilie, B., Lovenia, H., Ji, Z., Yu, T., Chung, W., et~al.}
\newblock A multitask, multilingual, multimodal evaluation of $\mathrm{ChatGPT}$ on reasoning, hallucination, and interactivity.
\newblock {\em arXiv preprint arXiv:2302.04023\/} (2023).

\bibitem{bastos2021hopfe}
{\sc Bastos, A., Singh, K., Nadgeri, A., Shekarpour, S., Mulang, I.~O., and Hoffart, J.}
\newblock Hopfe: Knowledge graph representation learning using inverse hopf fibrations.
\newblock In {\em International Conference on Information and Knowledge Management\/} (2021), pp.~89--99.

\bibitem{bellini2017auto}
{\sc Bellini, V., Anelli, V.~W., Di~Noia, T., and Di~Sciascio, E.}
\newblock Auto-encoding user ratings via knowledge graphs in recommendation scenarios.
\newblock In {\em Proceedings of the 2nd Workshop on Deep Learning for Recommender Systems\/} (2017), pp.~60--66.

\bibitem{bodenreider2004unified}
{\sc Bodenreider, O.}
\newblock The unified medical language system (umls): integrating biomedical terminology.
\newblock {\em Nucleic acids research 32}, suppl\_1 (2004), D267--D270.

\bibitem{bollacker2008freebase}
{\sc Bollacker, K., Evans, C., Paritosh, P., Sturge, T., and Taylor, J.}
\newblock Freebase: a collaboratively created graph database for structuring human knowledge.
\newblock In {\em ACM SIGMOD International Conference on Management of Data\/} (2008), pp.~1247--1250.

\bibitem{bordes2014semantic}
{\sc Bordes, A., Glorot, X., Weston, J., and Bengio, Y.}
\newblock A semantic matching energy function for learning with multi-relational data.
\newblock {\em Machine Learning 94}, 2 (2014), 233--259.

\bibitem{bordes2013irreflexive}
{\sc Bordes, A., Usunier, N., Garcia-Duran, A., Weston, J., and Yakhnenko, O.}
\newblock Irreflexive and hierarchical relations as translations.
\newblock {\em arXiv preprint arXiv:1304.7158\/} (2013).

\bibitem{bordes2013translating}
{\sc Bordes, A., Usunier, N., Garcia-Duran, A., Weston, J., and Yakhnenko, O.}
\newblock Translating embeddings for modeling multi-relational data.
\newblock {\em Advances in Neural Information Processing Systems 26\/} (2013).

\bibitem{bordes2011learning}
{\sc Bordes, A., Weston, J., Collobert, R., and Bengio, Y.}
\newblock Learning structured embeddings of knowledge bases.
\newblock In {\em AAAI Conference on Artificial Intelligence\/} (2011), pp.~301--306.

\bibitem{bourbaki1989commutative}
{\sc Bourbaki, N.}
\newblock {\em Commutative algebra: chapters 1-7}.
\newblock Springer, 1989.

\bibitem{brown2020language}
{\sc Brown, T., Mann, B., Ryder, N., Subbiah, M., Kaplan, J.~D., Dhariwal, P., Neelakantan, A., Shyam, P., Sastry, G., Askell, A., et~al.}
\newblock Language models are few-shot learners.
\newblock {\em Advances in Neural Information Processing Systems 33\/} (2020), 1877--1901.

\bibitem{burnaev2021manifold}
{\sc Burnaev, E., and Bernstein, A.}
\newblock Manifold modeling in machine learning.
\newblock {\em Journal of Communications Technology and Electronics 66}, 6 (2021), 754--763.

\bibitem{cao2023relmkg}
{\sc Cao, X., and Liu, Y.}
\newblock Relmkg: reasoning with pre-trained language models and knowledge graphs for complex question answering.
\newblock {\em Applied Intelligence 53}, 10 (2023), 12032--12046.

\bibitem{cao2021dual}
{\sc Cao, Z., Xu, Q., Yang, Z., Cao, X., and Huang, Q.}
\newblock Dual quaternion knowledge graph embeddings.
\newblock In {\em AAAI Conference on Artificial Intelligence\/} (2021), pp.~6894--6902.

\bibitem{cao2022geometry}
{\sc Cao, Z., Xu, Q., Yang, Z., Cao, X., and Huang, Q.}
\newblock Geometry interaction knowledge graph embeddings.
\newblock In {\em AAAI Conference on Artificial Intelligence\/} (2022), pp.~5521--5529.

\bibitem{carvalho2023knowledge}
{\sc Carvalho, R.~M., Oliveira, D., and Pesquita, C.}
\newblock Knowledge graph embeddings for icu readmission prediction.
\newblock {\em BMC Medical Informatics and Decision Making 23}, 1 (2023), 12.

\bibitem{chai2022module}
{\sc Chai, J., and Shi, G.}
\newblock Module: Module embedding for knowledge graphs.
\newblock {\em arXiv preprint arXiv:2203.04702\/} (2022).

\bibitem{chami2020low}
{\sc Chami, I., Wolf, A., Juan, D.-C., Sala, F., Ravi, S., and R{\'e}, C.}
\newblock Low-dimensional hyperbolic knowledge graph embeddings.
\newblock {\em arXiv preprint arXiv:2005.00545\/} (2020).

\bibitem{chao2021pairre}
{\sc Chao, L., He, J., Wang, T., and Chu, W.}
\newblock Pairre: Knowledge graph embeddings via paired relation vectors.
\newblock In {\em Annual Meeting of the Association for Computational Linguistics and International Joint Conference on Natural Language Processing\/} (2021), pp.~4360--4369.

\bibitem{chen2018neural}
{\sc Chen, R.~T., Rubanova, Y., Bettencourt, J., and Duvenaud, D.~K.}
\newblock Neural ordinary differential equations.
\newblock In {\em Advances in Neural Information Processing Systems\/} (2018), pp.~6572--6583.

\bibitem{chen2021mobiuse}
{\sc Chen, Y., Liu, J., Zhang, Z., Wen, S., and Xiong, W.}
\newblock M{\"o}biuse: Knowledge graph embedding on m{\"o}bius ring.
\newblock {\em Knowledge-Based Systems 227\/} (2021), 107181.

\bibitem{cheng2023editing}
{\sc Cheng, S., Zhang, N., Tian, B., Dai, Z., Xiong, F., Guo, W., and Chen, H.}
\newblock Editing language model-based knowledge graph embeddings.
\newblock {\em arXiv preprint arXiv:2301.10405\/} (2023).

\bibitem{cho2014learning}
{\sc Cho, K., Van~Merri{\"e}nboer, B., Gulcehre, C., Bahdanau, D., Bougares, F., Schwenk, H., and Bengio, Y.}
\newblock Learning phrase representations using rnn encoder-decoder for statistical machine translation.
\newblock {\em arXiv preprint arXiv:1406.1078\/} (2014).

\bibitem{choi2023knowledge}
{\sc Choi, B., and Ko, Y.}
\newblock Knowledge graph extension with a pre-trained language model via unified learning method.
\newblock {\em Knowledge-Based Systems 262\/} (2023), 110245.

\bibitem{cui2023efficient}
{\sc Cui, Y., Yang, Z., and Yao, X.}
\newblock Efficient and effective text encoding for chinese llama and alpaca.
\newblock {\em arXiv preprint arXiv:2304.08177\/} (2023).

\bibitem{demailly2012analytic}
{\sc Demailly, J.-P.}
\newblock {\em Analytic methods in algebraic geometry}, vol.~1.
\newblock International Press Somerville, MA, 2012.

\bibitem{dettmers2018convolutional}
{\sc Dettmers, T., Minervini, P., Stenetorp, P., and Riedel, S.}
\newblock Convolutional 2d knowledge graph embeddings.
\newblock In {\em AAAI Conference on Artificial Intelligence\/} (2018), pp.~1811--1818.

\bibitem{devlin2018bert}
{\sc Devlin, J., Chang, M.-W., Lee, K., and Toutanova, K.}
\newblock Bert: Pre-training of deep bidirectional transformers for language understanding.
\newblock {\em arXiv preprint arXiv:1810.04805\/} (2018).

\bibitem{ding2021deep}
{\sc Ding, J., and Regev, A.}
\newblock Deep generative model embedding of single-cell rna-seq profiles on hyperspheres and hyperbolic spaces.
\newblock {\em Nature communications 12}, 1 (2021), 1--17.

\bibitem{dong2015question}
{\sc Dong, L., Wei, F., Zhou, M., and Xu, K.}
\newblock Question answering over freebase with multi-column convolutional neural networks.
\newblock In {\em Annual Meeting of the Association for Computational Linguistics and International Joint Conference on Natural Language Processing\/} (2015), pp.~260--269.

\bibitem{dong2021hypersphere}
{\sc Dong, Y., Guo, X., Xiang, J., Liu, K., and Tang, Z.}
\newblock Hypersphere: An embedding method for knowledge graph completion based on hypersphere.
\newblock In {\em International Conference on Knowledge Science, Engineering and Management\/} (2021), pp.~517--528.

\bibitem{ebisu2018toruse}
{\sc Ebisu, T., and Ichise, R.}
\newblock Toruse: Knowledge graph embedding on a lie group.
\newblock In {\em AAAI Conference on Artificial Intelligence\/} (2018), pp.~1819--1826.

\bibitem{ebisu2019generalized}
{\sc Ebisu, T., and Ichise, R.}
\newblock Generalized translation-based embedding of knowledge graph.
\newblock {\em IEEE Transactions on Knowledge and Data Engineering 32}, 5 (2019), 941--951.

\bibitem{ensan2017document}
{\sc Ensan, F., and Bagheri, E.}
\newblock Document retrieval model through semantic linking.
\newblock In {\em ACM International Conference on Web Search and Data Mining\/} (2017), pp.~181--190.

\bibitem{faith2012algebra}
{\sc Faith, C.}
\newblock {\em Algebra II Ring Theory: Vol. 2: Ring Theory}, vol.~191.
\newblock Springer Science \& Business Media, 2012.

\bibitem{fei2021enriching}
{\sc Fei, H., Ren, Y., Zhang, Y., Ji, D., and Liang, X.}
\newblock Enriching contextualized language model from knowledge graph for biomedical information extraction.
\newblock {\em Briefings in bioinformatics 22}, 3 (2021), 110.

\bibitem{feng2016gake}
{\sc Feng, J., Huang, M., Yang, Y., and Zhu, X.}
\newblock Gake: Graph aware knowledge embedding.
\newblock In {\em International Conference on Computational Linguistics\/} (2016), pp.~641--651.

\bibitem{fitzpatrick2007euclid}
{\sc Fitzpatrick, R.}
\newblock {\em Euclid’s elements of geometry}.
\newblock Euclidis Elementa, 2007.

\bibitem{ganea2018hyperbolic}
{\sc Ganea, O., B{\'e}cigneul, G., and Hofmann, T.}
\newblock Hyperbolic entailment cones for learning hierarchical embeddings.
\newblock In {\em International Conference on Machine Learning\/} (2018), PMLR, pp.~1646--1655.

\bibitem{gao2020rotate3d}
{\sc Gao, C., Sun, C., Shan, L., Lin, L., and Wang, M.}
\newblock Rotate3d: Representing relations as rotations in three-dimensional space for knowledge graph embedding.
\newblock In {\em International Conference on Information and Knowledge Management\/} (2020), pp.~385--394.

\bibitem{gao2021quatde}
{\sc Gao, H., Yang, K., Yang, Y., Zakari, R.~Y., Owusu, J.~W., and Qin, K.}
\newblock Quatde: Dynamic quaternion embedding for knowledge graph completion.
\newblock {\em arXiv preprint arXiv:2105.09002\/} (2021).

\bibitem{gao2021dual}
{\sc Gao, L., Zhu, H., Zhuo, H.~H., and Xu, J.}
\newblock Dual quaternion embeddings for link prediction.
\newblock {\em Applied Sciences 11}, 12 (2021), 5572.

\bibitem{gaur2022iseeq}
{\sc Gaur, M., Gunaratna, K., Srinivasan, V., and Jin, H.}
\newblock Iseeq: Information seeking question generation using dynamic meta-information retrieval and knowledge graphs.
\newblock In {\em AAAI Conference on Artificial Intelligence\/} (2022), pp.~10672--10680.

\bibitem{ge2022compounde}
{\sc Ge, X., Wang, Y.-C., Wang, B., and Kuo, C.-C.~J.}
\newblock Compounde: Knowledge graph embedding with translation, rotation and scaling compound operations.
\newblock {\em arXiv preprint arXiv:2207.05324\/} (2022).

\bibitem{ge2023knowledge}
{\sc Ge, X., Wang, Y.-C., Wang, B., and Kuo, C.-C.~J.}
\newblock Knowledge graph embedding with 3d compound geometric transformations.
\newblock {\em arXiv preprint arXiv:2304.00378\/} (2023).

\bibitem{ge2022core}
{\sc Ge, X., Wang, Y.-C., Wang, B., and Kuo, C.~J.}
\newblock Core: A knowledge graph entity type prediction method via complex space regression and embedding.
\newblock {\em Pattern Recognition Letters 157\/} (2022), 97--103.

\bibitem{georgi1974unity}
{\sc Georgi, H., and Glashow, S.~L.}
\newblock Unity of all elementary-particle forces.
\newblock {\em Physical Review Letters 32}, 8 (1974), 438.

\bibitem{goldman1988varieties}
{\sc Goldman, W.~M.}
\newblock varieties of representations.
\newblock In {\em Geometry of Group Representations: Proceedings of the AMS-IMS-SIAM Joint Summer Research Conference\/} (1988), p.~169.

\bibitem{gregucci2023link}
{\sc Gregucci, C., Nayyeri, M., Hern{\'a}ndez, D., and Staab, S.}
\newblock Link prediction with attention applied on multiple knowledge graph embedding models.
\newblock In {\em International Conference on World Wide Web\/} (2023), pp.~2600--2610.

\bibitem{greub2012linear}
{\sc Greub, W.~H.}
\newblock {\em Linear algebra}, vol.~23.
\newblock Springer Science \& Business Media, 2012.

\bibitem{gu2018learning}
{\sc Gu, A., Sala, F., Gunel, B., and R{\'e}, C.}
\newblock Learning mixed-curvature representations in product spaces.
\newblock In {\em International Conference for Learning Representation\/} (2018).

\bibitem{guan2019knowledge}
{\sc Guan, N., Song, D., and Liao, L.}
\newblock Knowledge graph embedding with concepts.
\newblock {\em Knowledge-Based Systems 164\/} (2019), 38--44.

\bibitem{guo2021bique}
{\sc Guo, J., and Kok, S.}
\newblock Bique: Biquaternionic embeddings of knowledge graphs.
\newblock In {\em Conference on Empirical Methods in Natural Language Processing\/} (2021), pp.~8338--8351.

\bibitem{guo2019learning}
{\sc Guo, L., Sun, Z., and Hu, W.}
\newblock Learning to exploit long-term relational dependencies in knowledge graphs.
\newblock In {\em International Conference on Machine Learning\/} (2019), PMLR, pp.~2505--2514.

\bibitem{han2021temporal}
{\sc Han, Z., Ding, Z., Ma, Y., Gu, Y., and Tresp, V.}
\newblock Temporal knowledge graph forecasting with neural ode.
\newblock {\em arXiv preprint arXiv:2101.05151\/} (2021).

\bibitem{hao2019universal}
{\sc Hao, J., Chen, M., Yu, W., Sun, Y., and Wang, W.}
\newblock Universal representation learning of knowledge bases by jointly embedding instances and ontological concepts.
\newblock In {\em ACM SIGKDD Conference on Knowledge Discovery and Data Mining\/} (2019), pp.~1709--1719.

\bibitem{harris1998spherical}
{\sc Harris, J., and Stocker, H.}
\newblock Spherical geometry.
\newblock {\em Handbook of mathematics and computational science\/} (1998), 108--113.

\bibitem{hayashi2017equivalence}
{\sc Hayashi, K., and Shimbo, M.}
\newblock On the equivalence of holographic and complex embeddings for link prediction.
\newblock {\em arXiv preprint arXiv:1702.05563\/} (2017).

\bibitem{he2015learning}
{\sc He, S., Liu, K., Ji, G., and Zhao, J.}
\newblock Learning to represent knowledge graphs with gaussian embedding.
\newblock In {\em International Conference on Information and Knowledge Management\/} (2015), pp.~623--632.

\bibitem{hochreiter1997long}
{\sc Hochreiter, S., and Schmidhuber, J.}
\newblock Long short-term memory.
\newblock {\em Neural computation 9}, 8 (1997), 1735--1780.

\bibitem{hoffmann2011knowledge}
{\sc Hoffmann, R., Zhang, C., Ling, X., Zettlemoyer, L., and Weld, D.~S.}
\newblock Knowledge-based weak supervision for information extraction of overlapping relations.
\newblock In {\em Proceedings of the 49th annual meeting of the association for computational linguistics: human language technologies\/} (2011), pp.~541--550.

\bibitem{10.1145/3447772}
{\sc Hogan, A., Blomqvist, E., Cochez, M., D’amato, C., Melo, G.~D., Gutierrez, C., Kirrane, S., Gayo, J. E.~L., Navigli, R., Neumaier, S., Ngomo, A.-C.~N., Polleres, A., Rashid, S.~M., Rula, A., Schmelzeisen, L., Sequeda, J., Staab, S., and Zimmermann, A.}
\newblock Knowledge graphs.
\newblock {\em ACM Comput. Surv. 54}, 4 (2021).

\bibitem{host2023constructing}
{\sc H{\o}st, A.~M., Lison, P., and Moonen, L.}
\newblock Constructing a knowledge graph from textual descriptions of software vulnerabilities in the national vulnerability database.
\newblock {\em arXiv preprint arXiv:2305.00382\/} (2023).

\bibitem{householder1958unitary}
{\sc Householder, A.~S.}
\newblock Unitary triangularization of a nonsymmetric matrix.
\newblock {\em Journal of the ACM 5}, 4 (1958), 339--342.

\bibitem{huang2021knowledge}
{\sc Huang, X., Tang, J., Tan, Z., Zeng, W., Wang, J., and Zhao, X.}
\newblock Knowledge graph embedding by relational and entity rotation.
\newblock {\em Knowledge-Based Systems 229\/} (2021), 107310.

\bibitem{huang2023concept2box}
{\sc Huang, Z., Wang, D., Huang, B., Zhang, C., Shang, J., Liang, Y., Wang, Z., Li, X., Faloutsos, C., Sun, Y., et~al.}
\newblock Concept2box: Joint geometric embeddings for learning two-view knowledge graphs.
\newblock {\em arXiv preprint arXiv:2307.01933\/} (2023).

\bibitem{iyer2022dual}
{\sc Iyer, R.~G., Bai, Y., Wang, W., and Sun, Y.}
\newblock Dual-geometric space embedding model for two-view knowledge graphs.
\newblock In {\em ACM SIGKDD Conference on Knowledge Discovery and Data Mining\/} (2022), pp.~676--686.

\bibitem{izacard2020leveraging}
{\sc Izacard, G., and Grave, E.}
\newblock Leveraging passage retrieval with generative models for open domain question answering.
\newblock In {\em Proceedings of the 16th Conference of the European Chapter of the Association for Computational Linguistics: Main Volume, {EACL}\/} (2021), pp.~874--880.

\bibitem{ji2015knowledge}
{\sc Ji, G., He, S., Xu, L., Liu, K., and Zhao, J.}
\newblock Knowledge graph embedding via dynamic mapping matrix.
\newblock In {\em Annual Meeting of the Association for Computational Linguistics and International Joint Conference on Natural Language Processing\/} (2015), pp.~687--696.

\bibitem{ji2016knowledge}
{\sc Ji, G., Liu, K., He, S., and Zhao, J.}
\newblock Knowledge graph completion with adaptive sparse transfer matrix.
\newblock In {\em AAAI Conference on Artificial Intelligence\/} (2016), pp.~985--991.

\bibitem{ji2021survey}
{\sc Ji, S., Pan, S., Cambria, E., Marttinen, P., and Philip, S.~Y.}
\newblock A survey on knowledge graphs: Representation, acquisition, and applications.
\newblock {\em IEEE Transactions on Neural Networks and Learning Systems 33}, 2 (2022), 494--514.

\bibitem{ji2023survey}
{\sc Ji, Z., Lee, N., Frieske, R., Yu, T., Su, D., Xu, Y., Ishii, E., Bang, Y.~J., Madotto, A., and Fung, P.}
\newblock Survey of hallucination in natural language generation.
\newblock {\em ACM Computing Surveys 55}, 12 (2023), 1--38.

\bibitem{jiang2021learning}
{\sc Jiang, B., Zhang, Y., Wei, X., Xue, X., and Fu, Y.}
\newblock Learning compositional representation for 4d captures with neural ode.
\newblock In {\em IEEE/CVF Conference on Computer Vision and Pattern Recognition\/} (2021), pp.~5340--5350.

\bibitem{jiang20how}
{\sc Jiang, Z., Xu, F.~F., Araki, J., and Neubig, G.}
\newblock How can we know what language models know.
\newblock {\em Trans. Assoc. Comput. Linguistics 8\/} (2020), 423--438.

\bibitem{kang2021stable}
{\sc Kang, Q., Song, Y., Ding, Q., and Tay, W.~P.}
\newblock Stable neural ode with lyapunov-stable equilibrium points for defending against adversarial attacks.
\newblock {\em Advances in Neural Information Processing Systems\/} (2021), 14925--14937.

\bibitem{kazemi2018simple}
{\sc Kazemi, S.~M., and Poole, D.}
\newblock Simple embedding for link prediction in knowledge graphs.
\newblock {\em Advances in Neural Information Processing Systems 31\/} (2018).

\bibitem{kipf2016semi}
{\sc Kipf, T.~N., and Welling, M.}
\newblock Semi-supervised classification with graph convolutional networks.
\newblock {\em arXiv preprint arXiv:1609.02907\/} (2016).

\bibitem{kobyzev2020normalizing}
{\sc Kobyzev, I., Prince, S.~J., and Brubaker, M.~A.}
\newblock Normalizing flows: An introduction and review of current methods.
\newblock {\em IEEE Transactions on Pattern Analysis and Machine Intelligence 43}, 11 (2020), 3964--3979.

\bibitem{lacroix2018canonical}
{\sc Lacroix, T., Usunier, N., and Obozinski, G.}
\newblock Canonical tensor decomposition for knowledge base completion.
\newblock In {\em International Conference on Machine Learning\/} (2018), pp.~2863--2872.

\bibitem{lan2020query}
{\sc Lan, Y., and Jiang, J.}
\newblock Query graph generation for answering multi-hop complex questions from knowledge bases.
\newblock In {\em Annual Meeting of the Association for Computational Linguistics\/} (2020), pp.~969--974.

\bibitem{lang2012algebra}
{\sc Lang, S.}
\newblock {\em Algebra}, vol.~211.
\newblock Springer Science \& Business Media, 2012.

\bibitem{le2023knowledge}
{\sc Le, T., Huynh, N., and Le, B.}
\newblock Knowledge graph embedding by projection and rotation on hyperplanes for link prediction.
\newblock {\em Applied Intelligence 53}, 9 (2023), 10340--10364.

\bibitem{le2023rotate4D}
{\sc Le, T., Tran, H., and Le, B.}
\newblock Knowledge graph embedding with the special orthogonal group in quaternion space for link prediction.
\newblock {\em Knowledge-Based Systems 266\/} (2023), 110400.

\bibitem{lee2010introduction}
{\sc Lee, J.}
\newblock {\em Introduction to topological manifolds}, vol.~202.
\newblock Springer Science \& Business Media, 2010.

\bibitem{lewis2019bart}
{\sc Lewis, M., Liu, Y., Goyal, N., Ghazvininejad, M., Mohamed, A., Levy, O., Stoyanov, V., and Zettlemoyer, L.}
\newblock $\mathrm{BART:}$ denoising sequence-to-sequence pre-training for natural language generation, translation, and comprehension.
\newblock In {\em Proceedings of the 58th Annual Meeting of the Association for Computational Linguistics, {ACL}\/} (2020), pp.~7871--7880.

\bibitem{li2023position}
{\sc Li, G., Sun, Z., Hu, W., Cheng, G., and Qu, Y.}
\newblock Position-aware relational transformer for knowledge graph embedding.
\newblock {\em IEEE Transactions on Neural Networks and Learning Systems\/} (2023).

\bibitem{li2022pretrained}
{\sc Li, J., Tang, T., Zhao, W.~X., Nie, J.-Y., and Wen, J.-R.}
\newblock Pretrained language models for text generation: A survey.
\newblock {\em arXiv preprint arXiv:2201.05273\/} (2022).

\bibitem{li2020real}
{\sc Li, L., Wang, P., Yan, J., Wang, Y., Li, S., Jiang, J., Sun, Z., Tang, B., Chang, T.-H., Wang, S., et~al.}
\newblock Real-world data medical knowledge graph: construction and applications.
\newblock {\em Artificial Intelligence in Medicine 103\/} (2020), 101817.

\bibitem{li2022house}
{\sc Li, R., Zhao, J., Li, C., He, D., Wang, Y., Liu, Y., Sun, H., Wang, S., Deng, W., Shen, Y., et~al.}
\newblock House: Knowledge graph embedding with householder parameterization.
\newblock In {\em International Conference on Machine Learning\/} (2022), PMLR, pp.~13209--13224.

\bibitem{li2021robust}
{\sc Li, Y., Kong, D., Zhang, Y., Tan, Y., and Chen, L.}
\newblock Robust deep alignment network with remote sensing knowledge graph for zero-shot and generalized zero-shot remote sensing image scene classification.
\newblock {\em ISPRS Journal of Photogrammetry and Remote Sensing 179\/} (2021), 145--158.

\bibitem{li2021efficient}
{\sc Li, Z., Ji, J., Fu, Z., Ge, Y., Xu, S., Chen, C., and Zhang, Y.}
\newblock Efficient non-sampling knowledge graph embedding.
\newblock In {\em International Conference on World Wide Web\/} (2021), pp.~1727--1736.

\bibitem{li2023transo}
{\sc Li, Z., Liu, X., Wang, X., Liu, P., and Shen, Y.}
\newblock Transo: a knowledge-driven representation learning method with ontology information constraints.
\newblock {\em International Conference on World Wide Web 26}, 1 (2023), 297--319.

\bibitem{li2023k}
{\sc Li, Z.-X., Li, Y.-J., Liu, Y.-W., Liu, C., and Zhou, N.-X.}
\newblock K-ctiaa: Automatic analysis of cyber threat intelligence based on a knowledge graph.
\newblock {\em Symmetry 15}, 2 (2023), 337.

\bibitem{liang2016neural}
{\sc Liang, C., Berant, J., Le, Q., Forbus, K.~D., and Lao, N.}
\newblock Neural symbolic machines: Learning semantic parsers on freebase with weak supervision.
\newblock {\em arXiv preprint arXiv:1611.00020\/} (2016).

\bibitem{liang2019collaborative}
{\sc Liang, S.}
\newblock Collaborative, dynamic and diversified user profiling.
\newblock In {\em Proceedings of the AAAI Conference on Artificial Intelligence\/} (2019), vol.~33, pp.~4269--4276.

\bibitem{liang2019unsupervised}
{\sc Liang, S.}
\newblock Unsupervised semantic generative adversarial networks for expert retrieval.
\newblock In {\em The world wide web conference\/} (2019), pp.~1039--1050.

\bibitem{liang2021profiling}
{\sc Liang, S., Luo, Y., and Meng, Z.}
\newblock Profiling users for question answering communities via flow-based constrained co-embedding model.
\newblock {\em ACM Transactions on Information Systems 40}, 2 (2021), 1--38.

\bibitem{liang2021cross}
{\sc Liang, S., Tang, S., Meng, Z., and Zhang, Q.}
\newblock Cross-temporal snapshot alignment for dynamic networks.
\newblock {\em IEEE Transactions on Knowledge and Data Engineering\/} (2021).

\bibitem{liang2016dynamic}
{\sc Liang, S., Yilmaz, E., and Kanoulas, E.}
\newblock Dynamic clustering of streaming short documents.
\newblock In {\em Proceedings of the 22nd ACM SIGKDD international conference on knowledge discovery and data mining\/} (2016), pp.~995--1004.

\bibitem{liao2021learning}
{\sc Liao, S., Liang, S., Meng, Z., and Zhang, Q.}
\newblock Learning dynamic embeddings for temporal knowledge graphs.
\newblock In {\em ACM International Conference on Web Search and Data Mining\/} (2021), pp.~535--543.

\bibitem{lin19kagnet}
{\sc Lin, B.~Y., Chen, X., Chen, J., and Ren, X.}
\newblock Kagnet: Knowledge-aware graph networks for commonsense reasoning.
\newblock In {\em Proceedings of the 2019 Conference on Empirical Methods in Natural Language Processing and the 9th International Joint Conference on Natural Language Processing, {EMNLP-IJCNLP}\/} (2019), pp.~2829--2839.

\bibitem{lin2015modeling}
{\sc Lin, Y., Liu, Z., Luan, H., Sun, M., Rao, S., and Liu, S.}
\newblock Modeling relation paths for representation learning of knowledge bases.
\newblock {\em arXiv preprint arXiv:1506.00379\/} (2015).

\bibitem{lin2015learning}
{\sc Lin, Y., Liu, Z., Sun, M., Liu, Y., and Zhu, X.}
\newblock Learning entity and relation embeddings for knowledge graph completion.
\newblock In {\em AAAI Conference on Artificial Intelligence\/} (2015), pp.~2181--2187.

\bibitem{liu2017analogical}
{\sc Liu, H., Wu, Y., and Yang, Y.}
\newblock Analogical inference for multi-relational embeddings.
\newblock In {\em International Conference on Machine Learning\/} (2017), pp.~2168--2178.

\bibitem{liu2015latent}
{\sc Liu, X., and Fang, H.}
\newblock Latent entity space: a novel retrieval approach for entity-bearing queries.
\newblock {\em Information Retrieval Journal 18}, 6 (2015), 473--503.

\bibitem{long2020integrated}
{\sc Long, J., Chen, Z., He, W., Wu, T., and Ren, J.}
\newblock An integrated framework of deep learning and knowledge graph for prediction of stock price trend: An application in chinese stock exchange market.
\newblock {\em Applied Soft Computing 91\/} (2020), 106205.

\bibitem{lou2020differentiating}
{\sc Lou, A., Katsman, I., Jiang, Q., Belongie, S., Lim, S.-N., and De~Sa, C.}
\newblock Differentiating through the fr{\'e}chet mean.
\newblock In {\em International Conference on Machine Learning\/} (2020), pp.~6393--6403.

\bibitem{lu2020utilizing}
{\sc Lu, F., Cong, P., and Huang, X.}
\newblock Utilizing textual information in knowledge graph embedding: A survey of methods and applications.
\newblock {\em IEEE Access 8\/} (2020), 92072--92088.

\bibitem{lu2020dense}
{\sc Lu, H., and Hu, H.}
\newblock Dense: An enhanced non-abelian group representation for knowledge graph embedding.
\newblock {\em arXiv preprint arXiv:2008.04548\/} (2020).

\bibitem{lv2018differentiating}
{\sc Lv, X., Hou, L., Li, J., and Liu, Z.}
\newblock Differentiating concepts and instances for knowledge graph embedding.
\newblock {\em arXiv preprint arXiv:1811.04588\/} (2018).

\bibitem{ma2023conte}
{\sc Ma, T., Li, M., Lv, S., Zhu, F., Huang, L., and Hu, S.}
\newblock Conte: contextualized knowledge graph embedding for circular relations.
\newblock {\em Data Mining and Knowledge Discovery 37}, 1 (2023), 110--135.

\bibitem{marcheggiani2017encoding}
{\sc Marcheggiani, D., and Titov, I.}
\newblock Encoding sentences with graph convolutional networks for semantic role labeling.
\newblock In {\em Conference on Empirical Methods in Natural Language Processing\/} (2017), pp.~1506--1515.

\bibitem{meng2021mixture}
{\sc Meng, Z., Liu, F., Clark, T.~H., Shareghi, E., and Collier, N.}
\newblock Mixture-of-partitions: Infusing large biomedical knowledge graphs into bert.
\newblock {\em arXiv preprint arXiv:2109.04810\/} (2021).

\bibitem{meng21mixture}
{\sc Meng, Z., Liu, F., Clark, T.~H., Shareghi, E., and Collier, N.}
\newblock Mixture-of-partitions: Infusing large biomedical knowledge graphs into {BERT}.
\newblock In {\em Proceedings of the 2021 Conference on Empirical Methods in Natural Language Processing, {EMNLP}\/} (2021), pp.~4672--4681.

\bibitem{meng22rewire}
{\sc Meng, Z., Liu, F., Shareghi, E., Su, Y., Collins, C., and Collier, N.}
\newblock Rewire-then-probe: {A} contrastive recipe for probing biomedical knowledge of pre-trained language models.
\newblock In {\em Proceedings of the 60th Annual Meeting of the Association for Computational Linguistics (Volume 1: Long Papers), {ACL}\/} (2022), pp.~4798--4810.

\bibitem{mohamed2021biological}
{\sc Mohamed, S.~K., Nounu, A., and Nov{\'a}{\v{c}}ek, V.}
\newblock Biological applications of knowledge graph embedding models.
\newblock {\em Briefings in Bioinformatics 22}, 2 (2021), 1679--1693.

\bibitem{murphy2012machine}
{\sc Murphy, K.~P.}
\newblock {\em Machine learning: a probabilistic perspective}.
\newblock MIT press, 2012.

\bibitem{nayyeri2023knowledge}
{\sc Nayyeri, M., Xiong, B., Mohammadi, M., Akter, M.~M., Alam, M.~M., Lehmann, J., and Staab, S.}
\newblock Knowledge graph embeddings using neural ito process: From multiple walks to stochastic trajectories.
\newblock In {\em Findings of the Association for Computational Linguistics\/} (2023), pp.~7165--7179.

\bibitem{nayyeri2021knowledge}
{\sc Nayyeri, M., Xu, C., Hoffmann, F., Alam, M.~M., Lehmann, J., and Vahdati, S.}
\newblock Knowledge graph representation learning using ordinary differential equations.
\newblock In {\em Conference on Empirical Methods in Natural Language Processing\/} (2021), pp.~9529--9548.

\bibitem{nickel2017poincare}
{\sc Nickel, M., and Kiela, D.}
\newblock Poincar{\'e} embeddings for learning hierarchical representations.
\newblock {\em Advances in Neural Information Processing Systems 30\/} (2017), 6338--6347.

\bibitem{nickel2018learning}
{\sc Nickel, M., and Kiela, D.}
\newblock Learning continuous hierarchies in the lorentz model of hyperbolic geometry.
\newblock In {\em International Conference on Machine Learning\/} (2018), PMLR, pp.~3779--3788.

\bibitem{nickel2015review}
{\sc Nickel, M., Murphy, K., Tresp, V., and Gabrilovich, E.}
\newblock A review of relational machine learning for knowledge graphs.
\newblock {\em Proceedings of the IEEE 104}, 1 (2015), 11--33.

\bibitem{nickel2016holographic}
{\sc Nickel, M., Rosasco, L., and Poggio, T.}
\newblock Holographic embeddings of knowledge graphs.
\newblock In {\em AAAI Conference on Artificial Intelligence\/} (2016), pp.~1955--1961.

\bibitem{nickel2011three}
{\sc Nickel, M., Tresp, V., and Kriegel, H.-P.}
\newblock A three-way model for collective learning on multi-relational data.
\newblock In {\em International Conference on Machine Learning\/} (2011), pp.~809--816.

\bibitem{pan2023unifying}
{\sc Pan, S., Luo, L., Wang, Y., Chen, C., Wang, J., and Wu, X.}
\newblock Unifying large language models and knowledge graphs: A roadmap.
\newblock {\em arXiv preprint arXiv:2306.08302\/} (2023).

\bibitem{pan2021hyperbolic}
{\sc Pan, Z., and Wang, P.}
\newblock Hyperbolic hierarchy-aware knowledge graph embedding for link prediction.
\newblock In {\em Findings of Conference on Empirical Methods in Natural Language Processing\/} (2021), pp.~2941--2948.

\bibitem{pavlovic2022expressive}
{\sc Pavlovi{\'c}, A., and Sallinger, E.}
\newblock Expressive: A spatio-functional embedding for knowledge graph completion.
\newblock {\em arXiv preprint arXiv:2206.04192\/} (2022).

\bibitem{peng2021hyperbolic}
{\sc Peng, W., Varanka, T., Mostafa, A., Shi, H., and Zhao, G.}
\newblock Hyperbolic deep neural networks: A survey.
\newblock {\em arXiv preprint arXiv:2101.04562\/} (2021).

\bibitem{peng2021highly}
{\sc Peng, X., Chen, G., Lin, C., and Stevenson, M.}
\newblock Highly efficient knowledge graph embedding learning with orthogonal procrustes analysis.
\newblock {\em arXiv preprint arXiv:2104.04676\/} (2021).

\bibitem{peng2020lineare}
{\sc Peng, Y., and Zhang, J.}
\newblock Lineare: Simple but powerful knowledge graph embedding for link prediction.
\newblock In {\em IEEE International Conference on Data Mining\/} (2020), IEEE, pp.~422--431.

\bibitem{petroni19language}
{\sc Petroni, F., Rockt{\"{a}}schel, T., Riedel, S., Lewis, P. S.~H., Bakhtin, A., Wu, Y., and Miller, A.~H.}
\newblock Language models as knowledge bases?
\newblock In {\em Proceedings of the 2019 Conference on Empirical Methods in Natural Language Processing and the 9th International Joint Conference on Natural Language Processing, {EMNLP-IJCNLP}\/} (2019), pp.~2463--2473.

\bibitem{pflaum2001analytic}
{\sc Pflaum, M.}
\newblock {\em Analytic and geometric study of stratified spaces: contributions to analytic and geometric aspects}.
\newblock No.~1768. Springer Science \& Business Media, 2001.

\bibitem{qiu2020stepwise}
{\sc Qiu, Y., Wang, Y., Jin, X., and Zhang, K.}
\newblock Stepwise reasoning for multi-relation question answering over knowledge graph with weak supervision.
\newblock In {\em ACM International Conference on Web Search and Data Mining\/} (2020), pp.~474--482.

\bibitem{raffel2020exploring}
{\sc Raffel, C., Shazeer, N., Roberts, A., Lee, K., Narang, S., Matena, M., Zhou, Y., Li, W., and Liu, P.~J.}
\newblock Exploring the limits of transfer learning with a unified text-to-text transformer.
\newblock {\em The Journal of Machine Learning Research 21}, 1 (2020), 5485--5551.

\bibitem{raviv2016document}
{\sc Raviv, H., Kurland, O., and Carmel, D.}
\newblock Document retrieval using entity-based language models.
\newblock In {\em ACM SIGIR Conference on Research and Development in Information Retrieval\/} (2016), pp.~65--74.

\bibitem{robinson1938representations}
{\sc Robinson, G. d.~B.}
\newblock On the representations of the symmetric group.
\newblock {\em American Journal of Mathematics\/} (1938), 745--760.

\bibitem{rosset2020knowledge}
{\sc Rosset, C., Xiong, C., Phan, M., Song, X., Bennett, P., and Tiwary, S.}
\newblock Knowledge-aware language model pretraining.
\newblock {\em arXiv preprint arXiv:2007.00655\/} (2020).

\bibitem{rotman2000first}
{\sc Rotman, J.~J.}
\newblock {\em A first course in abstract algebra}.
\newblock Pearson College Division, 2000.

\bibitem{rudin1973functional}
{\sc Rudin, W.}
\newblock Functional analysis, 1973.

\bibitem{safavi2020evaluating}
{\sc Safavi, T., Koutra, D., and Meij, E.}
\newblock Evaluating the calibration of knowledge graph embeddings for trustworthy link prediction.
\newblock In {\em Empirical Methods in Natural Language Processing\/} (2020), pp.~8308--8321.

\bibitem{sala2018representation}
{\sc Sala, F., De~Sa, C., Gu, A., and R{\'e}, C.}
\newblock Representation tradeoffs for hyperbolic embeddings.
\newblock In {\em International Conference on Machine Learning\/} (2018), pp.~4460--4469.

\bibitem{saxena2020improving}
{\sc Saxena, A., Tripathi, A., and Talukdar, P.}
\newblock Improving multi-hop question answering over knowledge graphs using knowledge base embeddings.
\newblock In {\em Annual Meeting of the Association for Computational Linguistics\/} (2020), pp.~4498--4507.

\bibitem{scao2022bloom}
{\sc Scao, T.~L., Fan, A., Akiki, C., Pavlick, E., Ili{\'c}, S., Hesslow, D., Castagn{\'e}, R., Luccioni, A.~S., Yvon, F., Gall{\'e}, M., et~al.}
\newblock Bloom: A 176b-parameter open-access multilingual language model.
\newblock {\em arXiv preprint arXiv:2211.05100\/} (2022).

\bibitem{schlichtkrull2018modeling}
{\sc Schlichtkrull, M., Kipf, T.~N., Bloem, P., Van Den~Berg, R., Titov, I., and Welling, M.}
\newblock Modeling relational data with graph convolutional networks.
\newblock In {\em European Semantic Web Conference\/} (2018), Springer, pp.~593--607.

\bibitem{shang2019end}
{\sc Shang, C., Tang, Y., Huang, J., Bi, J., He, X., and Zhou, B.}
\newblock End-to-end structure-aware convolutional networks for knowledge base completion.
\newblock In {\em AAAI Conference on Artificial Intelligence\/} (2019), vol.~33, pp.~3060--3067.

\bibitem{shen20exploiting}
{\sc Shen, T., Mao, Y., He, P., Long, G., Trischler, A., and Chen, W.}
\newblock Exploiting structured knowledge in text via graph-guided representation learning.
\newblock In {\em Proceedings of the 2020 Conference on Empirical Methods in Natural Language Processing, {EMNLP}\/} (2020).

\bibitem{shen2012linden}
{\sc Shen, W., Wang, J., Luo, P., and Wang, M.}
\newblock Linden: linking named entities with knowledge base via semantic knowledge.
\newblock In {\em International Conference on World Wide Web\/} (2012), pp.~449--458.

\bibitem{socher2013reasoning}
{\sc Socher, R., Chen, D., Manning, C.~D., and Ng, A.}
\newblock Reasoning with neural tensor networks for knowledge base completion.
\newblock {\em Advances in Neural Information Processing Systems\/} (2013), 926--934.

\bibitem{suchanek2007yago}
{\sc Suchanek, F.~M., Kasneci, G., and Weikum, G.}
\newblock Yago: a core of semantic knowledge.
\newblock In {\em International Conference on World Wide Web\/} (2007), pp.~697--706.

\bibitem{sun2020multi}
{\sc Sun, R., Cao, X., Zhao, Y., Wan, J., Zhou, K., Zhang, F., Wang, Z., and Zheng, K.}
\newblock Multi-modal knowledge graphs for recommender systems.
\newblock In {\em International Conference on Information and Knowledge Management\/} (2020), pp.~1405--1414.

\bibitem{sun2020knowledge}
{\sc Sun, Z., Chen, M., Hu, W., Wang, C., Dai, J., and Zhang, W.}
\newblock Knowledge association with hyperbolic knowledge graph embeddings.
\newblock In {\em Conference on Empirical Methods in Natural Language Processing\/} (2020), pp.~5704--5716.

\bibitem{sun2019rotate}
{\sc Sun, Z., Deng, Z.-H., Nie, J.-Y., and Tang, J.}
\newblock Rotate: Knowledge graph embedding by relational rotation in complex space.
\newblock {\em arXiv preprint arXiv:1902.10197\/} (2019).

\bibitem{sung2021can}
{\sc Sung, M., Lee, J., Yi, S., Jeon, M., Kim, S., and Kang, J.}
\newblock Can language models be biomedical knowledge bases?
\newblock {\em arXiv preprint arXiv:2109.07154\/} (2021).

\bibitem{suzuki2018riemannian}
{\sc Suzuki, A., Enokida, Y., and Yamanishi, K.}
\newblock Riemannian transe: Multi-relational graph embedding in non-euclidean space.

\bibitem{tabacof2019probability}
{\sc Tabacof, P., and Costabello, L.}
\newblock Probability calibration for knowledge graph embedding models.
\newblock In {\em International Conference on Learning Representations\/} (2019).

\bibitem{tang2020orthogonal}
{\sc Tang, Y., Huang, J., Wang, G., He, X., and Zhou, B.}
\newblock Orthogonal relation transforms with graph context modeling for knowledge graph embedding.
\newblock In {\em Annual Meeting of the Association for Computational Linguistics\/} (2020), pp.~2713--2722.

\bibitem{tay2018hyperbolic}
{\sc Tay, Y., Tuan, L.~A., and Hui, S.~C.}
\newblock Hyperbolic representation learning for fast and efficient neural question answering.
\newblock In {\em ACM International Conference on Web Search and Data Mining\/} (2018), pp.~583--591.

\bibitem{thorpe2021grand++}
{\sc Thorpe, M., Nguyen, T.~M., Xia, H., Strohmer, T., Bertozzi, A., Osher, S., and Wang, B.}
\newblock Grand++: Graph neural diffusion with a source term.
\newblock In {\em International Conference for Learning Representation\/} (2021).

\bibitem{toutanova2015observed}
{\sc Toutanova, K., and Chen, D.}
\newblock Observed versus latent features for knowledge base and text inference.
\newblock In {\em Proceedings of the 3rd workshop on continuous vector space models and their compositionality\/} (2015), pp.~57--66.

\bibitem{touvron2023llama}
{\sc Touvron, H., Lavril, T., Izacard, G., Martinet, X., Lachaux, M.-A., Lacroix, T., Rozi{\`e}re, B., Goyal, N., Hambro, E., Azhar, F., et~al.}
\newblock Llama: Open and efficient foundation language models.
\newblock {\em arXiv preprint arXiv:2302.13971\/} (2023).

\bibitem{treves2016topological}
{\sc Treves, F.}
\newblock {\em Topological Vector Spaces, Distributions and Kernels: Pure and Applied Mathematics, Vol. 25}, vol.~25.
\newblock Elsevier, 2016.

\bibitem{trouillon2016complex}
{\sc Trouillon, T., Welbl, J., Riedel, S., Gaussier, {\'E}., and Bouchard, G.}
\newblock Complex embeddings for simple link prediction.
\newblock In {\em International Conference on Machine Learning\/} (2016), pp.~2071--2080.

\bibitem{ungar2001hyperbolic}
{\sc Ungar, A.~A.}
\newblock Hyperbolic trigonometry and its application in the poincar{\'e} ball model of hyperbolic geometry.
\newblock {\em Computers \& Mathematics with Applications 41}, 1-2 (2001), 135--147.

\bibitem{vashishth2019composition}
{\sc Vashishth, S., Sanyal, S., Nitin, V., and Talukdar, P.}
\newblock Composition-based multi-relational graph convolutional networks.
\newblock {\em arXiv preprint arXiv:1911.03082\/} (2019).

\bibitem{vaswani2017attention}
{\sc Vaswani, A., Shazeer, N., Parmar, N., Uszkoreit, J., Jones, L., Gomez, A.~N., Kaiser, {\L}., and Polosukhin, I.}
\newblock Attention is all you need.
\newblock {\em Advances in Neural Information Processing Systems 30\/} (2017).

\bibitem{vilela2023biomedical}
{\sc Vilela, J., Asif, M., Marques, A.~R., Santos, J.~X., Rasga, C., Vicente, A., and Martiniano, H.}
\newblock Biomedical knowledge graph embeddings for personalized medicine: Predicting disease-gene associations.
\newblock {\em Expert Systems 40}, 5 (2023), e13181.

\bibitem{vilnis2014word}
{\sc Vilnis, L., and McCallum, A.}
\newblock Word representations via gaussian embedding.
\newblock {\em arXiv preprint arXiv:1412.6623\/} (2014).

\bibitem{vrandevcic2014wikidata}
{\sc Vrande{\v{c}}i{\'c}, D., and Kr{\"o}tzsch, M.}
\newblock Wikidata: a free collaborative knowledgebase.
\newblock {\em Communications of the ACM 57}, 10 (2014), 78--85.

\bibitem{wan2021gaussianpath}
{\sc Wan, G., and Du, B.}
\newblock Gaussianpath: A bayesian multi-hop reasoning framework for knowledge graph reasoning.
\newblock In {\em AAAI Conference on Artificial Intelligence\/} (2021), vol.~35, pp.~4393--4401.

\bibitem{wang2022interht}
{\sc Wang, B., Meng, Q., Wang, Z., Wu, D., Che, W., Wang, S., Chen, Z., and Liu, C.}
\newblock Interht: Knowledge graph embeddings by interaction between head and tail entities.
\newblock {\em arXiv preprint arXiv:2202.04897\/} (2022).

\bibitem{wang2022dirie}
{\sc Wang, F., Zhang, Z., Sun, L., Ye, J., and Yan, Y.}
\newblock Dirie: knowledge graph embedding with dirichlet distribution.
\newblock In {\em International Conference on World Wide Web\/} (2022), pp.~3082--3091.

\bibitem{wang2019multi}
{\sc Wang, H., Zhang, F., Zhao, M., Li, W., Xie, X., and Guo, M.}
\newblock Multi-task feature learning for knowledge graph enhanced recommendation.
\newblock In {\em International Conference on World Wide Web\/} (2019), pp.~2000--2010.

\bibitem{wang2021hyperbolic}
{\sc Wang, K., Liu, Y., Lin, D., and Sheng, M.}
\newblock Hyperbolic geometry is not necessary: Lightweight euclidean-based models for low-dimensional knowledge graph embeddings.
\newblock In {\em Findings of Conference on Empirical Methods in Natural Language Processing\/} (2021), pp.~464--474.

\bibitem{wang2018incorporating}
{\sc Wang, P., Li, S., and Pan, R.}
\newblock Incorporating gan for negative sampling in knowledge representation learning.
\newblock In {\em AAAI Conference on Artificial Intelligence\/} (2018), vol.~32.

\bibitem{wang2017knowledge}
{\sc Wang, Q., Mao, Z., Wang, B., and Guo, L.}
\newblock Knowledge graph embedding: A survey of approaches and applications.
\newblock {\em IEEE Transactions on Knowledge and Data Engineering 29}, 12 (2017), 2724--2743.

\bibitem{wang2015knowledge}
{\sc Wang, Q., Wang, B., and Guo, L.}
\newblock Knowledge base completion using embeddings and rules.
\newblock In {\em International Joint Conference on Artificial Intelligence\/} (2015).

\bibitem{wang2021hierarchical}
{\sc Wang, S., Fu, K., Sun, X., Zhang, Z., Li, S., and Jin, L.}
\newblock Hierarchical-aware relation rotational knowledge graph embedding for link prediction.
\newblock {\em Neurocomputing 458\/} (2021), 259--270.

\bibitem{wang2022stke}
{\sc Wang, S., Liu, R., Shen, L., and Khattak, A.~M.}
\newblock Stke: Temporal knowledge graph embedding in the spherical coordinate system.
\newblock In {\em International Conference on Artificial Intelligence and Security\/} (2022), pp.~292--305.

\bibitem{wang2021knowledge}
{\sc Wang, S., Wei, X., Dos~Santos, C.~N., Wang, Z., Nallapati, R., Arnold, A., and Philip, S.~Y.}
\newblock Knowledge graph representation via hierarchical hyperbolic neural graph embedding.
\newblock In {\em IEEE International Conference on Big Data\/} (2021), pp.~540--549.

\bibitem{wang2021mixed}
{\sc Wang, S., Wei, X., Nogueira~dos Santos, C.~N., Wang, Z., Nallapati, R., Arnold, A., Xiang, B., Yu, P.~S., and Cruz, I.~F.}
\newblock Mixed-curvature multi-relational graph neural network for knowledge graph completion.
\newblock In {\em International Conference on World Wide Web\/} (2021), pp.~1761--1771.

\bibitem{wang21kepler}
{\sc Wang, X., Gao, T., Zhu, Z., Zhang, Z., Liu, Z., Li, J., and Tang, J.}
\newblock {KEPLER:} {A} unified model for knowledge embedding and pre-trained language representation.
\newblock {\em Trans. Assoc. Comput. Linguistics 9\/} (2021), 176--194.

\bibitem{wang2021kepler}
{\sc Wang, X., Gao, T., Zhu, Z., Zhang, Z., Liu, Z., Li, J., and Tang, J.}
\newblock Kepler: A unified model for knowledge embedding and pre-trained language representation.
\newblock {\em Transactions of the Association for Computational Linguistics 9\/} (2021), 176--194.

\bibitem{wang2019kgat}
{\sc Wang, X., He, X., Cao, Y., Liu, M., and Chua, T.-S.}
\newblock Kgat: Knowledge graph attention network for recommendation.
\newblock In {\em ACM SIGKDD Conference on Knowledge Discovery and Data Mining\/} (2019), pp.~950--958.

\bibitem{wang2014knowledge}
{\sc Wang, Z., Zhang, J., Feng, J., and Chen, Z.}
\newblock Knowledge graph embedding by translating on hyperplanes.
\newblock In {\em AAAI Conference on Artificial Intelligence\/} (2014), pp.~1112--1119.

\bibitem{weber2018curvature}
{\sc Weber, M., and Nickel, M.}
\newblock Curvature and representation learning: Identifying embedding spaces for relational data.
\newblock {\em Advances in Neural Information Processing Systems\/} (2018).

\bibitem{wilkinson2013linear}
{\sc Wilkinson, J.~H., Bauer, F.~L., and Reinsch, C.}
\newblock {\em Linear algebra}, vol.~2.
\newblock Springer, 2013.

\bibitem{wise2020covid}
{\sc Wise, C., Calvo, M.~R., Bhatia, P., Ioannidis, V., Karypus, G., Price, G., Song, X., Brand, R., and Kulkani, N.}
\newblock Covid-19 knowledge graph: Accelerating information retrieval and discovery for scientific literature.
\newblock In {\em Proceedings of Knowledgeable NLP: the First Workshop on Integrating Structured Knowledge and Neural Networks for NLP\/} (2020), pp.~1--10.

\bibitem{wu2010open}
{\sc Wu, F., and Weld, D.~S.}
\newblock Open information extraction using wikipedia.
\newblock In {\em Annual Meeting of the Association for Computational Linguistics\/} (2010), pp.~118--127.

\bibitem{wu2022learning}
{\sc Wu, Y., Lu, B., Tian, L., and Liang, S.}
\newblock Learning to co-embed queries and documents.
\newblock {\em Electronics 11}, 22 (2022), 3694.

\bibitem{xiao2015transa}
{\sc Xiao, H., Huang, M., Hao, Y., and Zhu, X.}
\newblock Transa: An adaptive approach for knowledge graph embedding.
\newblock {\em arXiv preprint arXiv:1509.05490\/} (2015).

\bibitem{xiao2015transg}
{\sc Xiao, H., Huang, M., Hao, Y., and Zhu, X.}
\newblock Transg: A generative mixture model for knowledge graph embedding.
\newblock {\em arXiv preprint arXiv:1509.05488\/} (2015).

\bibitem{xiao2015one}
{\sc Xiao, H., Huang, M., and Zhu, X.}
\newblock From one point to a manifold: Knowledge graph embedding for precise link prediction.
\newblock {\em arXiv preprint arXiv:1512.04792\/} (2015).

\bibitem{xiao2022complex}
{\sc Xiao, H., Liu, X., Song, Y., Wong, G.~Y., and See, S.}
\newblock Complex hyperbolic knowledge graph embeddings with fast fourier transform.
\newblock {\em arXiv preprint arXiv:2211.03635\/} (2022).

\bibitem{xie2016DKRL}
{\sc Xie, R., Liu, Z., Jia, J., Luan, H., and Sun, M.}
\newblock Representation learning of knowledge graphs with entity descriptions.
\newblock In {\em AAAI Conference on Artificial Intelligence\/} (2016), vol.~30.

\bibitem{xie2017image}
{\sc Xie, R., Liu, Z., Luan, H., and Sun, M.}
\newblock Image-embodied knowledge representation learning.
\newblock In {\em International Joint Conference on Artificial Intelligence\/} (2017), pp.~3140--3146.

\bibitem{xie2016representation}
{\sc Xie, R., Liu, Z., Sun, M., et~al.}
\newblock Representation learning of knowledge graphs with hierarchical types.
\newblock In {\em Annual Meeting of the Association for Computational Linguistics and International Joint Conference on Natural Language Processing\/} (2016), pp.~2965--2971.

\bibitem{xie2016TKRL}
{\sc Xie, R., Liu, Z., Sun, M., et~al.}
\newblock Representation learning of knowledge graphs with hierarchical types.
\newblock In {\em International Joint Conference on Artificial Intelligence\/} (2016), vol.~2016, pp.~2965--2971.

\bibitem{xiong2022ultrahyperbolic}
{\sc Xiong, B., Zhu, S., Nayyeri, M., Xu, C., Pan, S., Zhou, C., and Staab, S.}
\newblock Ultrahyperbolic knowledge graph embeddings.
\newblock {\em arXiv preprint arXiv:2206.00449\/} (2022).

\bibitem{xiong2017deeppath}
{\sc Xiong, W., Hoang, T., and Wang, W.~Y.}
\newblock Deeppath: A reinforcement learning method for knowledge graph reasoning.
\newblock {\em arXiv preprint arXiv:1707.06690\/} (2017).

\bibitem{xu2019relation}
{\sc Xu, C., and Li, R.}
\newblock Relation embedding with dihedral group in knowledge graph.
\newblock In {\em Annual Meeting of the Association for Computational Linguistics\/} (2019), pp.~263--272.

\bibitem{xu2016question}
{\sc Xu, K., Reddy, S., Feng, Y., Huang, S., and Zhao, D.}
\newblock Question answering on freebase via relation extraction and textual evidence.
\newblock {\em arXiv preprint arXiv:1603.00957\/} (2016).

\bibitem{xu2021understanding}
{\sc Xu, M.}
\newblock Understanding graph embedding methods and their applications.
\newblock {\em SIAM Review 63}, 4 (2021), 825--853.

\bibitem{yang2014embedding}
{\sc Yang, B., Yih, W.-t., He, X., Gao, J., and Deng, L.}
\newblock Embedding entities and relations for learning and inference in knowledge bases.
\newblock {\em arXiv preprint arXiv:1412.6575\/} (2014).

\bibitem{yang2021knowledge}
{\sc Yang, H., and Liu, J.}
\newblock Knowledge graph representation learning as groupoid: unifying transe, rotate, quate, complex.
\newblock In {\em International Conference on Information and Knowledge Management\/} (2021), pp.~2311--2320.

\bibitem{yang2020nage}
{\sc Yang, T., Sha, L., and Hong, P.}
\newblock Nage: Non-abelian group embedding for knowledge graphs.
\newblock In {\em International Conference on Information and Knowledge Management\/} (2020), pp.~1735--1742.

\bibitem{yao2019kg}
{\sc Yao, L., Mao, C., and Luo, Y.}
\newblock Kg-bert: Bert for knowledge graph completion.
\newblock {\em arXiv preprint arXiv:1909.03193\/} (2019).

\bibitem{yao2023improving}
{\sc Yao, T., Wang, Y., Zhang, K., and Liang, S.}
\newblock Improving the expressiveness of k-hop message-passing gnns by injecting contextualized substructure information.
\newblock In {\em Proceedings of the 29th ACM SIGKDD Conference on Knowledge Discovery and Data Mining\/} (2023), pp.~3070--3081.

\bibitem{yasunaga2021qa}
{\sc Yasunaga, M., Ren, H., Bosselut, A., Liang, P., and Leskovec, J.}
\newblock Qa-gnn: Reasoning with language models and knowledge graphs for question answering.
\newblock {\em arXiv preprint arXiv:2104.06378\/} (2021).

\bibitem{yasunaga21qa}
{\sc Yasunaga, M., Ren, H., Bosselut, A., Liang, P., and Leskovec, J.}
\newblock {QA-GNN:} reasoning with language models and knowledge graphs for question answering.
\newblock In {\em Proceedings of the 2021 Conference of the North American Chapter of the Association for Computational Linguistics: Human Language Technologies, {NAACL-HLT}\/} (2021), pp.~535--546.

\bibitem{yih2015semantic}
{\sc Yih, S. W.-t., Chang, M.-W., He, X., and Gao, J.}
\newblock Semantic parsing via staged query graph generation: Question answering with knowledge base.
\newblock In {\em Joint Conference of the Annual Meeting of the ACL and International Joint Conference on Natural Language Processing of the AFNLP\/} (2015), pp.~1321--1331.

\bibitem{yu2022jaket}
{\sc Yu, D., Zhu, C., Yang, Y., and Zeng, M.}
\newblock Jaket: Joint pre-training of knowledge graph and language understanding.
\newblock In {\em AAAI Conference on Artificial Intelligence\/} (2022), pp.~11630--11638.

\bibitem{yu2021mquade}
{\sc Yu, J., Cai, Y., Sun, M., and Li, P.}
\newblock Mquade: a unified model for knowledge fact embedding.
\newblock In {\em International Conference on World Wide Web\/} (2021), pp.~3442--3452.

\bibitem{yu2022triplere}
{\sc Yu, L., Luo, Z., Liu, H., Lin, D., Li, H., and Deng, Y.}
\newblock Triplere: Knowledge graph embeddings via tripled relation vectors.
\newblock {\em arXiv preprint arXiv:2209.08271\/} (2022).

\bibitem{zhang2020few}
{\sc Zhang, C., Yao, H., Huang, C., Jiang, M., Li, Z., and Chawla, N.~V.}
\newblock Few-shot knowledge graph completion.
\newblock In {\em AAAI Conference on Artificial Intelligence\/} (2020), pp.~3041--3048.

\bibitem{zhang2021knowledge}
{\sc Zhang, Q., Wang, R., Yang, J., and Xue, L.}
\newblock Knowledge graph embedding by translating in time domain space for link prediction.
\newblock {\em Knowledge-Based Systems 212\/} (2021), 106564.

\bibitem{zhang2022knowledge}
{\sc Zhang, Q., Wang, R., Yang, J., and Xue, L.}
\newblock Knowledge graph embedding by reflection transformation.
\newblock {\em Knowledge-Based Systems 238\/} (2022), 107861.

\bibitem{zhang2022structural}
{\sc Zhang, Q., Wang, R., Yang, J., and Xue, L.}
\newblock Structural context-based knowledge graph embedding for link prediction.
\newblock {\em Neurocomputing 470\/} (2022), 109--120.

\bibitem{zhang2021drug}
{\sc Zhang, R., Hristovski, D., Schutte, D., Kastrin, A., Fiszman, M., and Kilicoglu, H.}
\newblock Drug repurposing for covid-19 via knowledge graph completion.
\newblock {\em Journal of biomedical informatics 115\/} (2021), 103696.

\bibitem{zhang2019quaternion}
{\sc Zhang, S., Tay, Y., Yao, L., and Liu, Q.}
\newblock Quaternion knowledge graph embeddings.
\newblock {\em Advances in Neural Information Processing Systems\/} (2019), 2731--2741.

\bibitem{zhang2013optimizing}
{\sc Zhang, W., Chen, T., Wang, J., and Yu, Y.}
\newblock Optimizing top-n collaborative filtering via dynamic negative item sampling.
\newblock In {\em ACM SIGIR Conference on Research and Development in Information Retrieval\/} (2013), pp.~785--788.

\bibitem{zhang2019interaction}
{\sc Zhang, W., Paudel, B., Zhang, W., Bernstein, A., and Chen, H.}
\newblock Interaction embeddings for prediction and explanation in knowledge graphs.
\newblock In {\em ACM International Conference on Web Search and Data Mining\/} (2019), pp.~96--104.

\bibitem{zhang22greaselm}
{\sc Zhang, X., Bosselut, A., Yasunaga, M., Ren, H., Liang, P., Manning, C.~D., and Leskovec, J.}
\newblock Greaselm: Graph reasoning enhanced language models.
\newblock In {\em International Conference on Learning Representations\/} (2022).

\bibitem{zhang2022trans}
{\sc Zhang, X., Yang, Q., and Xu, D.}
\newblock Trans: Transition-based knowledge graph embedding with synthetic relation representation.
\newblock {\em arXiv preprint arXiv:2204.08401\/} (2022).

\bibitem{zhang2020interstellar}
{\sc Zhang, Y., Yao, Q., and Chen, L.}
\newblock Interstellar: Searching recurrent architecture for knowledge graph embedding.
\newblock {\em Advances in Neural Information Processing Systems 33\/} (2020), 10030--10040.

\bibitem{zhang2020learning}
{\sc Zhang, Z., Cai, J., Zhang, Y., and Wang, J.}
\newblock Learning hierarchy-aware knowledge graph embeddings for link prediction.
\newblock In {\em AAAI Conference on Artificial Intelligence\/} (2020), pp.~3065--3072.

\bibitem{zhang2019ernie}
{\sc Zhang, Z., Han, X., Liu, Z., Jiang, X., Sun, M., and Liu, Q.}
\newblock Ernie: Enhanced language representation with informative entities.
\newblock In {\em Annual Meeting of the Association for Computational Linguistics\/} (2019), pp.~1441--1451.

\bibitem{zhang2020pretrain}
{\sc Zhang, Z., Liu, X., Zhang, Y., Su, Q., Sun, X., and He, B.}
\newblock Pretrain-kge: learning knowledge representation from pretrained language models.
\newblock In {\em Findings of the Association for Computational Linguistics\/} (2020), pp.~259--266.

\bibitem{zhang2021fine}
{\sc Zhang, Z., Parulian, N., Ji, H., Elsayed, A., Myers, S., and Palmer, M.}
\newblock Fine-grained information extraction from biomedical literature based on knowledge-enriched abstract meaning representation.
\newblock In {\em Annual Meeting of the Association for Computational Linguistics and International Joint Conference on Natural Language Processing\/} (2021), pp.~6261--6270.

\bibitem{zhang2018knowledge}
{\sc Zhang, Z., Zhuang, F., Qu, M., Lin, F., and He, Q.}
\newblock Knowledge graph embedding with hierarchical relation structure.
\newblock In {\em Conference on Empirical Methods in Natural Language Processing\/} (2018), pp.~3198--3207.

\bibitem{zhao2017meta}
{\sc Zhao, H., Yao, Q., Li, J., Song, Y., and Lee, D.~L.}
\newblock Meta-graph based recommendation fusion over heterogeneous information networks.
\newblock In {\em ACM SIGKDD Conference on Knowledge Discovery and Data Mining\/} (2017), pp.~635--644.

\bibitem{zheng2022hyperbolic}
{\sc Zheng, W., Wang, W., Qian, F., Zhao, S., and Zhang, Y.}
\newblock Hyperbolic hierarchical knowledge graph embeddings for link prediction in low dimensions.
\newblock {\em arXiv preprint arXiv:2204.13704\/} (2022).

\bibitem{zheng2021knowledge}
{\sc Zheng, W., Yin, L., Chen, X., Ma, Z., Liu, S., and Yang, B.}
\newblock Knowledge base graph embedding module design for visual question answering model.
\newblock {\em Pattern Recognition 120\/} (2021), 108153.

\bibitem{zhou2020improving}
{\sc Zhou, K., Zhao, W.~X., Bian, S., Zhou, Y., Wen, J.-R., and Yu, J.}
\newblock Improving conversational recommender systems via knowledge graph based semantic fusion.
\newblock In {\em ACM SIGKDD Conference on Knowledge Discovery and Data Mining\/} (2020), pp.~1006--1014.

\end{thebibliography}
\end{document}